\documentclass[12pt,letterpaper,twoside,reqno,nosumlimits]{amsart}
\usepackage{times}
\usepackage[margin=1in]{geometry}
\usepackage{etoolbox}
\usepackage{commath}

\usepackage[export]{adjustbox}
\newcommand*\diff{\mathop{}\!\mathrm{d}}
\usepackage{thmtools}
\patchcmd{\section}{\scshape}{\bfseries}{}{}
\makeatletter
\renewcommand{\@secnumfont}{\bfseries}
\makeatother

\usepackage[dvipsnames]{xcolor}
\patchcmd{\section}{\normalfont}{\normalfont\color{MidnightBlue}}{}{}
\patchcmd{\subsection}{\normalfont}{\normalfont\color{MidnightBlue}}{}{}

\makeatletter
\def\subsubsection{\@startsection{subsubsection}{3}%
\z@{.5\linespacing\@plus.7\linespacing}{-.5em}%
{\normalfont\bfseries}}
\makeatother

\usepackage{fancyhdr}
\usepackage{amsmath,amsfonts,amsbsy,amsgen,amscd,mathrsfs,amssymb,amsthm,mathtools,tensor}

\usepackage{todonotes}
\usepackage{tikz}
\usepackage{overpic}

\usepackage{stmaryrd}

\usepackage{amsopn}

\DeclareMathOperator*{\argmin}{arg\,min}

\usepackage{caption}
\usepackage{graphicx}
\usepackage{color}
\usepackage{url}
\usepackage{epstopdf}
\usepackage{pdfsync}
\usepackage[colorlinks=true,
            linkcolor=black,
            urlcolor=black,
            citecolor=black]{hyperref}
\usepackage{comment}
\usepackage{mathabx}
\usepackage{upgreek}

\usepackage{caption}
\usepackage{subcaption}

\usepackage{mathrsfs}
\usepackage{yfonts}
\hypersetup{
    linkcolor=blue,
}

\newlength{\fixboxwidth}
\usepackage{epsfig,amsbsy,graphicx,multirow}

\usepackage{ algorithm, algorithmic}
\renewcommand{\algorithmiccomment}[1]{\bgroup\hfill//~#1\egroup}

\usepackage[all,cmtip]{xy}

\usepackage{accents}

\def\R{\mathbb{R}}

\def\U{\mathcal{U}}

\newcommand{\Matern}[3]{K_{#1}\!\left(#2;#3\right)} 
\newcommand{\BesselK}{\operatorname{K}}              

\def\restrict#1{\raise-.5ex\hbox{\ensuremath|}_{#1}}

\def\<{\big\langle}
\def\>{\big\rangle}

\def\Tr{\operatorname{Tr}}

\def\det{\operatorname{det}}

\def\argmin{{\operatorname{argmin}}}

\definecolor{red}{rgb}{0.9, 0, 0}

\newtheorem{Theorem}{Theorem}[section]

\declaretheorem[name=Remark,sibling=Theorem,qed=\qedsymbol]{Remark}

\makeatletter
\newcommand{\oset}[3][0ex]{%
  \mathrel{\mathop{#3}\limits^{
    \vbox to#1{\kern-2\ex@
    \hbox{$\scriptstyle#2$}\vss}}}}
\makeatother

\makeatletter
\newcommand{\uset}[3][0ex]{%
  \mathrel{\mathop{#3}\limits_{
    \vbox to#1{\kern-2\ex@
    \hbox{$\scriptstyle#2$}\vss}}}}
\makeatother

\usepackage{aurical}
\usepackage[T1]{fontenc}

\usepackage{cleveref}

\usepackage{tikz,pgflibraryshapes}
\usetikzlibrary{shadows}
\usetikzlibrary{arrows}

\newcounter{MatthieuCounter}
\newcommand{\matthieu}[1]{{\small \color{purple}
		\refstepcounter{MatthieuCounter}\textsf{[Matthieu]$_{\arabic{MatthieuCounter}}$:{#1}}}}

\newcommand{\corrauthor}{\textsuperscript{*}}

\newcommand{\XY}[1]{\textcolor{black}{#1}}      
\newcommand{\GE}[1]{\textcolor{black}{#1}}     
\newcommand{\HO}[1]{\textcolor{black}{#1}}    

\begin{document}

\title[Solving Functional PDEs with Gaussian Processes]{Solving Functional PDEs with Gaussian Processes and applications to Functional Renormalization Group Equations}

\author[X.~Yang]{Xianjin Yang\corrauthor}
\address[X.~Yang]{%
  Computing \& Mathematical Sciences, Caltech, Pasadena, CA 91125, USA}
\email[X.~Yang]{yxjmath@caltech.edu}
\thanks{\corrauthor Corresponding author: \texttt{yxjmath@caltech.edu}.}

\author[M.~Darcy]{Matthieu Darcy}
\address[M.~Darcy]{%
  Computing \& Mathematical Sciences, Caltech, Pasadena, CA 91125, USA}
\email[M.~Darcy]{mdarcy@caltech.edu}

\author[M.~Hudes]{Matthew Hudes}
\address[M.~Hudes]{%
  Department of Applied Mathematics \& Statistics, Johns Hopkins University, Baltimore, MD 21218, USA}
\email[M.~Hudes]{mhudes1@jh.edu}

\author[F.~J.~Alexander]{Francis J. Alexander}
\address[F.~J.~Alexander]{Argonne National Laboratory, Lemont, IL 60439, USA}
\email[F.~J.~Alexander]{fja@anl.gov}

\author[G.~Eyink]{Gregory Eyink}
\address[G.~Eyink]{%
  Department of Applied Mathematics \& Statistics, Johns Hopkins University, Baltimore, MD 21218, USA}
\email[G.~Eyink]{eyink@jhu.edu}

\author[H.~Owhadi]{Houman Owhadi}
\address[H.~Owhadi]{%
  Computing \& Mathematical Sciences, Caltech, Pasadena, CA 91125, USA}
\email[H.~Owhadi]{owhadi@caltech.edu}

\date{\today}

\begin{abstract}
We present an operator learning framework for solving non-perturbative functional renormalization group equations, which are integro-differential equations defined on functionals. Our proposed approach uses Gaussian process operator learning to construct a flexible functional representation formulated directly on function space, making it independent of a particular equation or discretization. Our method is flexible, and can apply to a broad range of functional differential equations while still allowing for the incorporation of physical priors in either the prior mean or the kernel design. We demonstrate the performance of our method on several relevant equations, such as the Wetterich and Wilson--Polchinski equations, showing that it achieves equal or better performance than existing approximations such as the local-potential approximation, while being significantly more flexible. In particular, our method can handle non-constant fields, making it promising for the study of more complex field configurations, such as instantons.


\end{abstract}

\maketitle

\section{Introduction}
Functional renormalization group (FRG) equations provide a powerful non-perturbative framework for studying quantum and statistical field theories, enabling interpolation between microscopic and macroscopic descriptions of physical systems \cite{Delamotte_2012,dupuis2021nonperturbative}. Mathematically, they take the form of exact functional integro-differential equations for 
\GE{field-theoretic generating functionals}, but solving these analytically is generally intractable, requiring numerical approximations. Traditional approaches rely on the truncation of expansions, 
such as the derivative expansion or loop expansion, restricting their validity to specific applications where those approximations are accurate. Moreover, numerical schemes are often tailored to particular models, requiring detailed knowledge of the underlying physics. While there have been efforts to propose general and flexible approaches to solving FRG equations \cite{flowpy}, including machine learning methods \cite{PINN-FRG}, existing frameworks are either tied to a specific discretization of the underlying continuous functional PDE,  restricted to specific fields, or rely upon a prescribed ansatz.

In this work, we introduce an operator learning perspective for solving FRG equations. Our framework, based on Gaussian processes (GPs), directly treats FRG flows in function space, sidestepping discretization dependence and enabling application across a wide range of FRG equations. This represents a step toward a general and flexible methodology for functional differential equations in theoretical physics. The remainder of this introduction reviews operator learning, provides an overview of FRG equations, surveys existing numerical methods, and outlines how our approach links the two.

\subsection{Functional data and operator learning}
\label{sec: operator learning}

The functional regression problem, where we aim to learn a mapping with functional inputs and outputs, has emerged as an important problem within the scientific machine learning literature where many applications naturally have functions as data points. A major challenge in this context is how to effectively adapt existing learning methods to this setting. The earliest kernel-based approach to functional regression appears in~\cite{kadriopvaluedkernel}, which generalizes vector-valued function learning to the setting of operator-valued kernels (see also \cite{owhadi2020ideas}, which extends these concepts to compositions of operator-valued Gaussian processes). More recently, there has also been renewed interest in the field of operator learning, especially in the context of accelerating the numerical computation of partial differential equations (PDEs). This problem can be traced back to the use of reduced basis methods \cite{almroth1978automatic, lucia2004reduced, noor1980reduced}, and in more recent years, the development of neural network methods such as DeepONet \cite{ lu2019deeponet}, the Fourier neural operator \cite{li2020fourier}, the convolutional neural operators \cite{convolutionalneuraloperatorsrobust}, or PCA-Net \cite{pcanet}, have extended neural networks to functional inputs and outputs. However, the theoretical analysis of these methods is often limited to proving an existence theorem without guarantees about whether these networks are computable in practice. To address this, a general kernel-based operator-learning framework is introduced in~\cite{kernelmethodsoperatorlearning}, together with theoretical guarantees and convergence rates. A notable feature of such frameworks is their discretization-agnostic character~\cite{li2020fourier}: they are formulated independently of any specific discretization and can therefore be applied to data given on different grids at prediction time.

Although the functional data in these problems often arise from PDE models, most operator-learning approaches treat the task purely as a regression problem and do not exploit the PDE structure that generates the data. In parallel, there has been extensive development of machine learning methods for approximating solutions of PDEs without relying on paired input-output data: the governing equations are enforced directly through a physics-informed loss function. Representative examples include physics-informed neural networks (PINNs)~\cite{PINNs, PINNstraining} and kernel- or GP-based schemes~\cite{chen2021solving, mou2022numerical, meng2023sparse}. Kernel methods in this setting also enjoy stronger theoretical guarantees~\cite{kernelroughpde, BATLLE2025113488}.

While there has been a growing interest in combining elements of both approaches, for example by combining operator learning with a physics-informed loss \cite{physicsinformedneuraloperatorlearning}, there has been little research into the applications of machine learning methods to \textit{partial differential equations defined over functionals}, i.e., mappings  from functions to a real number. Solving such equations requires elements from both of the outlined problems above: the ability to effectively deal with functions as inputs to the algorithm and the ability to impose the PDE constraints. Such a problem naturally arises in the context of FRG equations  which define integro-differential equations over functionals. We briefly describe these equations in the next section. Recently, deep learning methods have been applied to problems related to the FRG flow. For example, \cite{NeuralMCFRG, NeuralODE_learning_FRG} learn the flow directly from data, and \cite{NNinversereFRG} uses a simple one layer convolutional neural network to learn the inverse of flow, from coarse scale to fine scale. A physics-informed neural network (PINN) \cite{PINNs} has recently been proposed to solve FRG equations on a lattice \cite{PINN-FRG}, thereby reducing the problem to solving a PDE in a high-dimensional lattice space. The approach presented in the latter paper is analogous to the \textit{discretize then optimize} paradigm where the infinite-dimensional problem is first discretized into a finite dimensional problem and solved through existing methods. In this paper, we take an \textit{optimize-then-discretize}    perspective: we formulate  a solution of the flow using the governing equations directly in function space before discretizing. To the best of our knowledge, there is no pre-existing \GE{FRG solution} framework formulated directly on function space. This formulation has the potential to enable rigorous a priori error analysis taking into account the various sources of error, similar to the one presented in \cite{kernelmethodsoperatorlearning}.

\subsection{Functional renormalization group equations}
\label{sec:frg-equations-pres}

The nonperturbative FRG provides a
framework for interpolating between microscopic and macroscopic
descriptions of many systems, from condensed matter to statistical field
theory.  A central object in this framework is a family of scale-dependent
functionals, such as the effective average action.
These functionals evolve with the renormalization scale \(\kappa\) and
encode how fluctuations at momenta above \(\kappa\) are progressively
integrated out. Two widely used FRG flow equations are the Wetterich--Morris equation
\begin{align}\label{eq: WM renorm}
  \partial_\kappa \Gamma_\kappa[\phi]
  &= \frac{1}{2}\int_{x,y} \partial_\kappa R_\kappa(x,y)\,
     \Bigl( \Gamma_\kappa^{(2)}[\phi]
            + R_\kappa \Bigr)^{-1}(x,y)\diff{x}\diff{y},
\end{align}
which governs the evolution of the effective average action
\(\Gamma_\kappa\), and the Wilson--Polchinski equation
\begin{equation}\label{eq:physWPE_intro}
\partial_\kappa W_\kappa[J]
=\frac{1}{2}\int_{x, y}\partial_\kappa R_\kappa(x,y)
\left(\frac{\delta^2 W_\kappa[J]}{\delta J_x \delta J_y}
      +\frac{\delta W_\kappa[J]}{\delta J_x}
             \frac{\delta W_\kappa[J]}{\delta J_y}\right)\,\mathrm{d}x\,\mathrm{d}y,
\end{equation}
which describes the corresponding flow of the cumulant-generating functional 
\(W_\kappa\), related to the effective-average action by a generalized Legendre transform duality relation
\begin{equation} \Gamma_\kappa[\phi] + W_\kappa[J] = \int_x J_x \phi_x  
-\frac{1}{2}\int_{x,y} R_\kappa(x,y) \phi_x\phi_y. \label{legendre} \end{equation}

\noindent Here, $\phi$ and $J$ are fields (functions), \(R_\kappa\) is a scale-dependent regulator kernel that suppresses
long-wavelength modes below the running scale \(\kappa\), and
\(\Gamma_\kappa^{(2)}[\phi]\) denotes the \GE{operator defined by the} second functional derivative of
\(\Gamma_\kappa\) with respect to \(\phi\). For clarity of exposition, in what follows we primarily formulate the discussion in terms of the Wetterich equation and its \GE{solution, the} effective average action \(\Gamma_\kappa\). The corresponding formulations for the Wilson--Polchinski flow \(W_\kappa\) are entirely analogous, and in \Cref{sec: numerical results} we include experiments for both formulations.

In typical applications one prescribes an initial condition for \eqref{eq: WM renorm} at a large
\emph{ultraviolet} (UV) scale \(\kappa_{\mathrm{UV}}\).  For such large values of
\(\kappa\), the regulator strongly suppresses \GE{fluctuations at all resolved scales}, 
so that the description remains close to the underlying microscopic
model.  At this scale one sets $\Gamma_{\kappa_{\mathrm{UV}}} = S_{\mathrm{bare}}$,
 where \(S_{\mathrm{bare}}\), called the \emph{bare action}, is a prescribed
action functional specifying the microscopic model (for example, the local
interactions and parameters before fluctuations are integrated out).
The flow
equation is then integrated in \(\kappa\) towards smaller \emph{infrared} (IR) scales
\(\kappa_{\mathrm{IR}}>0\), corresponding to lower momenta and larger
distances.  Along this flow, \(\Gamma_\kappa\) evolves from a microscopic
description to an effective macroscopic one.  Analytic solutions are available
only in very simple models, so practical computations typically rely on
truncations or modeling assumptions tailored to the problem at hand.

From a mathematical perspective, equations such as
\eqref{eq: WM renorm} and \eqref{eq:physWPE_intro} are examples of functional
PDEs.  Abstractly, one may view \(\Gamma_\kappa\) as a functional that assigns
a real number to each field configuration \(\phi\), and
\(\kappa \mapsto \Gamma_\kappa\) as an evolution in the scale variable on some
interval \(I \subset \mathbb{R}\).  We write this in the generic form
\begin{equation}\label{eq:generic-functional-PDE}
  \partial_\kappa \Gamma_\kappa[\phi]
  = F\Bigl(
      \kappa,\,
      \phi,\,
      \Gamma_\kappa[\phi],\,
      \frac{\delta \Gamma_\kappa}{\delta \phi}[\phi],\,
      \frac{\delta^2 \Gamma_\kappa}{\delta \phi^2}[\phi]
    \Bigr),
  \qquad
  \kappa \in I,
\end{equation}
with an initial condition prescribed at some reference scale
\(\kappa_0 \in I\),
\begin{equation}\label{eq:generic-functional-IC}
  \Gamma_{\kappa_0}[\phi] = \Gamma_{\kappa_0}^{\mathrm{init}}[\phi].
\end{equation}
Here \(\Gamma_{\kappa_0}^{\mathrm{init}}\) is a prescribed functional that encodes
the initial data of the problem (for instance, the bare action in an FRG
flow, or a given energy functional in other applications).
The map \(F\) is a (typically nonlocal) functional that may involve spatial
integrals and depends on \(\Gamma_\kappa\) and its functional derivatives. The Wetterich--Morris equation arises from a
particular choice of \(F\), in which the right-hand side is written as an
integral involving the inverse operator of
\(\Gamma_\kappa^{(2)}[\phi] + R_\kappa\), and
\(\Gamma_{\kappa_0}^{\mathrm{init}}\) is given by the bare action
\(S_{\mathrm{bare}}\) at the ultraviolet scale \(\kappa_0 = \kappa_{\mathrm{UV}}\).
In FRG applications one usually takes
\(I = [\kappa_{\mathrm{IR}}, \kappa_{\mathrm{UV}}]\) and integrates the
flow from \(\kappa_{\mathrm{UV}}\) down to \(\kappa_{\mathrm{IR}}\), but the
abstract formulation \eqref{eq:generic-functional-PDE}–\eqref{eq:generic-functional-IC}
equally covers forward evolutions in \(\kappa\).

In this work we develop a GP method for solving such functional PDEs.
The basic idea is to approximate, at each scale \(\kappa\), the unknown
functional \(\Gamma_\kappa[\cdot]\) by a GP surrogate
\(\Gamma_\kappa^\dagger[\cdot]\). We first fix a finite set of collocation
fields \(\Phi = \{\phi^{(i)}\}_{i=1}^N\). For a given \(\kappa\), the values
\(\{\Gamma_\kappa[\phi^{(i)}]\}_{i=1}^N\) are regarded as observations of a GP
prior on \(\Gamma_\kappa\). Concretely, we place a GP prior on the map
\(\phi \mapsto \Gamma_\kappa[\phi]\) with a covariance kernel between fields,
and define the surrogate \(\Gamma_\kappa^\dagger\) as the corresponding GP
posterior mean. In other words, for any input field \(\phi\), 
\begin{align}
\label{eq:intro:gamma_exp}
\Gamma_\kappa^\dagger[\phi]
  := \mathbb{E}\bigl[\,\Gamma_\kappa[\phi]
      \,\big|\,
      \Gamma_\kappa[\phi^{(1)}],\dots,\Gamma_\kappa[\phi^{(N)}]\,\bigr],
\end{align}
so \(\Gamma_\kappa^\dagger\) is uniquely determined by the chosen kernel and by
the functional values on the collocation set. The functional PDE is then enforced only on the collocation set. At a fixed
scale \(\kappa\) and for each \(\phi^{(i)} \in \Phi\), we insert
\(\Gamma_\kappa^\dagger\) into the right-hand side of the evolution equation,
and compute the required functional derivatives (for example, functional
gradients and Hessians) at \(\phi^{(i)}\) through $\Gamma_\kappa^\dagger$.
Matching this right-hand side with the \(\kappa\)-derivative of
\(\Gamma_\kappa^\dagger[\phi^{(i)}]\) yields, for \(i=1,\dots,N\), a closed
system of ordinary differential equations (ODEs) in \(\kappa\) for the unknown values
\(\Gamma_\kappa[\phi^{(i)}]\) on the collocation set. Solving this finite-dimensional ODE system over the \(\kappa\)-interval of
interest determines \(\Gamma_\kappa[\phi^{(i)}]\) for all \(\kappa\) and all
collocation fields \(\phi^{(i)}\). At any given scale, the GP surrogate
\(\Gamma_\kappa^\dagger[\cdot]\) given by \eqref{eq:intro:gamma_exp} can then be evaluated on arbitrary input
fields, providing a global approximation to the solution of the functional PDE.

For clarity of exposition, in \Cref{sec: methodology} we first develop the methodology in detail for a general class of functional PDEs of the form~\eqref{eq:generic-functional-PDE}. We then illustrate the framework, and provide the corresponding implementation details, for the Wetterich--Morris equation~\eqref{eq: WM renorm}.

\subsection{Numerical methods for FRG and functional PDEs}
\label{sec: numerical methods frg}

The FRG leads to exact flow equations for
scale-dependent functionals such as the effective average action
$\Gamma_\kappa$ or the cumulant-generating functional $W_\kappa$.
Prototypical examples are the Wetterich--Morris equation~\eqref{eq: WM renorm}
and the Wilson--Polchinski equation~\eqref{eq:physWPE_intro}.  Except in
special Gaussian or mean-field limits, these functional PDEs are not
analytically solvable, and practical FRG computations rely on truncations in
functional space followed by numerical solution of the resulting PDEs
or ODE systems; see, e.g., the reviews~\cite{berges2002non,Delamotte_2012,pawlowski2007aspects,MetznerRMP2012,dupuis2021nonperturbative}.

A large class of methods begins by projecting the flow onto a finite number of
field-dependent functions or couplings.  The simplest and most widely used
example is the local potential approximation (LPA), where the flow is
restricted to a scale-dependent local potential  evaluated on
homogeneous backgrounds.  Derivative expansions and vertex expansions extend
this ansatz by including higher-derivative operators or $n$-point vertex
functions, but they still reduce the functional PDE to a coupled system of
equations in a few field or momentum variables.  Once such a truncation is
fixed, standard grid-based methods (finite differences or finite volumes) are
applied to the remaining variables.  In particular, finite-volume and
discontinuous Galerkin schemes have been designed to handle non-analytic
features in the flow without spurious
oscillations and with good stability properties~\cite{GrossiWink2023}. 

Higher-order Galerkin and finite-element methods have recently been developed
to improve accuracy and adaptivity.  In these approaches, the scale-dependent
quantities (e.g., effective potentials or vertices) are expanded in local basis
functions, and the FRG flow is enforced in a weak sense.  This allows for
mesh refinement near sharp transitions or non-analytic regions while keeping a
coarser resolution elsewhere.  The DiFfRG framework is a representative
example, providing a finite-element toolbox for a broad class of bosonic and
fermionic FRG flows, with parallel and GPU implementations for demanding
applications~\cite{sattler2024diffrg}.  Discontinuous Galerkin variants
combine the flexibility of finite elements with robust shock-capturing and
have been used, for instance, to resolve first and second order phase
transitions in scalar models with high accuracy~\cite{GrossiWink2023}.

Spectral and pseudo-spectral methods take a complementary, global viewpoint.
In these approaches the scale-dependent functions (for instance effective
potentials or fixed-point solutions) are expanded in global bases such as
Chebyshev polynomials, and the FRG flow equations are enforced at appropriate
collocation points.  The articles~\cite{BorchardtKnorr2015,BorchardtKnorr2016}
demonstrate that such pseudo-spectral schemes can produce exponentially
convergent, high-precision solutions for global fixed points and critical
exponents in scalar and Yukawa models, capturing the behavior at both small
and large fields.  Spectral methods are particularly effective when the flow
remains sufficiently smooth, while non-analyticities typically require domain
decompositions or hybrid strategies.

A common feature of these FRG approaches is that they start from a
model-specific truncation (such as the local potential approximation,
derivative expansions, or vertex expansions) and then approximate the
resulting flow equations in terms of a small number of field or momentum variables; see, for example, the reviews~\cite{berges2002non,Delamotte_2012}.
\XY{While the functional dependence $\phi \mapsto \Gamma_\kappa[\phi]$ is approximated, one works ultimately with a finite family of scalar or tensor functions defined on physical or momentum space.}

More recently, machine-learning-based FRG solvers have been proposed. In particular, physics-informed neural networks (PINNs) have been used to encode FRG flow equations in the loss functional and to approximate solutions without an explicit spatial grid, for both scalar and fermionic models~\cite{PINN-FRG}.  Such approaches can in principle handle
higher-dimensional truncations, although issues of training cost, robustness,
and a posteriori error control remain active topics of research.


By contrast, we represent the functional with a GP and impose the FRG flow equation as a structural constraint, yielding a GP-based numerical scheme for functional PDEs.

\subsection{Summary of contributions}

The main contributions of this work can be summarized as follows.

\begin{enumerate}
  \item \textbf{GP framework for functional PDEs.}
  We develop a GP method for solving functional evolution equations of the form~\eqref{eq:generic-functional-PDE}.  
  The method represents the unknown functional directly in function space, so it is independent of any particular spatial or temporal discretization.  
  Different physical representations (real or complex Fourier modes, pointwise samples, lattice fields) enter only through the choice of feature map and kernel, while the core algorithm remains unchanged.

  \item \textbf{Application to functional renormalization group flows.}
  We specialize this framework to the FRG, with a detailed formulation for the Wetterich--Morris equation~\eqref{eq: WM renorm}, and numerical experiments for both the Wetterich--Morris and the Wilson--Polchinski equations~\eqref{eq:physWPE_intro}.

  \item \textbf{Treatment of non-homogeneous fields.}
  \XY{In contrast to standard truncation schemes such as the local potential approximation, which are formulated on constant or homogeneous backgrounds, our GP method can be trained on nonconstant collocation fields when feasible, and it can then evaluate the functional on arbitrary inputs.}

  \item \textbf{Physics-informed GP priors.}
  We show how known structure from standard FRG approximations can be encoded in the GP prior, for example through local-potential-type ansatzes.
  This allows the surrogate to incorporate key physical insights while remaining fully nonparametric in function space.

   \item \textbf{Numerical validation on FRG benchmarks.}
  We test the method on several FRG flows, including Gaussian models with closed-form solutions and interacting \(\phi^4\) 
  field-theory models where no explicit solution is available.  \XY{All applications here consider fields defined on a one-dimensional input domain.}
  In the Gaussian cases, the GP surrogate reproduces the exact flow to high accuracy both on and off the collocation set.  
  For the interacting models, we consider  a lattice formulation.  \XY{In fact, among the spatial dimensions considered in \cite{zorbach2025lattice}, the approximation error of the LPA is largest in the one-dimensional case.}   Building on the benchmarks of \cite{zorbach2025lattice}, we find that our method yields solutions with smaller pointwise errors than a standard LPA flow, over most of the parameter range, for standard physical observables.
\end{enumerate}

\subsection{Notation.}
We write \(\R = (-\infty,\infty)\) for the real numbers.  
We are interested in approximating functionals (i.e., functions of functions)
that arise as solutions of integro-differential equations.  
Throughout we use the following conventions:

\begin{itemize}
  \item Capital letters \(W\) and \(\Gamma\) are reserved for 
  \emph{functionals}, i.e.\ maps from the function space \(\mathcal{U}\) to 
  the real line \(\R\).

\item For a functional \(\Gamma\), we write
  \(\delta \Gamma / \delta \phi\) for its first functional derivative and
 \(\delta^2 \Gamma / \delta \phi\,\delta \phi\) for its second functional
  derivative. The latter may be taken as formal kernel defining an integral 
  operator, which, following standard practice, we denote as \(\Gamma^{(2)}[\phi]\).  
  \item 
The evolution parameter is the RG scale and is
denoted by $\kappa\in [\kappa_{\mathrm{IR}},\kappa_{\mathrm{UV}}].$
  \item Lower-case Greek letters such as \(\phi\), \(\varphi\), and \(\psi\) are
used for real-valued functions (fields)  in the context of the
Wetterich equation and, more generally, functional PDEs. Following the
conventional notation of the Wilson--Polchinski equation in the field theory
literature, we keep \(J\) as the field variable for the Wilson--Polchinski
equation.

  \item Elements of the function space, \(\phi \in \mathcal{U}\), are referred
  to as \emph{function-space collocation functions or fields}.
\end{itemize}

We  use the following vector broadcasting convention: given a function
\(f\colon\R\to\R\) and a vector
\(\boldsymbol{\phi} = (\phi_i)_{i=1}^n\), the notation \(f(\boldsymbol{\phi})\) denotes
elementwise application, $f(\boldsymbol{\phi}) := \bigl(f(\phi_1),\dots,f(\phi_n)\bigr)$.

\subsection{Outline of the rest of the paper}
The rest of the paper is organized as follows.  
In \Cref{sec: methodology}, we present our methodology for solving general
functional PDEs with GPs and its application to the illustrative example of the
Wetterich--Morris equation. \XY{\Cref{sec: numerical results} presents our main numerical results for two FRG flows of interest, the Wilson--Polchinski equation and the Wetterich--Morris equation.
}  Several appendices fill in background information. \XY{\Cref{app: optimal recovery} reviews the connection between optimal recovery and kernel-based interpolants. We report additional numerical experiments in \Cref{app:extra_experiments}.}
In \Cref{secfre}, we provide a self-contained mathematical presentation of the
FRG flow problem.  
In \Cref{app: derivative and hessian}, we discuss functional derivatives and
Hessians of kernel interpolants.  
In \Cref{appendix:LPA_intro}, we give a brief introduction to the local
potential approximation (LPA), a standard method for obtaining approximate
solutions of the Wetterich equation on constant fields.  
In \Cref{app_sec:TM}, we review the transfer-matrix method for computing the limit solution of the Wetterich equation in one dimension.  \Cref{app:wetterich_gp_details} details the implementation of the
Wetterich equation on the lattice with \(\phi^{4}\) bare action used in the
numerical experiment of \Cref{subsec:num:wetterich_lattice}. Finally, \Cref{app_sec:mag_sus} briefly reviews the definitions of two physical observables, magnetization and susceptibility, which can be computed from the infrared local potential and used as numerical benchmarks for the Wetterich equation.
 
\section{Methodology}\label{sec: methodology}
We now outline the methodology used in this work and its application to
FRG flows. Our starting point is the kernel-based operator learning
framework of~\cite{kernelmethodsoperatorlearning}, which is designed to
learn maps between function spaces from data. In this work we adapt this
framework to represent functionals, that is, maps from function spaces
to scalars. The key distinction from data-driven operator learning
is that, in our setting, the functional values are not directly observed.
Instead, we incorporate the FRG flow equation as an additional constraint
on the surrogate, so that the reconstructed functional is informed by the
underlying functional PDE rather than by paired input-output data. We begin in \Cref{subsec:GP_kernel_interpolant} by reviewing the relationship between GP regression and kernel interpolation. \Cref{sec: kernel approx func} then discusses kernel-based approximation of functionals. In \Cref{subsec:functional-evolution}, we  develop a general framework for solving functional PDEs.
Finally, in \Cref{sec:methodology-wetterich} we specialize this framework to the Wetterich--Morris equation~\eqref{eq: WM renorm} as an illustrative example.

\subsection{Gaussian processes and kernel interpolation}
\label{subsec:GP_kernel_interpolant}
{In this subsection, we briefly review GPs and their connection with kernel interpolation. A real-valued GP is a collection of random
variables \(\{f(x)\}_{x\in\mathcal{X}}\) indexed by points of an input
space \(\mathcal{X}\), such that every finite vector
\((f(x_1),\dots,f(x_n))\) is jointly Gaussian.
A GP is characterised by its mean function
\(m(x) = \mathbb{E}[f(x)]\) and its covariance kernel
\(k(x,x') = \mathrm{Cov}(f(x),f(x'))\).
For a zero-mean prior, \(m(x)\equiv 0\), the law of the random field
\(f\) is completely determined by the positive-definite kernel \(k\).

The same kernel \(k\) defines a reproducing kernel Hilbert space (RKHS)
\(\mathcal{H}_k\).  Given exact observations \(y_i = f(x_i)\) at points
\(\{x_i\}_{i=1}^n\), the RKHS kernel interpolant
\(f^\dagger \in \mathcal{H}_k\) is the unique minimum-norm function satisfying
\[
  f^\dagger(x_i) = y_i, \qquad i=1,\dots,n,
\]
and it admits the explicit representation
\[
  f^\dagger(x) = k(x,X)\,K(X,X)^{-1}y,
\]
where \(X=(x_1,\dots,x_n)\), \(y = (y_1,\dots,y_n)^\top\), and
\(K(X,X)_{ij} = k(x_i,x_j)\).

If we place a zero-mean GP prior with covariance \(k\) on the random
field \(f\) and condition on the same noiseless data, the posterior GP
has mean
\[
  \mathbb{E}[f(x)\mid y]
  = k(x,X)\,K(X,X)^{-1}y.
\]
Thus the GP posterior mean coincides exactly with the RKHS kernel
interpolant \(f^\dagger\).
}

\subsection{Kernel approximation of functionals}
\label{sec: kernel approx func}

\begin{figure}[h]
\centerline{%
\begin{tikzpicture}[->,>=stealth',shorten >=1pt,auto,
                    thick,
                    node distance=3.8cm,
                    main node/.style={font=\sffamily\Large\bfseries,inner sep=3pt}]

\node[main node] (1) {$\mathcal{U}$};
\node[main node] (3) [below of=1,node distance=2.2cm] {$\mathbb{R}^{m}$};
\node[main node] (2) [right of=1] {$\mathbb{R}$};

\path[every node/.style={font=\sffamily\Large\bfseries},red]
      (1) edge node[above,red] {$\Gamma$} (2)
      (3) edge node[above,red] {$\gamma$} (2);

\begin{scope}[transform canvas={xshift=.3em}]
  \draw (1) to node[right] {$w$} (3);
\end{scope}

\begin{scope}[transform canvas={xshift=-.3em}]
  \draw (3) to node[left] {$\tilde{w}$} (1);
\end{scope}
\end{tikzpicture}}%
\caption{Commutative diagram for the functional learning setup.}
\label{fig:projection diagram} 
\end{figure}
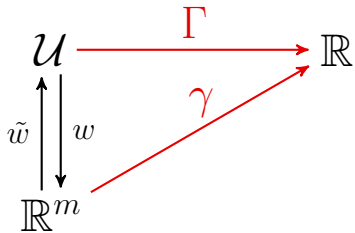

\GE{Let \(\mathcal{U}\) be a Banach space of functions and further let \(\Gamma:\mathcal{U}\to\mathbb{R}\) 
be an unknown functional that we wish to approximate.  The basic idea is to 
encode each field \(\phi\in\mathcal{U}\) by a finite vector of features \(w(\phi)\in\mathbb{R}^{m}\), 
learn a scalar function \(\gamma:\mathbb{R}^{m}\to\mathbb{R}\) on this finite-dimensional
feature space, and then approximate the functional $\Gamma(\phi)$ by the composition
\(\gamma(w(\phi))\). A commutative diagram illustrating the relationships among operators, function spaces, and functionals is shown in \Cref{fig:projection diagram}, where the reconstruction 
operator $\tilde{w}$ which ``inverts'' $w$ is discussed later in Remark~\ref{rk:pic_remark}. Concretely, when \(\mathcal{U}\) is infinite-dimensional, one does not compute directly with
\(\phi\in\mathcal{U}\), but with the finite-dimensional projections
\(w(\phi)\in\mathbb{R}^{m}\) as in~\cite{owhadi2019operator}.  
In the applications we mainly use two concrete choices. The first choice is based on pointwise measurements in space (or time):
\begin{align*}
  w(\phi) &= \bigl(\phi(x_j)\bigr)_{j=1}^{m},
\end{align*}
where \(\{x_j\}_{j=1}^{m}\) are prescribed collocation points.  This requires
the evaluation functionals \(\delta_{x_j}\) to lie in the dual space of 
\(\mathcal{U}\).  A second choice is to use spectral coefficients in an orthonormal basis 
for the case that \(\mathcal{U}\) is a Hilbert space.  For example, if \(\mathcal{U}\subset L^2([0,1])\), we
may expand
\[
  \phi(x) = \sum_{j=-\infty}^{\infty} c_j\,\mathrm{e}^{\,i2\pi j x},
\]
and define the feature vector as a truncated set of Fourier coefficients,
\[
  w(\phi) = \bigl(c_j\bigr)_{j=-m}^{m}.
\]
This construction only requires that the fields belong to \(L^2([0,1])\); other
orthonormal bases (for instance, real Fourier bases or wavelets) can be used in
the same way.}

\GE{Once a projection \(w\) is fixed, our aim is to approximate \(\Gamma\) by a
functional that depends on \(\phi\) only through the finite-dimensional
feature vector \(w(\phi)\).  Concretely, we seek a scalar function
\(\gamma:\mathbb{R}^{m}\to\mathbb{R}\) such that
\[
  \Gamma(\phi)\;\approx\;\gamma\bigl(w(\phi)\bigr)
  \qquad\text{for }\phi\in\mathcal{U}.
\]
Thus the problem of approximating the functional \(\Gamma\) is reduced to
approximating the finite-dimensional function \(\gamma\).
To construct \(\gamma\) we use optimal recovery and kernel
interpolation~\cite{owhadi2019operator}.  Suppose we are given training fields
\(\Phi=\{\phi^{(1)},\dots,\phi^{(N)}\}\subset\mathcal{U}\) and their functional
values
\[
  Y :=
  \bigl(\Gamma(\phi^{(1)}),\dots,\Gamma(\phi^{(N)})\bigr)^{\!\top}
  \in\mathbb{R}^{N}.
\]
Let \(z^{(i)}:=w(\phi^{(i)})\in\mathbb{R}^{m}\) denote the corresponding
feature vectors, and let \(\mathcal{H}_K\)
be the RKHS associated with the kernel \(K\).
We seek \(\gamma\) as the minimum-norm interpolant
\begin{align}
  \gamma^\dagger
  &= \arg\min_{\gamma\in\mathcal{H}_K}
     \|\gamma\|_{\mathcal{H}_K}^{2} \\
  &\text{subject to}\quad
     \gamma\bigl(z^{(i)}\bigr) = \Gamma(\phi^{(i)}),
     \qquad i=1,\dots,N.
\end{align}
The representer theorem (see \cite[Sec.~17.8]{owhadi2019operator}) implies that \(\gamma^\dagger\) has the explicit form
\begin{align}
  \gamma^\dagger(z)
  &= K(z,Z)^{\top}\,K(Z,Z)^{-1} Y,
  \label{eq:gamma-representer}
\end{align}
where \(Z=(z^{(1)},\dots,z^{(N)})\), \(K(z,Z)\in\mathbb{R}^{N}\) has entries
\(K(z,Z)_i = K(z,z^{(i)})\), and \(K(Z,Z)\in\mathbb{R}^{N\times N}\) is the Gram
matrix with entries \(K(Z,Z)_{ij}=K(z^{(i)},z^{(j)})\).  Our kernel surrogate
for the functional is then
\begin{align}
  \Gamma^\dagger(\phi)
  := \gamma^\dagger\bigl(w(\phi)\bigr) =
  K\bigl(w(\phi),Z\bigr)^{\top}\,
  K(Z,Z)^{-1} Y,
  \label{eq:Gamma-dagger-def}
\end{align}
which we will use in the sequel as an approximation of \(\Gamma\).}

\GE{The framework of~\cite{kernelmethodsoperatorlearning} uses this projection-%
reconstruction machinery to address data-driven operator learning, where one
is given evaluations of \(\Gamma\) at many input functions (typically from an
existing numerical solver) and no physics is imposed.  In our setting we use
the same kernel representation \eqref{eq:Gamma-dagger-def} for \(\Gamma\), but
the functional values are not directly available.  Instead, we enforce that the GP surrogate
\(\Gamma^\dagger\) satisfies the FRG flow
equations, so that the surrogate is informed by the underlying functional PDE
rather than by paired input-output data.  
Compared to neural network approaches to approximating functionals \cite{lu2019deeponet, li2020fourier, owhadi2020ideas}, the proposed approach benefits from the solid theoretical foundations of GP methods \cite{owhadi2019operator}, and can be employed to solve
integro-differential equations in functional spaces, as shown in the following section.}

\GE{
\begin{Remark}
\label{rk:pic_remark}
An important theoretical advantage of this formulation is that it may be rendered discretization-agnostic by means
of a reconstruction operator \(\tilde w:\mathbb{R}^m\to\mathcal{U}\) that maps measurement vectors back to
elements of \(\,\mathcal{U}\). See \Cref{fig:projection diagram}. 
This reconstruction operator can be defined 
using the framework of optimal recovery (see \cref{app: optimal recovery} ), which enables rigorous a priori error estimates for the recovery of $\phi\in{\mathcal U}$. Combined with the error rates associated with kernel interpolation, this can yield rigorous convergence guarantees and error rates for the recovery of $\Gamma_\kappa$. We leave this to future work.
\end{Remark}}

\subsection{A GP method for solving functional PDEs}
\label{subsec:functional-evolution}

Let \(\mathcal{U}\) be a (typically infinite-dimensional) function space and
let \(\Gamma_\kappa : \mathcal{U} \to \mathbb{R}\), \(\kappa \in I \subset \mathbb{R}\), be
a one-parameter family of functionals.
We consider the functional evolution problem
\eqref{eq:generic-functional-PDE}–\eqref{eq:generic-functional-IC}, where the
right-hand side \(F\) is built from \(\phi\), \(\Gamma_\kappa[\phi]\), and a
finite number of functional derivatives of \(\Gamma_\kappa\).
The variable \(\kappa\) plays the role of an evolution parameter and may be
advanced either forward or backward.
In FRG applications \(\kappa\) is interpreted as a momentum scale and the flow
is typically integrated backward from a UV scale \(\kappa_{\mathrm{UV}}\) to an
IR scale \(\kappa_{\mathrm{IR}}\), but the construction below is independent of
this choice of direction.

The main difficulty is that \(\Gamma_\kappa\) lives on the infinite-dimensional
space \(\mathcal{U}\).
Our strategy is to represent the evolution entirely through the values of
\(\Gamma_\kappa\) on a finite set of collocation fields and to use the GP
surrogate from \Cref{sec: kernel approx func} to extend these values to all
\(\phi\in\mathcal{U}\).
In this way the functional PDE is reduced to a closed ODE system in
finitely many unknowns.

We recall the kernel approximation of functionals from
\Cref{sec: kernel approx func}.
We fix once and for all a feature map \(w:\mathcal{U}\to\mathbb{R}^m\), a
positive-definite kernel \(K\) on \(\mathbb{R}^m\) with reproducing kernel
Hilbert space \(\mathcal{H}_K\), and a set of functional collocation fields
\[
  \Phi = \{\phi^{(1)},\dots,\phi^{(N)}\}\subset\mathcal{U}.
\]
\XY{For instance, one may take \(w_k(\phi)=\langle e_k,\phi\rangle\), where \(e_k\in\mathcal{U}\) are prescribed basis functions and \(\langle\cdot,\cdot\rangle\) denotes the inner product in the underlying Hilbert space. Other projections tailored to the problem can also be used.}
The associated feature vectors are
\[
  z^{(i)} := w(\phi^{(i)}) \in \mathbb{R}^m,
  \qquad
  Z := (z^{(1)},\dots,z^{(N)}),
\]
and we write \(K(Z,Z)\in\mathbb{R}^{N\times N}\) for the Gram matrix introduced
in \Cref{sec: kernel approx func}.
Given a vector \(Y\in\mathbb{R}^N\) of functional values on \(\Phi\), the
kernel construction \eqref{eq:Gamma-dagger-def} defines a GP surrogate
\(\Gamma^\dagger\). 
In numerical computations we replace \(K(Z,Z)^{-1}\) by
\(\bigl(K(Z,Z)+\varepsilon I\bigr)^{-1}\) for a small \(\varepsilon>0\) for
stability, but the representer form itself is unchanged.

In the evolutionary setting the role of \(Y\) is played by the 
values of the scale-dependent functional on the collocation set.
For each \(\kappa\in I\) we introduce the vector
\begin{align}
\label{eq:def_gk}
Y_\kappa
  :=
  \bigl(\Gamma_\kappa[\phi^{(1)}],\dots,
        \Gamma_\kappa[\phi^{(N)}]\bigr)^{\top}
  \in\mathbb{R}^N,
\end{align}
which is the primary unknown.
At the initial scale \(\kappa_0\in I\) these values are prescribed by the
initial functional \(\Gamma_{\kappa_0}^{\mathrm{init}}\),
\[
  Y_{\kappa_0}
  = \Gamma_{\kappa_0}^{\mathrm{init}}(\Phi)
  =
  \bigl(\Gamma_{\kappa_0}^{\mathrm{init}}[\phi^{(1)}],\dots,
        \Gamma_{\kappa_0}^{\mathrm{init}}[\phi^{(N)}]\bigr)^{\top},
\]
which will serve as the initial condition for the finite–dimensional evolution.

Given \(Y_\kappa\), we approximate \(\Gamma_\kappa[\cdot]\) at scale \(\kappa\)
by the GP surrogate defined, as in \eqref{eq:Gamma-dagger-def}, with
\(Y = Y_\kappa\):
\begin{equation}
  \Gamma_\kappa^\dagger[\phi]
  :=
  K\bigl(w(\phi),Z\bigr)\,
  \bigl(K(Z,Z)+\varepsilon I\bigr)^{-1} Y_\kappa,
  \qquad \phi\in\mathcal{U}.
  \label{eq:Gamma-surrogate-kappa}
\end{equation} 
Thus \(\Gamma_\kappa^\dagger\) is defined for all input fields and depends on
\(\kappa\)  through the finite-dimensional vector \(Y_\kappa\).

To obtain a closed evolution for \(Y_\kappa\), we enforce the functional PDE
\eqref{eq:generic-functional-PDE} only on the collocation set \(\Phi\).
For a fixed \(\kappa\in I\) and each \(\phi^{(i)}\in\Phi\), we approximate the
right-hand side of \eqref{eq:generic-functional-PDE} by inserting the surrogate
\eqref{eq:Gamma-surrogate-kappa}:
\begin{equation}
\label{eq:pde-surrogate}
  \partial_\kappa \Gamma_\kappa[\phi^{(i)}]
  \;\approx\;
  F\!\left(
    \kappa,\,
    \phi^{(i)},\,
    \Gamma_\kappa^\dagger[\phi^{(i)}],\,
    \frac{\delta \Gamma_\kappa^\dagger}{\delta\phi}\Big|_{\phi^{(i)}},\,
    \frac{\delta^2 \Gamma_\kappa^\dagger}{\delta\phi^2}\Big|_{\phi^{(i)}}
  \right),
  \qquad i=1,\dots,N,
\end{equation}
where the functional derivatives of \(\Gamma_\kappa^\dagger\) are computed
analytically from the kernel representation; see
\Cref{app: derivative and hessian} for details.
By definition \(Y_\kappa^{(i)}=\Gamma_\kappa[\phi^{(i)}]\), so the left-hand
side of \eqref{eq:pde-surrogate} is exactly
\[
  \partial_\kappa \Gamma_\kappa[\phi^{(i)}]
  = \frac{\mathrm{d}}{\mathrm{d}\kappa}\,Y_\kappa^{(i)}.
\]
Hence \eqref{eq:pde-surrogate} naturally induces a closed system of ODEs for the vector \(Y_\kappa\):
\begin{equation}
  \frac{\mathrm{d}}{\mathrm{d}\kappa} Y_\kappa^{(i)}
  =
  F\!\left(
    \kappa,\,
    \phi^{(i)},\,
    \Gamma_\kappa^\dagger[\phi^{(i)}],\,
    \frac{\delta \Gamma_\kappa^\dagger}{\delta\phi}\Big|_{\phi^{(i)}},\,
    \frac{\delta^2 \Gamma_\kappa^\dagger}{\delta\phi^2}\Big|_{\phi^{(i)}}
  \right),
  \qquad i=1,\dots,N,
  \label{eq:finite-dim-evolution-compact}
\end{equation}
with initial condition
\[
  Y_{\kappa_0}
  = \Gamma_{\kappa_0}^{\mathrm{init}}(\Phi).
\]

Equation~\eqref{eq:finite-dim-evolution-compact} is a standard ODE system on
\(\mathbb{R}^N\).
In FRG applications one typically takes
\(I = [\kappa_{\mathrm{IR}},\kappa_{\mathrm{UV}}]\) and integrates this ODE
backward from \(\kappa_{\mathrm{UV}}\) to \(\kappa_{\mathrm{IR}}\), but the same
formulation applies equally to forward evolutions in \(\kappa\).
Once \(Y_\kappa\) has been computed on the interval of interest, the surrogate
\eqref{eq:Gamma-surrogate-kappa} provides an approximation of
\(\Gamma_\kappa[\phi]\) for arbitrary fields \(\phi\in\mathcal{U}\).

\subsection{A functional RG example: the Wetterich--Morris equation}
\label{sec:methodology-wetterich}
\XY{\Cref{subsec:functional-evolution} provides a framework for solving general functional PDEs. To illustrate how this framework applies to FRG flows,} we now specialize it to the
Wetterich--Morris equation~\eqref{eq: WM renorm}, for fields that are
either periodic on the one-dimensional torus \(\mathbb{T}\), identified with
the interval \([0,1]\), or
defined on a lattice.  Recall that the scale parameter \(\kappa\) plays the
role of a sliding momentum cutoff.  At a large ultraviolet scale
\(\kappa_{\mathrm{UV}}\) the functional \(\Gamma_{\kappa_{\mathrm{UV}}}\) is
prescribed by a microscopic action \(S_{\mathrm{bare}}\) that encodes the model
at short distances.  We integrate \eqref{eq: WM renorm} backward in \(\kappa\) from
\(\kappa_{\mathrm{UV}}\) down to \(\kappa_{\mathrm{IR}}\). As \(\kappa\) is lowered towards an infrared scale
\(\kappa_{\mathrm{IR}}>0\), more and more high-wavenumber modes are integrated out and
\(\Gamma_\kappa\) evolves towards the full effective action.  

\medskip
\noindent\textbf{Finite-dimensional representation of fields.}
In the GP method of \Cref{subsec:functional-evolution} the first step is to
fix a feature map \(w:\mathcal{U}\to\mathbb{R}^m\) that encodes each field by a
finite vector.
We now specify this map for the Wetterich--Morris equation.

We work with a function space \(\mathcal{U}\) of fields  and view
\(\mathcal{U}\) as continuously embedded in a Hilbert space \(\mathcal{X}\) that
carries the natural inner product of the problem.
For continuum examples in this work, \(\mathcal{X}\) is the periodic
\(L^2\)-space on the one-dimensional torus \(\mathbb{T}\) or the corresponding
Sobolev space \(H^1(\mathbb{T})\).
For lattice examples, \(\mathcal{X}\) is the discrete \(L^2\)-space of grid
functions.

It is convenient to represent fields in an orthonormal basis of Laplacian eigenfunctions, so that the Laplacian and standard translation-invariant regulators act diagonally in this basis.
We therefore fix a real orthonormal basis \(\{e_\alpha\}_{\alpha\in\mathcal{A}}\)
of \(\mathcal{X}\) consisting of eigenfunctions of the  continuum Laplacian  or
the discrete Laplacian on the lattice. 
On the torus \(\mathbb{T}\), a standard choice is the complex Fourier basis
\(\{\mathrm{e}^{2\pi i p x}\}_{p\in\mathbb{Z}}\), or, equivalently, the
associated real Fourier basis
\[
  e_0(x)=1,\qquad
  e_{2p-1}(x)=\sqrt{2}\cos(2\pi p x),\qquad
  e_{2p}(x)=\sqrt{2}\sin(2\pi p x),
  \quad p\ge 1.
\]
On a periodic lattice we use the corresponding discrete Fourier modes.

For numerical purposes we truncate to a finite index set
\(\mathcal{A}_P \subset \mathcal{A}\) of active modes and work on the
finite-dimensional subspace $
  \mathcal{X}_P
  := \operatorname{span}\{e_\alpha : \alpha \in \mathcal{A}_P\}$. 
Here \(P\) is a truncation parameter that controls how many Fourier modes (or,
more generally, eigenmodes) are retained: increasing \(P\) increases both the
dimension of \(\mathcal{X}_P\) and the highest resolved frequency. The truncation level \(P\) is chosen so that
\(\{e_\alpha\}_{\alpha\in\mathcal{A}_P}\) captures all physically relevant
modes while keeping the associated matrices of moderate size. \GE{Such wavenumber truncation is
standard in physics practice as well, e.g. in the case of a spatial lattice model with spacing $a$ naturally 
$P=\pi/a.$}

Given this basis, we define the feature map \(w\) by the truncated expansion
coefficients
\[
  c_\alpha[\phi] := \langle \phi,e_\alpha\rangle_{\mathcal{X}},
  \qquad \alpha\in\mathcal{A}_P,
\]
and set
\[
  w(\phi)
  := \bigl(c_\alpha[\phi]\bigr)_{\alpha\in\mathcal{A}_P}
  \in \mathbb{R}^{m},
  \qquad m := |\mathcal{A}_P|.
\]
On a lattice the inner product \(\langle\cdot,\cdot\rangle_{\mathcal{X}}\) is
understood as the corresponding discrete \(L^2\) inner product.
With this choice, all subsequent GP surrogates \(\Gamma_\kappa^\dagger\) are
constructed from the finite-dimensional feature vectors \(w(\phi)\) as in
\Cref{sec: kernel approx func}.

\medskip
\noindent\textbf{GP formulation of the Wetterich--Morris flow.}
We now specialise the general GP framework of \Cref{subsec:functional-evolution}
to the Wetterich--Morris equation~\eqref{eq: WM renorm}.

On the truncated mode set \(\mathcal{A}_P\) we represent each field
\(\phi\in\mathcal{U}\) by its coefficients in the eigenbasis discussed above,
and use these coefficients as features:
\[
  w(\phi)
  :=
  \bigl(c_\alpha[\phi]\bigr)_{\alpha\in\mathcal{A}_P},
  \qquad
  c_\alpha[\phi]
  = \langle \phi, e_\alpha\rangle_{\mathcal{X}}.
\]
We then fix a finite set of collocation fields
\[
  \Phi = \{\phi^{(1)},\dots,\phi^{(N)}\}\subset\mathcal{U},
  \qquad
  z^{(i)} := w(\phi^{(i)})\in\mathbb{R}^m,
\]
and collect the feature vectors into \(Z=(z^{(1)},\dots,z^{(N)})\).
A covariance kernel \(K\) on feature space is fixed once and for all, and its
Gram matrix \(K(Z,Z)\) enters the GP surrogate exactly as in
\eqref{eq:Gamma-surrogate-kappa}.

For each scale \(\kappa\), the unknown values on the collocation set are
collected into the vector \(Y_\kappa\) defined in~\eqref{eq:def_gk}, which in
turn determines the GP surrogate \(\Gamma_\kappa^\dagger[\cdot]\) through
\eqref{eq:Gamma-surrogate-kappa}.
At the ultraviolet scale \(\kappa_{\mathrm{UV}}\) the initial condition is
given by the bare action,
\[
  Y_{\kappa_{\mathrm{UV}}}
  =
  \Gamma_{\kappa_{\mathrm{UV}}}(\Phi)
  =
  \bigl(S_{\mathrm{bare}}[\phi^{(1)}],\dots,
        S_{\mathrm{bare}}[\phi^{(N)}]\bigr)^\top,
\]
which plays the role of \(\Gamma_{\kappa_0}^{\mathrm{init}}(\Phi)\) in
\eqref{eq:finite-dim-evolution-compact}.
The evolution of \(Y_\kappa\) is governed by the finite-dimensional ODE system 
\eqref{eq:finite-dim-evolution-compact}, whose right-hand side is obtained by
evaluating the Wetterich--Morris flow with the surrogate
\(\Gamma_\kappa^\dagger\).
In this setting, we obtain
\begin{equation}
  \frac{\mathrm{d}}{\mathrm{d}\kappa} Y_\kappa^{(i)}
  = \frac{1}{2}\int_{x,y} \partial_\kappa R_\kappa(x,y)\,
     \Bigl( \Gamma_\kappa^{\dagger(2)}[\phi^{(i)}]
            + R_\kappa \Bigr)^{-1}(x,y)\,\diff{x}\,\diff{y},
  \qquad i=1,\dots,N,
  \label{eq:finite-dim-evolution:wetterich}
\end{equation}
where the second functional derivative \GE{operator} \(\Gamma_\kappa^{\dagger(2)}\) is
computed from the surrogate \(\Gamma_\kappa^\dagger\).

\medskip
\noindent\textbf{A computable ODE system for the Wetterich--Morris equation.}
Next, our goal is to rewrite the right-hand side of
\eqref{eq:finite-dim-evolution:wetterich} in an explicit, numerically tractable
form.

In our GP framework all information about fields enters through the feature map
\(w\), which only retains their projection onto the truncated subspace
\(\mathcal{X}_P = \operatorname{span}\{e_\alpha : \alpha\in\mathcal{A}_P\}\).
It is therefore natural to evaluate the  integral in 
\eqref{eq:finite-dim-evolution:wetterich} on the same finite-dimensional subspace.
Concretely, we replace \(\Gamma_\kappa^{\dagger (2)}[\phi]\) and \(R_\kappa\) by their
restrictions to \(\mathcal{X}_P\), represented in the basis
\(\{e_\alpha\}_{\alpha\in\mathcal{A}_P}\).
This approximation is consistent with the  truncation implicit
in \(w\) and reduces the  integration to a finite sum over the  modes in \(\mathcal{A}_P\). 

For a fixed field \(\phi\in\mathcal{U}\) and scale \(\kappa\), the second
functional derivative \GE{defines a linear operator} \(\Gamma_\kappa^{\dagger (2)}[\phi]\) on
\(\mathcal{X}\). 
Restricted to the truncated subspace \(\mathcal{X}_P\), we represent the
operator \(\Gamma_\kappa^{\dagger (2)}[\phi]\) by the matrix
\(H_\kappa(\phi)\) with entries
\[
  H_\kappa(\phi)_{\alpha\beta}
  :=
  \bigl\langle e_\alpha,\,
          \Gamma_\kappa^{\dagger (2)}[\phi]\,e_\beta
  \bigr\rangle_{\mathcal{X}},
  \qquad \alpha,\beta\in\mathcal{A}_P.
\]
The matrix elements \(H_\kappa(\phi)_{\alpha\beta}\) are computed from the
surrogate \(\Gamma_\kappa^\dagger\) by taking directional second derivatives
along \(e_\alpha\) and \(e_\beta\); see \Cref{app: derivative and hessian} for
the computations.

In line with standard FRG practice \GE{for field theory and critical phenomena}, we restrict attention to
translation-invariant regulators (see, e.g.,~\cite{berges2002non,Delamotte_2012}).
In the Laplacian eigenbasis \(\{e_\alpha\}_{\alpha\in\mathcal{A}}\), with
\(-\Delta e_\alpha = \lambda_\alpha e_\alpha\), such regulators are diagonal and admit
the spectral representation
\[
  R_\kappa(x,y)
  \;\approx\;
  \sum_{\alpha\in\mathcal{A}_P}
    r_\kappa(\lambda_\alpha)\,e_\alpha(x)e_\alpha(y),
\]
for some scalar profile \(\lambda\mapsto r_\kappa(\lambda)\), where the sum is
restricted to the truncated mode set \(\mathcal{A}_P\).
Its scale derivative has the analogous expansion
\[
  \partial_\kappa R_\kappa(x,y)
  \;\approx\;
  \sum_{\alpha\in\mathcal{A}_P}
    \partial_\kappa r_\kappa(\lambda_\alpha)\,e_\alpha(x)e_\alpha(y).
\]

On the truncated subspace \(\mathcal{X}_P\) we represent
\(\Gamma_\kappa^{\dagger (2)}[\phi] + R_\kappa\) by the matrix \(A_\kappa(\phi)\) with
entries
\[
  A_\kappa(\phi)_{\alpha\beta}
  :=
  H_\kappa(\phi)_{\alpha\beta}
  + r_\kappa(\lambda_\alpha)\,\delta_{\alpha\beta},
  \qquad \alpha,\beta\in\mathcal{A}_P,
\]
and we denote its inverse by
\[
  B_\kappa(\phi) := A_\kappa(\phi)^{-1}.
\]
In this basis, the inverse kernel
\[
  \mathcal{G}_\kappa[\phi](x,y)
  := \bigl(\Gamma_\kappa^{\dagger (2)}[\phi]+R_\kappa\bigr)^{-1}(x,y)
\]
is then approximated by
\[
  \mathcal{G}_\kappa[\phi](x,y)
  \;\approx\;
  \sum_{\beta,\ell\in\mathcal{A}_P}
    B_\kappa(\phi)_{\beta\ell}\,e_\beta(x)\,e_\ell(y).
\]

Using orthonormality in \(\mathcal{X}\),
\(\int e_\alpha(x)e_\beta(x)\,\mathrm{d}x = \delta_{\alpha\beta}\) (and
similarly in \(y\)), the  approximation of
\eqref{eq:finite-dim-evolution:wetterich} on \(\mathcal{X}_P\) takes the form
\begin{equation}\label{eq:wetterich-ODE-g-final}
  \frac{\mathrm{d}}{\mathrm{d}\kappa} Y_\kappa^{(i)}
  =
  \frac{1}{2}
  \sum_{\alpha\in\mathcal{A}_P}
    \partial_\kappa r_\kappa(\lambda_\alpha)\,
    B_\kappa\bigl(\phi^{(i)}\bigr)_{\alpha\alpha},
  \qquad i=1,\dots,N.
\end{equation}
This ODE system is the numerical approximation of the flow
\eqref{eq:finite-dim-evolution:wetterich} used in our computations.
Together with the initial condition
\(
  Y_{\kappa_{\mathrm{UV}}}
  = \Gamma_{\kappa_{\mathrm{UV}}}(\Phi)
\),
this defines a closed ODE system for the collocated values \(Y_\kappa\).
We integrate \eqref{eq:wetterich-ODE-g-final} backward in \(\kappa\), from
\(\kappa_{\mathrm{UV}}\) down to \(\kappa_{\mathrm{IR}}\), using a standard ODE
solver.
At any intermediate scale, the surrogate \(\Gamma_\kappa^\dagger[\cdot]\) for
arbitrary \(\phi\in\mathcal{U}\) is recovered by inserting the current
\(Y_\kappa\) into \eqref{eq:Gamma-surrogate-kappa}.

\section{Numerical results}\label{sec: numerical results}

We evaluate the proposed GP solver on FRG flows. The first example is an  explicitly solvable benchmark: the exact quadratic solution of the Wilson--Polchinski equation for the Gaussian model (Subsection~\ref{subsec:num:wp_gaussian}). This case is used to validate correctness under controlled conditions. Here, we vary the number and type of collocation functions, report errors both at the collocation set and on independently drawn test functions from a different distribution, and study convergence as the number of collocation functions increases.

The second experiment is carried out for the Wetterich equation on a periodic lattice. As documented in \cite{zorbach2025lattice}, the results produced by the LPA when restricted to constant fields can differ from those obtained by lattice Monte Carlo (MC) \cite{zorbach2025lattice} for certain choices of bare parameters. 
In one spatial dimension, both lattice Monte Carlo (MC)~\cite{zorbach2025lattice} and transfer matrix (TM) \cite{baxter2016exactly} methods provide highly accurate reference solutions for some physical observables  of the system.
We therefore benchmark our method against LPA, lattice MC, and TM. 
  Our primary accuracy metrics are \GE{three} physical observables: \GE{the effective potential,} the magnetization and the susceptibility (see Appendix \ref{app_sec:mag_sus}). The magnetization is the field value that minimizes the infrared effective potential once a homogeneous source has tilted it by a linear term in the field, and it captures the system’s equilibrium response to that source. The susceptibility measures how rapidly this equilibrium magnetization changes as the source varies, serving as the local slope of the magnetization curve and characterizing the strength of the system’s linear response. In Subsection~\ref{subsec:num:wetterich_lattice}, we apply our GP method to solve the Wetterich equation. \XY{Here the GP prior differs from the one in the first case: it is derived from the LPA ansatz, which consists of a kinetic term and a local potential term. We model the local potential as a GP, and by the linearity of the ansatz in this potential, the full functional inherits a GP prior. This construction readily extends to other settings where a physical ansatz is available and can be encoded into the structure of the GP prior.}
 Across the tests, our method produces magnetization and susceptibility curves that agree with the TM and MC references more closely than LPA over most of the sampled range, demonstrating systematic improvements.

Throughout the experiments, we study FRG flows on fixed finite spatial domains.  For a discussion of the differences between finite-volume and infinite-volume formulations of the FRG, we refer the reader to~\cite{zorbach2025lattice}.

All experiments are implemented in Python with JAX~\cite{jax}. The resulting ODE systems are integrated using the LSODA algorithm to automatically handle both nonstiff and stiff regimes. Detailed solver settings for each problem, together with scripts to reproduce all experiments, are available in the public repository%
\footnote{\href{https://github.com/yangx0e/KernelsRenormalizationGroups}%
{\texttt{https://github.com/yangx0e/KernelsRenormalizationGroups}}}%
\textsuperscript{,}%
\footnote{\XY{All calculations are performed on a MacBook Pro equipped with an Apple M1 Max processor and 64\,GB of memory.}}.

\subsection{The Wilson--Polchinski equation: Gaussian-model}
\label{subsec:num:wp_gaussian}
We consider the Wilson--Polchinski equation
\begin{equation}\label{eq:physWPE}
\partial_\kappa W_\kappa[J]
=\frac{1}{2}\int_{0}^1 \int_{0}^{1} \partial_\kappa R_\kappa(x,y)
\left(\frac{\delta^2 W_\kappa[J]}{\delta J_x \delta J_y}
      +\frac{\delta W_\kappa[J]}{\delta J_x}\frac{\delta W_\kappa[J]}{\delta J_y}\right)\,\mathrm{d}x\,\mathrm{d}y,
\end{equation}
which admits the following explicit solution for a Gaussian free-field theory
\begin{align}
\label{eq:wp:explicit}
W_\kappa[J]
= -\frac{1}{2}\int_{0}^{1}\!\!\int_{0}^{1}
J(x)\,[\Gamma_2+R_\kappa]^{-1}(x,y)\,J(y)\,\mathrm{d}x\,\mathrm{d}y
-\frac{1}{2}\ln \frac{{\det}(\Gamma_2+R_\kappa)}{\det \Gamma_2},
\end{align}
with \(\Gamma_2=-\Delta+1\). Here, the initial data for \eqref{eq:physWPE} at the UV scale $\kappa_{\mathrm{UV}}$ is prescribed by the explicit expression \eqref{eq:wp:explicit} evaluated at $\kappa = \kappa_{\mathrm{UV}}$. We work on the periodic Sobolev space
\[
  H^{1}_{\mathrm{per}}([0,1];\R)
  = \bigl\{J\in H^{1}([0,1];\R)\colon J(0)=J(1)\bigr\},
\]
and choose the Litim regulator \cite{litim2001optimized}
\[
R_{\kappa}(p)
=\bigl(\kappa^{2}-p^{2}\bigr)\,\Theta(\kappa^{2}-p^{2}),
\]
which adds a mass-like term \(\kappa^{2}-p^{2}\) for Fourier modes with \(p^{2}<\kappa^{2}\) and vanishes otherwise. Here, we recall that \(\Theta\) denotes the Heaviside step function, equal to \(1\) for positive arguments and \(0\) for nonpositive arguments.


For computations we truncate to the \((2P+1)\)-dimensional  Fourier subspace
\[
  V_{P}
  = \Bigl\{
      J(x)=\sum_{p=-P}^{P}c_{p}\,e^{2\pi i p x}
      \;\colon\;
      c_{p}\in\R,\;c_{p}=c_{-p}
    \Bigr\}
  \;\subset\;H^{1}_{\mathrm{per}}([0,1];\R).
\]

\subsubsection{Numerical setup}
\begin{table}[ht]
  \centering
  \caption{Gaussian-Model Solution of the Wilson--Polchinski Equation: Average relative $L^2$ errors on collocation and test sets for varying sample sizes $N$.}
  \label{tab:avg-rel-l2-errors-horizontal}
  \begin{tabular}{|c|c|c|c|c|c|}
    \hline
                  & $N=50$  & $N=100$ & $N=200$ & $N=500$ & $N=1000$ \\ 
    \hline
    Collocation set 
                  & $1.91 \times 10^{-8}$ & $2.51\times 10^{-9}$  & $1.28\times 10^{-9}$  & $8.29 \times 10^{-10}$  & $8.29 \times 10^{-10}$   \\ 
    \hline
    Test set    
                  & $5.10 \times 10^{-8}$  & $3.99\times 10^{-9}$  & $1.46\times 10^{-9}$  & $7.48\times 10^{-10}$  & $6.59\times 10^{-10}$   \\ 
    \hline
  \end{tabular}
\label{tb:wp}
\end{table}

\begin{figure}[!hbtp]
	\centering  
    \begin{subfigure}[b]{0.32\textwidth}		\adjustbox{valign=b}{\includegraphics[width=\textwidth]{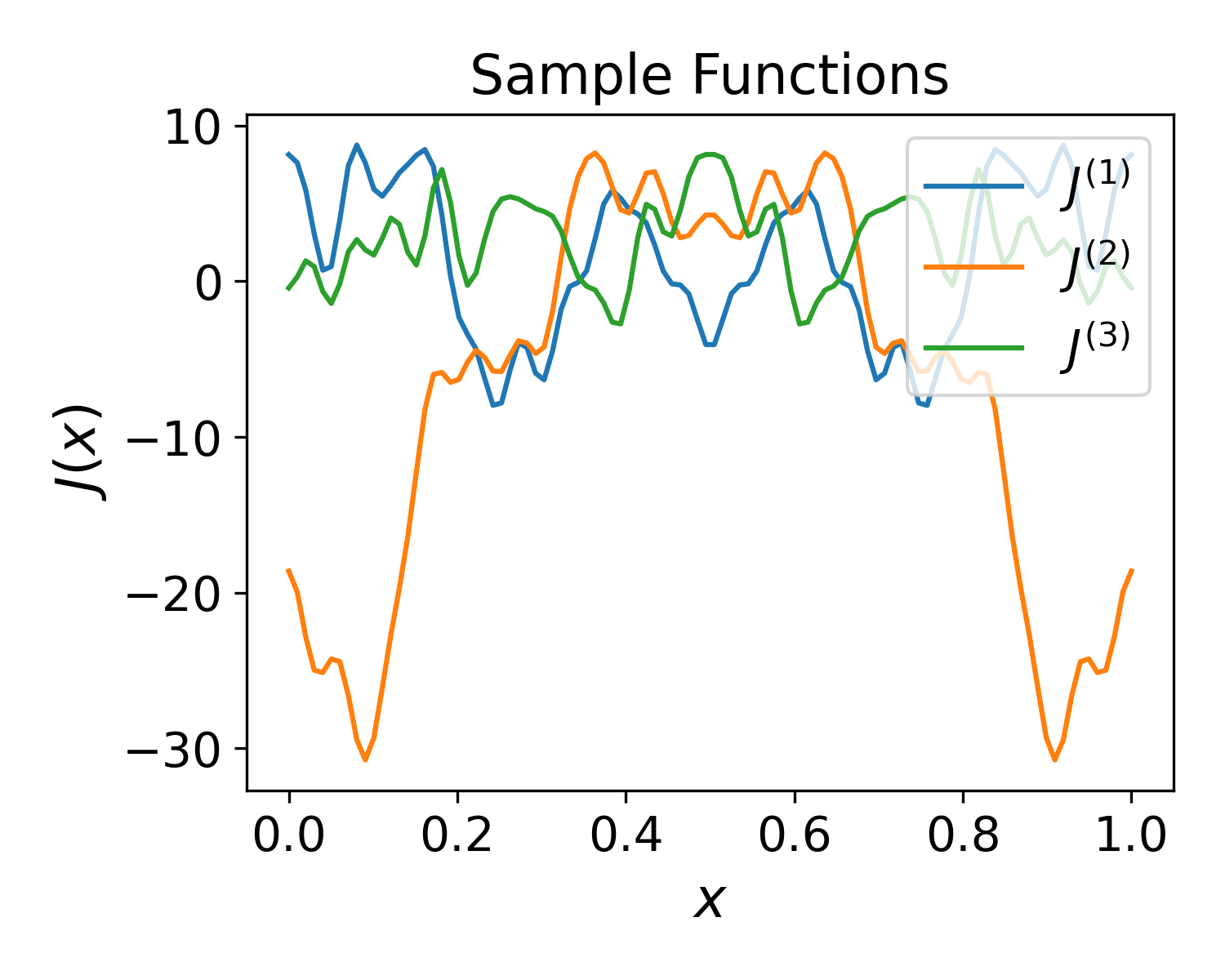}}
		\caption{The first three of 1000 collocation points.}
		\label{fig:Wilsonpolchinsky:collocation}
	\end{subfigure} 
	\begin{subfigure}[b]{0.32\textwidth}
		\adjustbox{valign=b}{\includegraphics[width=\textwidth]{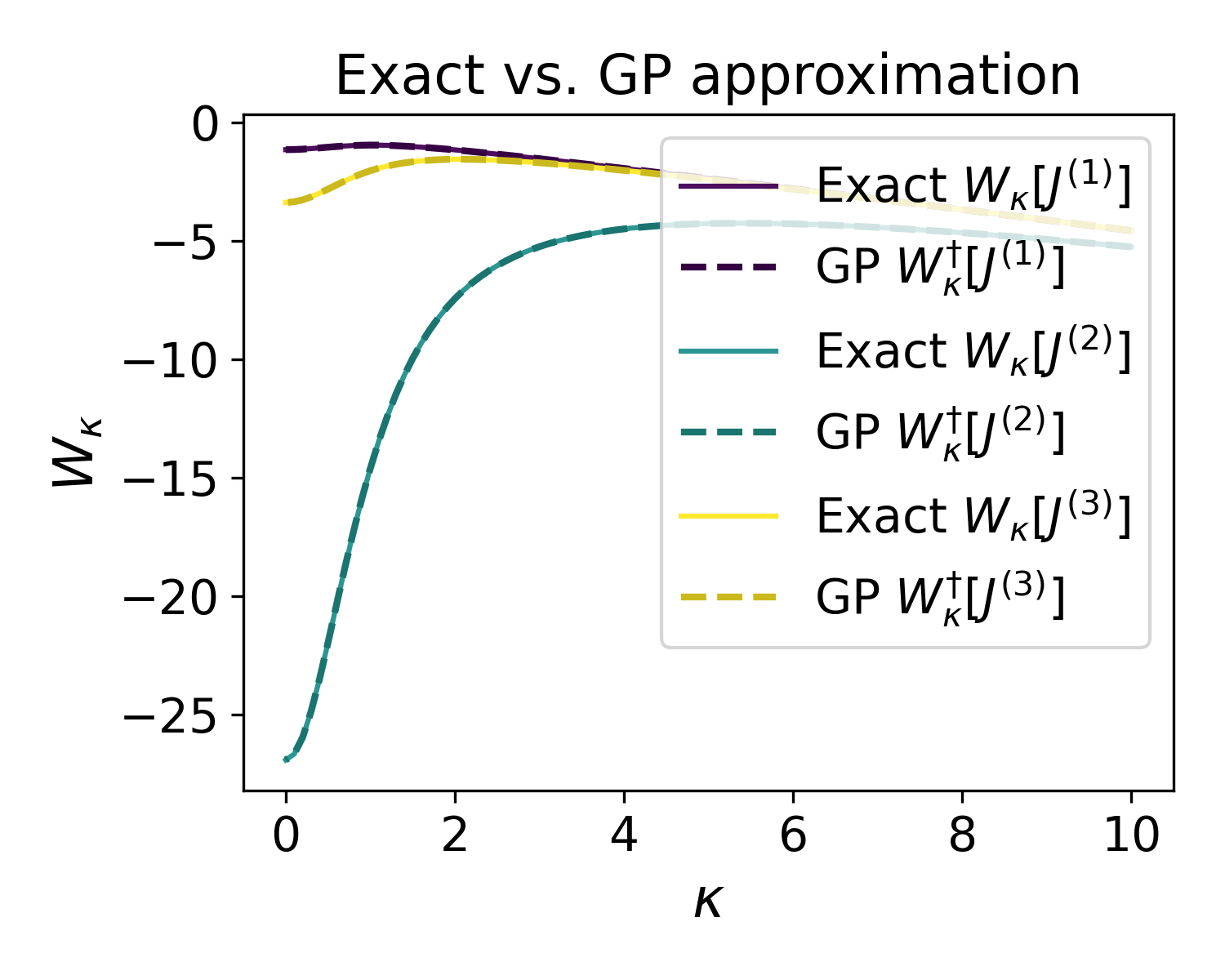}}
		\caption{Exact solutions vs. GP approximations.}
		\label{fig:Wilsonpolchinsky:collocation_fields}
	\end{subfigure} 
    \begin{subfigure}[b]{0.32\textwidth}
		\adjustbox{valign=b}{\includegraphics[width=\textwidth]{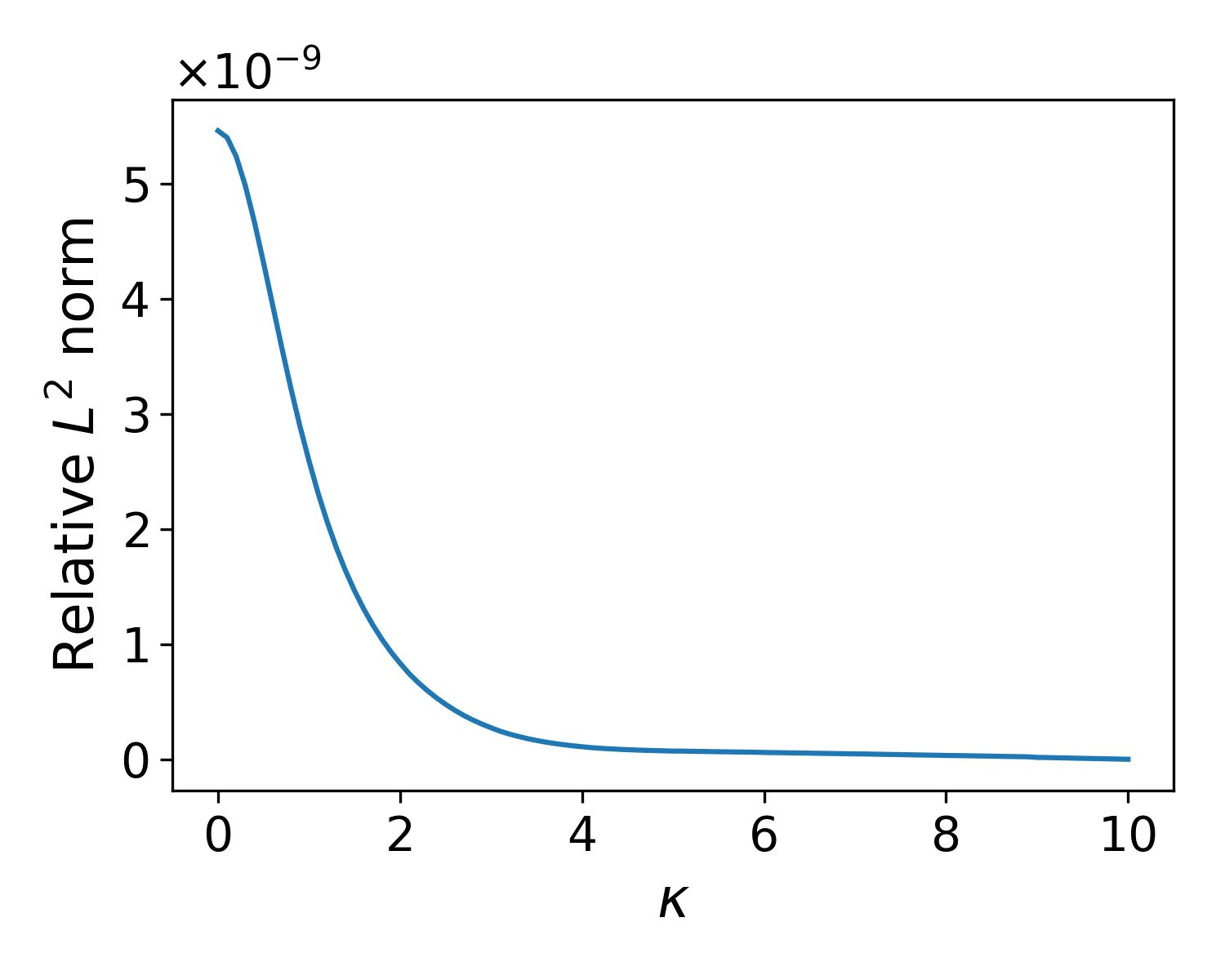}}
		\caption{Relative $L^2$ error per scale variable $\kappa$ at collocation points.}
		\label{fig::Wilsonpolchinsky:collocation_fields:RL2}
	\end{subfigure} \bigskip
    \begin{subfigure}[b]{0.32\textwidth}
		\includegraphics[width=\textwidth]{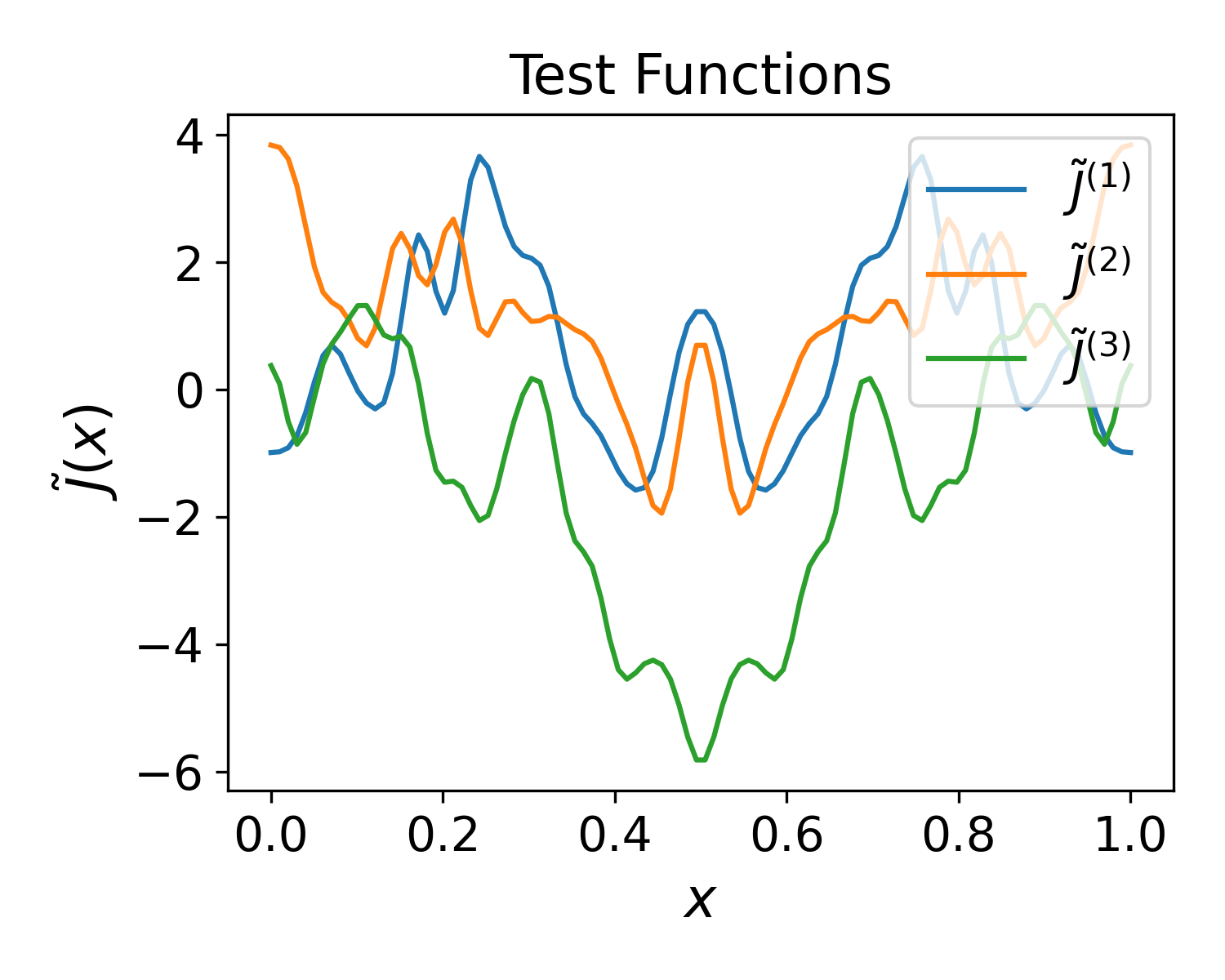}
		\caption{The first three of 200 test points.}
		\label{fig:Wilsonpolchinsky:test_fields}
	\end{subfigure}
	\begin{subfigure}[b]{0.32\textwidth}
		\includegraphics[width=\textwidth]{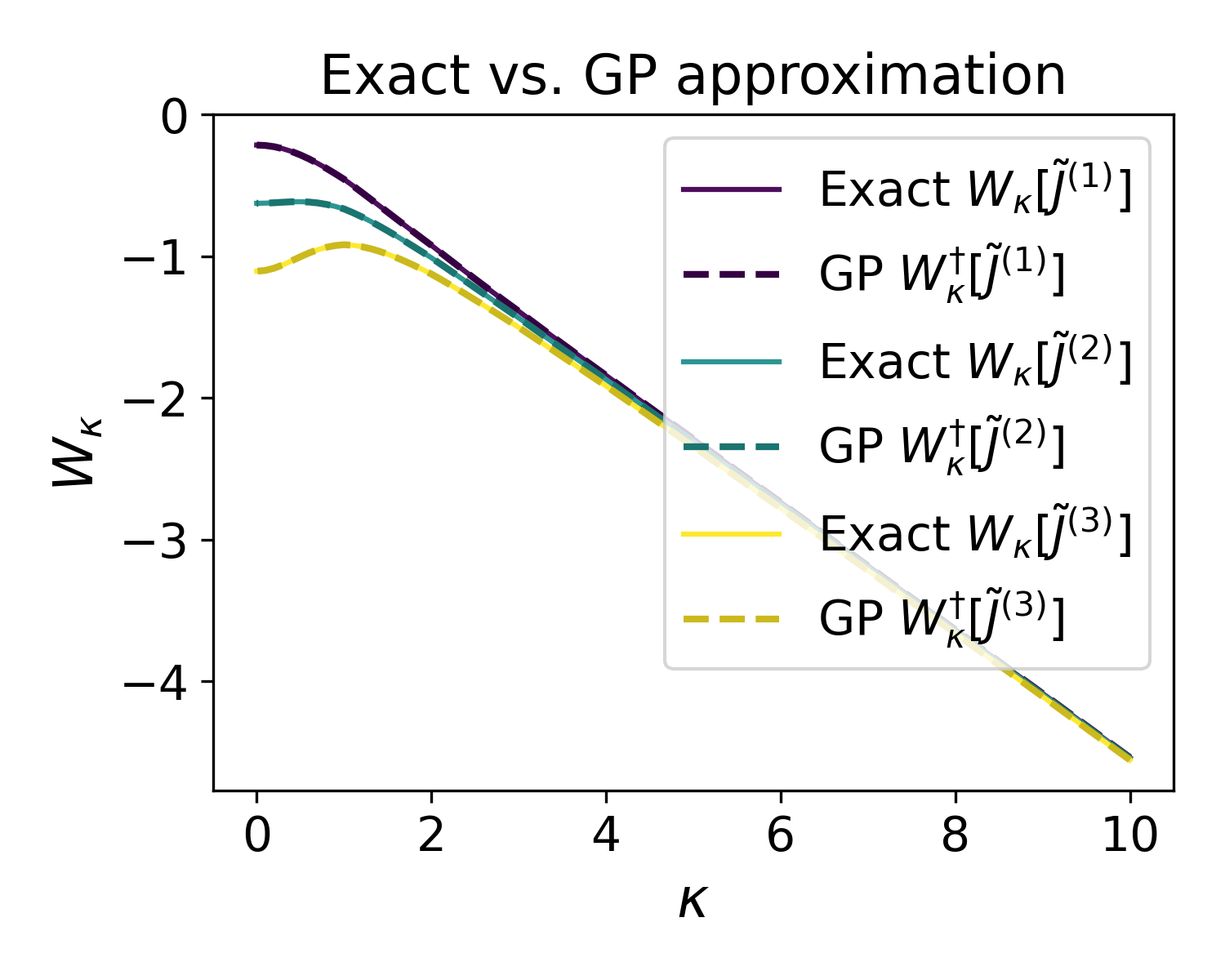}
		\caption{Exact solutions vs. GP approximations.}
		\label{fig:Wilsonpolchinsky:test_fields}
	\end{subfigure}
	\begin{subfigure}[b]{0.32\textwidth}
		\includegraphics[width=\textwidth]{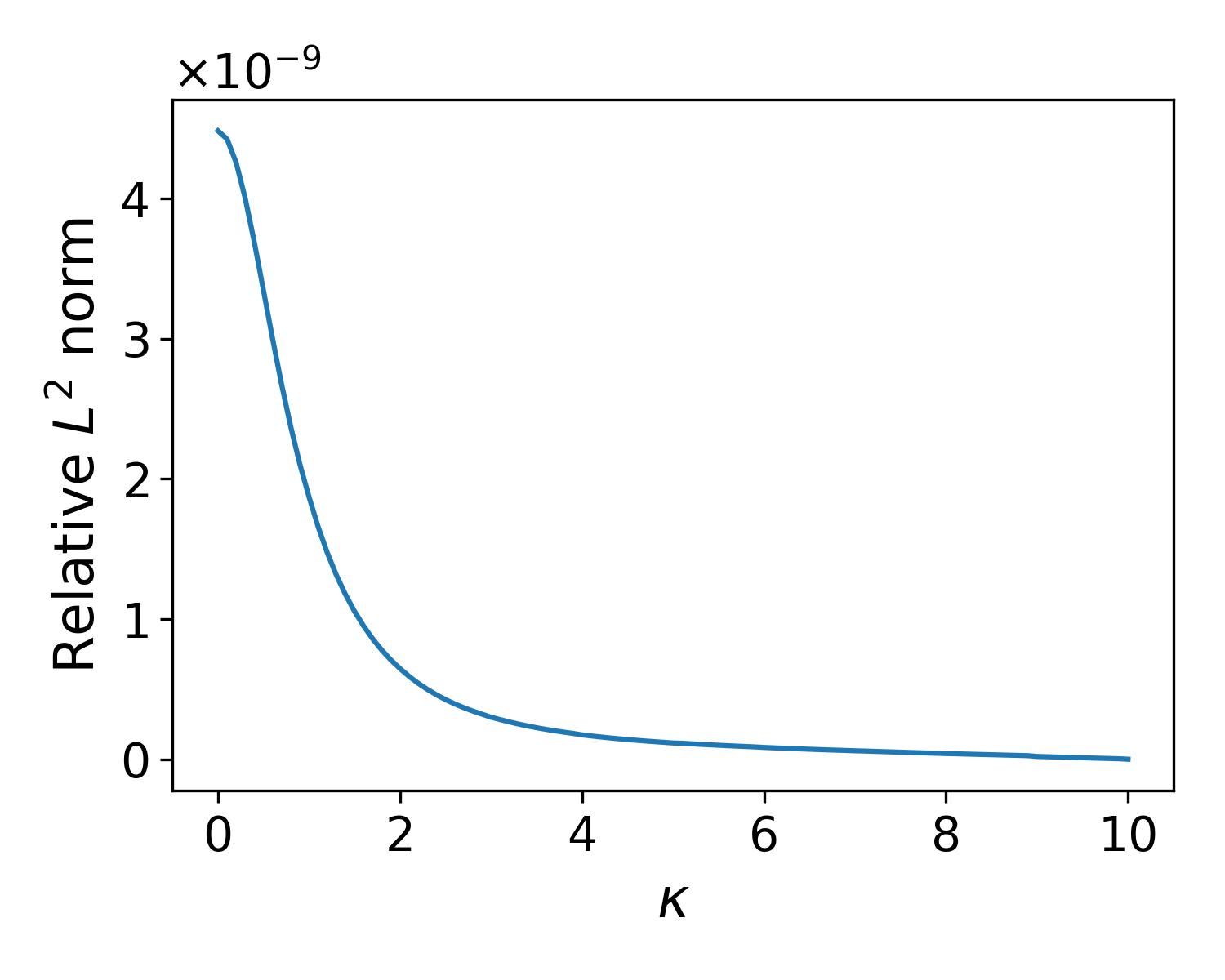}
		\caption{Relative $L^2$ error per scale variable $\kappa$ at test points.}
		\label{fig:Wilsonpolchinsky:test_fields:RL2}
	\end{subfigure} \\[1ex]
	\caption{Numerical results for Gaussian-model solution of the Wilson--Polchinski equation. Subfigures (A)--(C) show results at collocation points, while (D)--(F) present corresponding results at testing points. For visualization, only the first 3 of 1000 collocation points and the first 3 of 200 testing points are displayed. The relative $L^2$ error is evaluated for each scale variable $\kappa$.}
	\label{fig:Wilsonpolchinsky:GM}
\end{figure}

In this benchmark we fix \(P=20\).   We compute the reference solution \(W_\kappa\) using the explicit formula \eqref{eq:wp:explicit}. To study how the solver behaves on a broad collection of fields, and to check its ability to generalize beyond the inputs used to drive the flow, we work with a train-test split in function space. We first construct a random positive-definite covariance on Fourier coefficients by drawing a matrix \(A\in\R^{(P+1)\times(P+1)}\) with i.i.d.\ standard normal entries and setting $\Sigma \;=\; A^\top A$.   To generate the collocation functions, we draw \(N=1000\) coefficient vectors
\(\{\hat c^{(i)}\}_{i=1}^N\subset\R^{2P+1}\) from \(\mathcal N(0,\Sigma)\).
We enforce reality by mirroring \(\hat c_p\mapsto \hat c_{-p}\) and impose a mild high-frequency decay
\[
c_p\;\leftarrow\;\begin{cases}
\hat c_0, & p=0,\\[4pt]
\hat c_p/\lvert p\rvert, & p\neq0,
\end{cases}
\]
so that higher Fourier modes are present but slightly damped. Each resulting vector defines a real field
\(
J^{(i)}(x)=\sum_{p=-P}^{P}c_p^{(i)}\,e^{2\pi i p x}\in V_P,
\)
yielding the collocation set ${\rm J}=\bigl\{J^{(1)},\dots,J^{(N)}\bigr\}\subset V_P.$

To evaluate out-of-sample performance, we construct an independent test set of size \(N_{\mathrm{test}}=200\) using the same symmetry and decay, but with  coefficients \((\hat c_{-P}^{(j)},\dots,\hat c_{P}^{(j)})\sim\mathcal N(0,I)\) instead of the correlated covariance \(\Sigma\).  We denote these fields by $\tilde{{\rm J}}=\{\tilde J^{(j)}\}_{j=1}^{N_{\mathrm{test}}}.$
This independent ensemble plays the role of a hold-out set in function space and is used only for testing the generalization of the learned surrogate.

\XY{Since the Gaussian model solution depends quadratically on the Fourier coefficients}, we equip the field space on \(V_P\) with the  quadratic feature-space kernel
\begin{equation}
\label{eq:wp_wr_kernel}
k\bigl(J,J'\bigr)
=\sum_{p=-P}^{P}\bigl(c_p[J]\,c_p[J']+1\bigr)^2,
\end{equation}
where \(c_p[J]\) is the \(p\)th Fourier coefficient of \(J\).  
For the GP method described in \Cref{sec: methodology},  
a small nugget \(\varepsilon=10^{-12}\) is added to the Gram matrix for numerical stability. We denote the GP solution at scale $\kappa$ by $W_\kappa^\dagger$.

The initial condition of the ODE system is computed from the explicit solution in \eqref{eq:wp:explicit} at the ultraviolet scale \(\kappa_{\mathrm{UV}}=10\). We then integrate the associated ODE system in \(\kappa\) from \(\kappa_{\mathrm{UV}}\) down to \(\kappa_{\mathrm{IR}}=10^{-10}\).
The trajectories are sampled at \(N_T=100\) uniformly spaced scales
\[
  \kappa_\ell = \kappa_{\mathrm{UV}} + \ell\,\Delta\kappa,\qquad
  \Delta\kappa=\frac{\kappa_{\mathrm{IR}}-\kappa_{\mathrm{UV}}}{N_T},
  \quad \ell=0,\dots,N_T.
\]
At each \(\kappa_\ell\) we evaluate the exact solution and the GP surrogate on both 
\({\rm J}\) and \(\widetilde{\rm J}\).
To measure accuracy we use relative discrete \(L^2\) errors
\[
  \varepsilon_W^{\mathrm{coll}}(\kappa_\ell)
  := \frac{\bigl\|W_{\kappa_\ell}({\rm J})-W_{\kappa_\ell}^{\dagger}({\rm J})\bigr\|_2}
           {\bigl\|W_{\kappa_\ell}({\rm J})\bigr\|_2},
  \qquad
  \varepsilon_W^{\mathrm{test}}(\kappa_\ell)
  := \frac{\bigl\|W_{\kappa_\ell}(\widetilde{\rm J})-W_{\kappa_\ell}^{\dagger}(\widetilde{\rm J})\bigr\|_2}
           {\bigl\|W_{\kappa_\ell}(\widetilde{\rm J})\bigr\|_2},
\]
and their averages over the scale grid,
\[
  \bar\varepsilon_W^{\mathrm{coll}}
  = \frac{1}{N_T+1}\sum_{\ell=0}^{N_T}\varepsilon_W^{\mathrm{coll}}(\kappa_\ell),
  \qquad
  \bar\varepsilon_W^{\mathrm{test}}
  = \frac{1}{N_T+1}\sum_{\ell=0}^{N_T}\varepsilon_W^{\mathrm{test}}(\kappa_\ell).
\]

\subsubsection{Numerical results and discussion}
Figure~\ref{fig:Wilsonpolchinsky:GM} summarizes the performance of the GP solver on this quadratic benchmark.  
Panel~(A) shows three representative collocation fields generated by the sampling procedure above.
Panel~(B) compares, for the same three fields, the exact trajectories  obtained from \eqref{eq:wp:explicit} with the GP surrogates. The curves are visually indistinguishable across the whole range, indicating that the GP surrogate $W^\dagger$ accurately captures the functional dependence on the RG scale.

Panel~(C) reports the collocation relative errors \(\varepsilon_W^{\mathrm{coll}}(\kappa)\) as a function of \(\kappa\).
Panels~(D)–(F) repeat the same diagnostics for three randomly chosen test fields and for the full test ensemble
\(\widetilde\Phi\).  Again, the pointwise trajectories of \(W_\kappa\) and \(W_\kappa^\dagger\) coincide to plotting
accuracy, and the test errors \(\varepsilon_W^{\mathrm{test}}(\kappa)\) match the collocation errors in magnitude,
confirming out-of-sample generalization in function space.

Table~\ref{tb:wp} aggregates these results by listing \(\bar\varepsilon_W^{\mathrm{coll}}\) and
\(\bar\varepsilon_W^{\mathrm{test}}\) for increasing numbers of collocation fields \(N\).  
As \(N\) increases, the errors decay rapidly, providing numerical evidence for convergence. These results demonstrate that the proposed GP surrogate achieves accurate interpolation and generalization across the entire RG flow.

\subsection{Wetterich--Morris equation: lattice $\phi^4$-model}
\label{subsec:num:wetterich_lattice}
We work on a one-dimensional periodic lattice with \(N_x\in\mathbb{N}\) sites,
lattice spacing equal to \(1\), and index set
\(\mathbb{T}_{N_x}:=\mathbb{Z}/N_x\mathbb{Z}\) (indices taken modulo \(N_x\)).  
A lattice field is \(\phi=(\phi_i)_{i\in\mathbb{T}_{N_x}}\in\mathbb{R}^{N_x}\).  
The forward difference and second-order discrete Laplacian are
\[
(\nabla \phi)_i = \phi_{i+1}-\phi_i,\qquad
(-\Delta \phi)_i = 2\phi_i-\phi_{i+1}-\phi_{i-1}.
\]
We use the discrete inner product
\(\langle u,v\rangle := \sum_{i\in\mathbb{T}_{N_x}}u_i v_i\)
and the standard matrix trace \(\Tr(\cdot)\) on \(\mathbb{R}^{{N_x}\times {N_x}}\).

The effective average action \(\Gamma_\kappa[\phi]\) depends on the RG scale \(\kappa\) that flows backward from a large ultraviolet value \(\kappa_{\mathrm{UV}}\geq \pi\) to an infrared value \(0\leq \kappa_{\mathrm{IR}}<\pi/N_x\). For each \(\kappa\), a symmetric translation-invariant regulator \(R_\kappa\in\mathbb{R}^{N\times N}\) is given and \(\partial_\kappa R_\kappa\) denotes its scale derivative. The lattice Wetterich equation is
\begin{equation}
\label{eq:wett:lattice}
\partial_\kappa \Gamma_k[\phi]
=\frac{1}{2}\,\Tr\!\Big[\big(\Gamma_\kappa^{(2)}[\phi]+R_\kappa\big)^{-1}\,\partial_\kappa R_\kappa\Big],
\qquad
\Gamma_\kappa^{(2)}[\phi]:=\frac{\delta^2\Gamma_k[\phi]}{\delta \phi\,\delta \phi}\in\mathbb{R}^{N_x\times N_x}.
\end{equation}
At \(\kappa_{\mathrm{UV}}\), we prescribe the bare lattice \(\phi^4\) action
\begin{equation}
\label{eq:UV}
\Gamma_{\kappa_{\mathrm{UV}}}[\phi]
=\sum_{i\in\mathbb{T}_{N_x}}\Big(\tfrac12|(\nabla \phi)_i|^2+\tfrac{m^2}{2}\,\phi_i^2+\tfrac{\lambda}{24}\,\phi_i^4\Big),
\quad m^2\in\mathbb{R},\ \lambda>0 .
\end{equation}

To solve \eqref{eq:wett:lattice}, we place a GP prior on $\Gamma_\kappa$, induced by the local ansatz
\begin{equation}
\label{eq:local}
\Gamma_\kappa[\phi]
  = \sum_{i\in\mathbb{T}_{N_x}}
    \Big(\tfrac12 \lvert (\nabla \phi)_i \rvert^2 + U_\kappa(\phi_i)\Big),
\end{equation}
where the scalar potential $U_\kappa$ is modeled as a GP on a fixed field window $[-\phi_{\mathrm{max}},\phi_{\mathrm{max}}]$. 
\XY{Hence, by linearity, $\Gamma_\kappa$ inherits a GP prior induced by that on $U_\kappa$.}
Concretely, we decompose
\begin{equation}
\label{eq:parametrized_Uk_main}
U_\kappa(\phi) = m_0(\phi) + v_\kappa(\phi) + g_\kappa(\phi),
\qquad
\phi\in[-L,L],
\end{equation}
where
\[
m_0(\phi)=\tfrac12 m^2\phi^2+\tfrac{\lambda}{24}\phi^4
\]
is the bare \(\phi^4\) potential. We track the running quadratic and quartic couplings through a low-dimensional parametric term
\[
v_\kappa(\phi)
=
\operatorname{softplus}(\theta_{2,\kappa})\,\phi^2/2
+
\operatorname{softplus}(\theta_{4,\kappa})\,\phi^4/12,
\qquad
\operatorname{softplus}(\theta)=\ln\!\bigl(1+e^{\theta}\bigr),
\]
with latent parameters \(\theta_\kappa=(\theta_{2,\kappa},\theta_{4,\kappa})^\top\in\mathbb{R}^2\).
Finally,
\[
g_\kappa \sim \mathcal{GP}(0,K_U)
\]
is a zero-mean GP remainder with fixed covariance kernel \(K_U\) on
\([-\phi_{\mathrm{max}},\phi_{\mathrm{max}}]\), capturing the residual nonpolynomial structure.

{
Jointly determining $v_\kappa$ and $g_\kappa$ from the flow at each scale would amount to solving a fully nonlinear constrained regression problem in function space at every $\kappa$, which is prohibitively expensive over long RG trajectories. Instead, we adopt a predictor-projector scheme that interleaves ODE integration with constrained reconstruction. At each scale step, we first form a predictor by replacing the nonlinear parametrization of $v_\kappa$ with an unconstrained linear-Gaussian surrogate, yielding a closed, linearly parameterized curvature map $U_\kappa''$ as a function of the sampled potential values. We then apply a projection (correction) step: we fit a constrained polynomial parametrization together with a GP remainder to the predicted potential and replace the predicted state by the reconstructed potential. The corrected potential is used as the initial condition for the next predictor step. This produces an interleaved algorithm that enforces the constraints along the RG evolution while keeping the per-step cost manageable. Implementation details of the predictor closure, the projection objective, and the numerical integrator are given in Appendix~\ref{app:wetterich_gp_details}. Here we only summarize the numerical setup and results.

}


\subsubsection*{Numerical setup.} 
We discretize the periodic lattice with \(N_x=32\) sites and take the field window \([-\phi_{\mathrm{max}},\phi_{\mathrm{max}}]\) with \(\phi_{\mathrm{max}}=5.0\) on a uniform set of \(N_\phi=800\) constant-field collocation points, which is denoted by $X$. The remainder \(g_\kappa\)  is modeled by a GP  on \([-\phi_{\mathrm{max}},\phi_{\mathrm{max}}]\) with the kernel $K_U$ given explicitly by 
\begin{align}
\label{eq:phi4lattice}
K_U(\phi,\phi')
\;=\; \sum_{q=0}^{Q_{\max}}
\sigma^2\,\big(\eta^2 + (\pi q/\phi_{\mathrm{max}})^2\big)^{-\beta}\,
\cos\!\Big(\tfrac{\pi q}{\phi_{\mathrm{max}}}\,\phi\Big)\,
\cos\!\Big(\tfrac{\pi q}{\phi_{\mathrm{max}}}\,\phi'\Big),
\end{align}
with hyperparameters \(\sigma^2=1.0\), \(\eta=0.98\), \(\beta=2.0\), and spectral cutoff \(Q_{\max}=\min\{N_\phi-1,1024\}\).  The kernel Gram matrix is stabilized with a nugget \(\epsilon=10^{-11}\). \XY{The kernel choice in \eqref{eq:phi4lattice} reflects that the effective potential $U_\kappa$ should vary smoothly with the field, and that the model is even in the field, i.e., $U_\kappa(\phi)=U_\kappa(-\phi)$.
}

{
The flow is integrated backward in $\kappa$ from $\kappa_{\mathrm{UV}}=100$ to $\kappa_{\mathrm{IR}}=10^{-10}$, and we record the solution at $1001$ uniformly spaced output scales $\{\kappa_s\}_{s=0}^S$ with $S=1000$. We initialize the constrained parameters at the UV scale by $
\theta_{\kappa_{\mathrm{UV}},2}=\theta_{\kappa_{\mathrm{UV}},4}=-8.0$,
so that $\operatorname{softplus}(\theta)\approx 0$, and warm-start each subsequent projection at $\kappa_{s+1}$ using the previous iterate. At each scale, the projected parameters are obtained by minimizing the projection objective (Appendix~\ref{app:wetterich_gp_details}) with Tikhonov weight $\gamma=10^{-6}$ via a damped Gauss--Newton method (at most six iterations, unit damping). At the infrared scale $\kappa_{\mathrm{IR}}$, we set $\theta_{\mathrm{IR}}:=\theta_{\kappa_S}$ and denote by $g_{\mathrm{IR}}$ the GP posterior mean from the final projection. Together these define the reconstructed infrared potential
\[
U_{\mathrm{IR}}(\phi)
=
m_0(\phi)
+\operatorname{softplus}\!\big(\theta_{\mathrm{IR},2}\big)\,\phi^2/2
+\operatorname{softplus}\!\big(\theta_{\mathrm{IR},4}\big)\,\phi^4/12
+g_{\mathrm{IR}}(\phi).
\]
}

The reconstructed infrared potential \(U_{\mathrm{IR}}(\phi)\) is then used to extract observables (see Appendix~\ref{app_sec:mag_sus}). In particular, the magnetization \(m(c)\) is obtained by solving \(U'_{\mathrm{IR}}(m(c))=c\) with the Newton method on \(c\in[10^{-4},10^{0.4}]\) sampled at 18 logarithmic points, and the longitudinal susceptibility is given by \(\chi(c)=1/U''_{\mathrm{IR}}(m(c))\).

\begin{figure}
  \centering
  \begin{subfigure}{0.33\textwidth}
\centering\includegraphics[width=\linewidth]{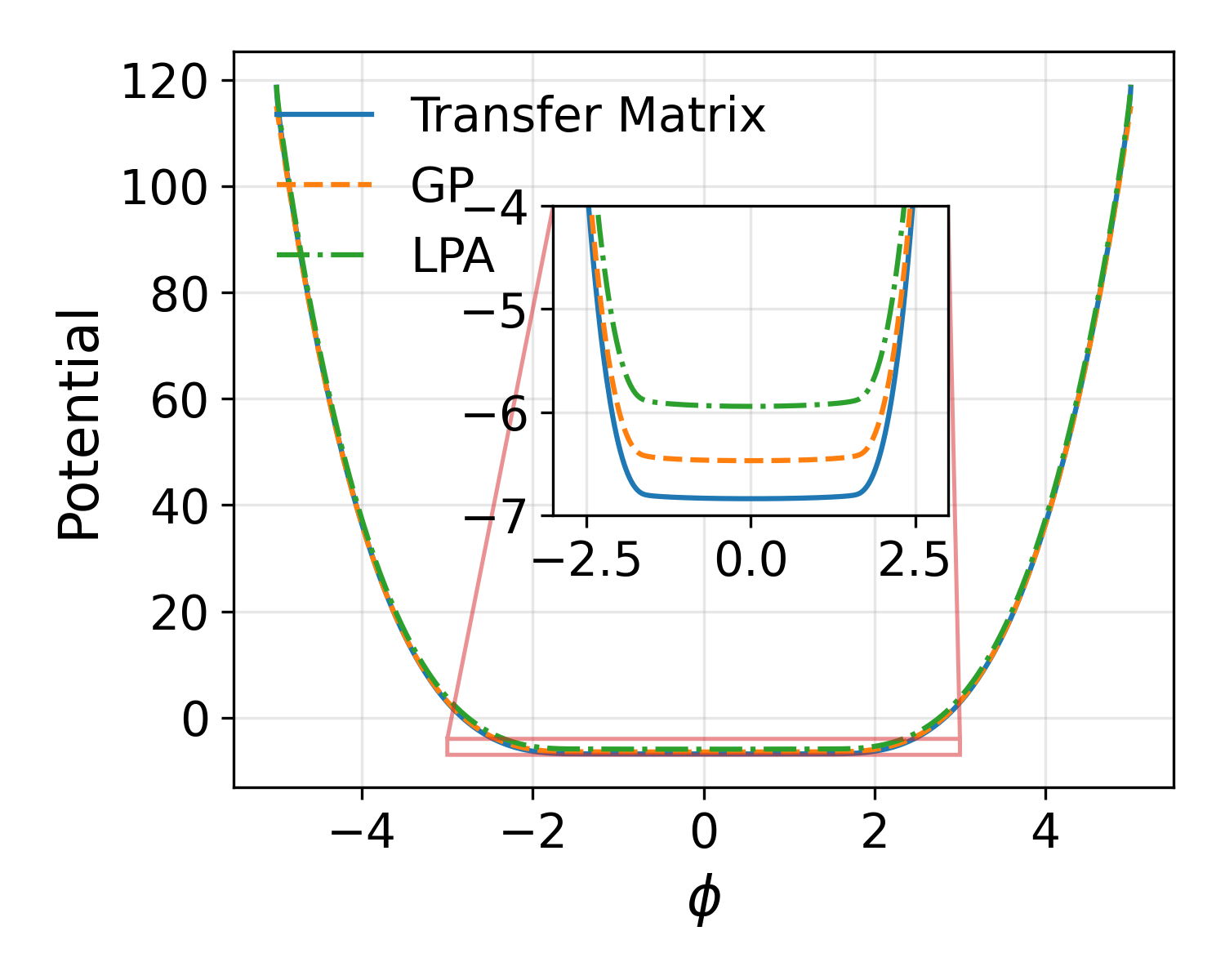}
    \caption{Potential $U_{\kappa_{\mathrm{IR}}}$}
    \label{fig:img1}
  \end{subfigure}\hfill
  \begin{subfigure}{0.33\textwidth}
    \centering
    \includegraphics[width=\linewidth]{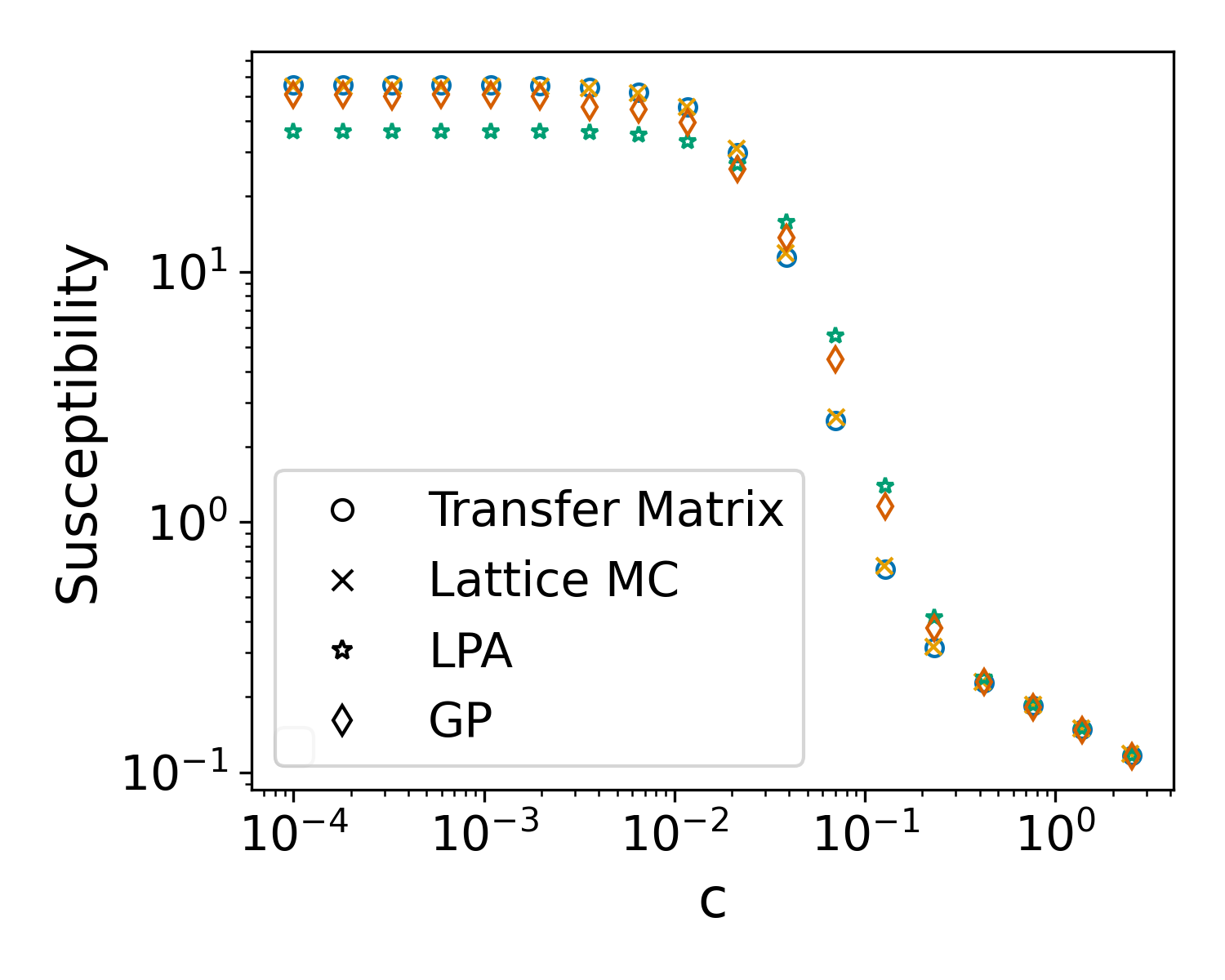}
    \caption{Susceptibility}
    \label{fig:img2}
  \end{subfigure}\hfill
  \begin{subfigure}{0.33\textwidth}
    \centering
    \includegraphics[width=\linewidth]{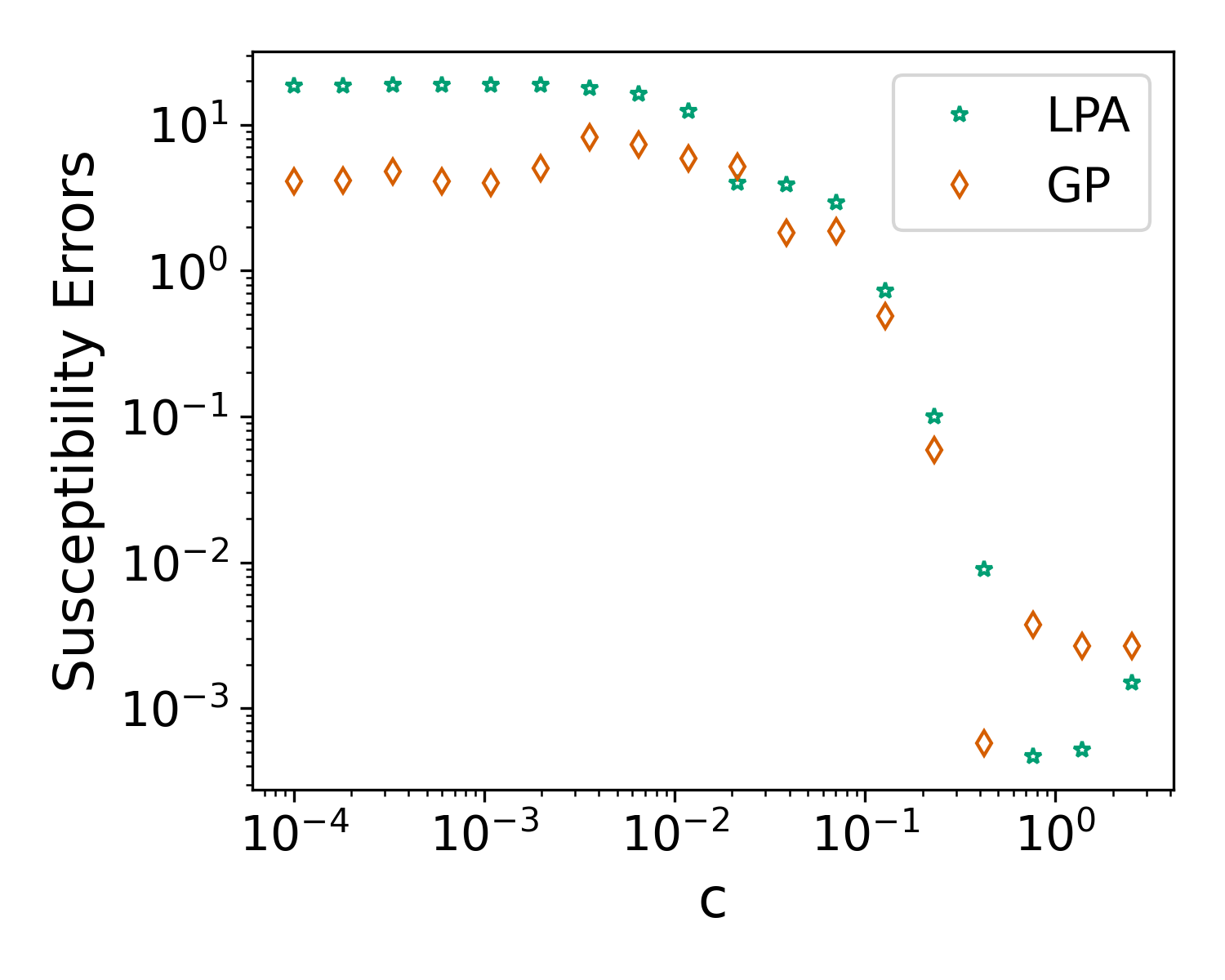}
    \caption{Susceptibility Errors}
    \label{fig:img3}
  \end{subfigure}

  \medskip

  \begin{subfigure}{0.33\textwidth}
    \centering
    \includegraphics[width=\linewidth]{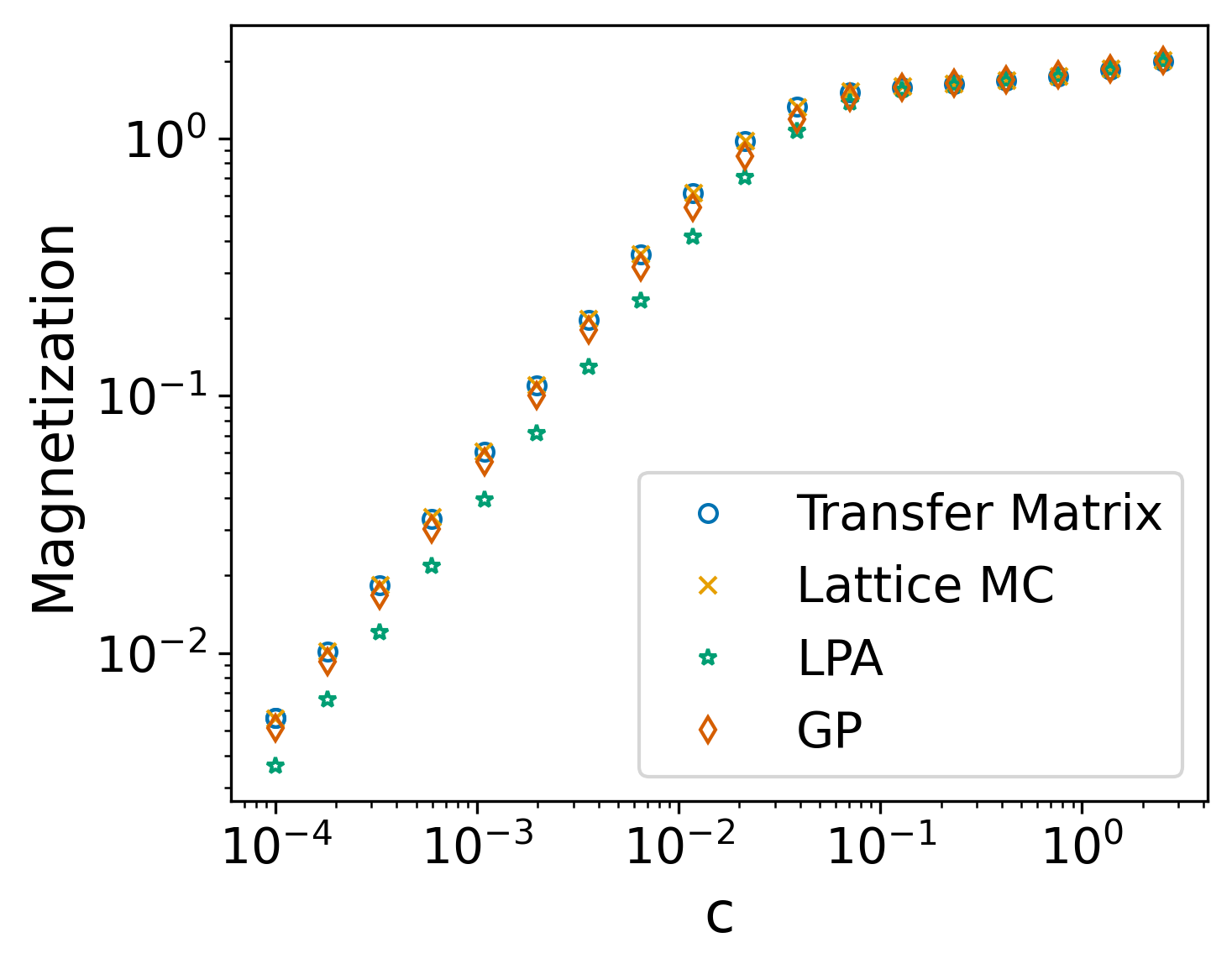}
    \caption{Magnetization}
    \label{fig:img4}
  \end{subfigure}
  \begin{subfigure}{0.33\textwidth}
    \centering
    \includegraphics[width=\linewidth]{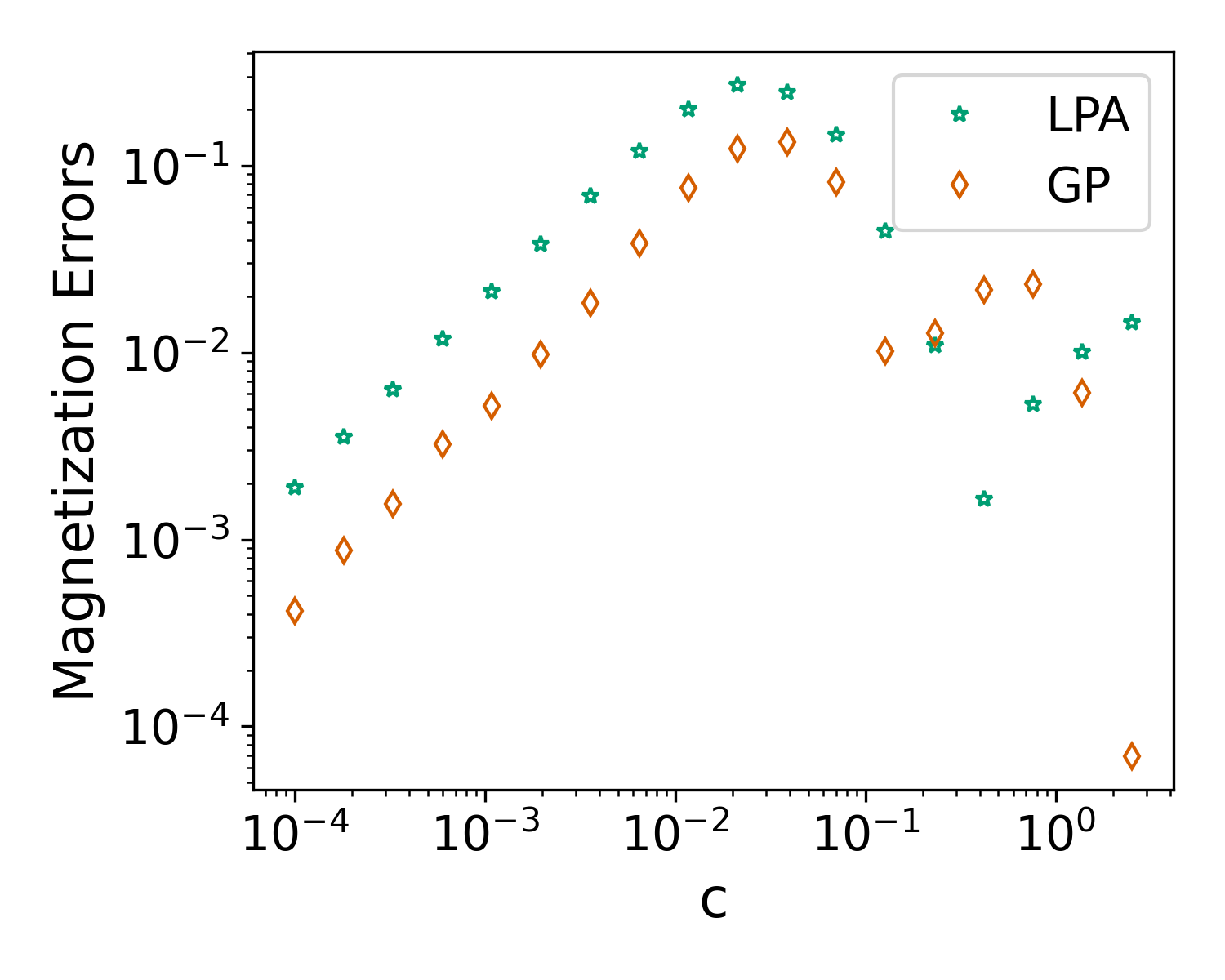}
    \caption{Magnetization Errors}
    \label{fig:img5}
  \end{subfigure}
\caption{
    Benchmark of the GP reconstruction versus standard LPA, lattice Monte Carlo, and the transfer matrix method.
    (A) Values of the potential $U_{\kappa_{\mathrm{IR}}}(\phi)$ on $[-5,5]$ of different methods.
    (B) Susceptibility $\chi(c)$ versus external field $c$ (log-log).
    (C) Susceptibility errors (relative to transfer matrix); for most values of $c$, the GP exhibits
    smaller errors than LPA, especially at small $c$.
    (D) Magnetization $m(c)$ (log-log).
    (E) Magnetization errors (relative to transfer matrix); in most of the $c$ range the GP is more
    accurate than LPA. }
  \label{fig:five-panel}
\end{figure}

\subsubsection*{Numerical results and discussion.}
Figure~\ref{fig:five-panel} reports a five-panel comparison of our GP-based reconstruction against a standard local potential flow (LPA) (see Appendix \ref{appendix:LPA_intro}), lattice Monte Carlo (MC) \cite{zorbach2025lattice}, and a transfer-matrix (TM) reference (see Appendix \ref{app_sec:TM}). For the one-dimensional lattice $\phi^4$ model with the parameters considered here, TM and high-precision lattice MC provide essentially state-of-the-art benchmarks for the magnetization $m(c)$ and the susceptibility $\chi(c)$ as functions of the external field $c$, see, e.g.,~\cite{zorbach2025lattice} and Appendix~\ref{app_sec:TM}. Panel~(A) shows the infrared effective potential $U_{\kappa_{\mathrm{IR}}}(\phi)$ on $[-5,5]$. Panel~(B) plots the susceptibility $\chi(c)$ on log-log axes, and panel~(C) displays the corresponding pointwise absolute errors $|\chi(c)-\chi_{\mathrm{TM}}(c)|$ relative to the TM reference. Panel~(D) shows the magnetization $m(c)$ on log-log axes, while panel~(E) reports the pointwise absolute errors $|m(c)-m_{\mathrm{TM}}(c)|$. In panels~(C) and~(E), smaller values indicate closer agreement with the TM benchmark.

The study \cite{zorbach2025lattice} demonstrates that, for this model, the standard LPA flow  already captures the qualitative behavior of $U_{\kappa_{\mathrm{IR}}}(\phi)$, $m(c)$, and $\chi(c)$, as well known, but exhibits systematic quantitative deviations when compared to lattice MC. These discrepancies originate from the fact that the truncation keeps only a local potential $U_\kappa(\phi)$ and neglects momentum-dependent corrections in $\Gamma_\kappa^{(2)}$, including wave-function renormalization and higher-derivative operators. As a result, the local flow misrepresents how fluctuations redistribute curvature in field space, and this manifests as a visible gap between the LPA and the lattice MC/TM results.

Our GP reconstruction is built on the same local ansatz for $\Gamma_\kappa$ that underlies the standard flow, but it replaces a finite low-order polynomial parameterization of $U_\kappa$ by a physics-informed GP prior. Concretely, we encode the bare $\phi^4$ potential in the mean function $m_0$, represent the running quadratic and quartic couplings in a low-dimensional parametric component $v_\kappa$, and model the remaining nonpolynomial structure by a zero-mean GP remainder $g_\kappa$ with a covariance kernel $K_U$ derived from a spectral decomposition of the lattice kinetic operator along constant fields. In this way, the GP kernel is not a generic smoothness prior but a kernel engineered from the  RG ansatz.

Using TM as the reference, the pointwise errors of $m(c)$ and $\chi(c)$ obtained from $U_{\kappa_{\mathrm{IR}}}$ are reduced for the GP reconstruction compared to the purely LPA flow on most values of $c$. \XY{For LPA, the computation requires $27.90$ seconds, whereas for the GP method it requires $59.95$ seconds.} Conceptually, this illustrates how physics-driven kernel design can be used to refine the effective description of $\Gamma_\kappa$ along selected observables: an ansatz or additional physical information (symmetries or invariances, etc.) for $\Gamma_\kappa$ can be incorporated into the construction of GP priors.

\section{Conclusion}
\label{sec:Conclusion}
We have introduced a general framework for solving functional renormalization
group (FRG) equations directly in function space, leveraging tools from
operator learning. The method is both general and flexible, and can be applied
to a broad class of FRG problems. We have demonstrated that it accurately
reproduces several flows of interest, while also revealing multiple directions
for further development.

On the methodological side, we have worked with a fixed kernel when solving the
FRG equations. Kernel-learning techniques such as cross-validation, maximum
likelihood estimation (MLE), or maximum a posteriori (MAP) estimation
\cite{nelsen2025bilevel, chenkf, hastie2009elements, rasmussen} could be used
to tune kernel hyperparameters and improve accuracy. In addition, approaches
based on programmable kernels \cite{kernelmodedecomposition, hypergraphdiscovery}
may help uncover structural properties of the solution. Once additional
physical insight into the underlying FRG solution is available, this knowledge
can be encoded directly into the GP kernel.

On the theoretical side, kernel-based methods open the door to rigorous error
analysis. Although we do not pursue this here, error estimates analogous to
those in \cite{kernelmethodsoperatorlearning} could be derived in our setting. \XY{The encoder-decoder structure of diagram in \Cref{fig:projection diagram} and the optimal-recovery viewpoint of Remark~\ref{rk:pic_remark} were first used in the operator-learning analyses of
\cite{kernelmethodsoperatorlearning} to provide convergence guarantees and rigorous a priori error estimates. We do not pursue such convergence analysis here, but note that it could be used in conjunction with an error analysis of time-stepping schemes to provide detailed error analysis of the recovery of $\Gamma$, and instead leave it for future work. In contrast to purely data-driven settings as in \cite{kernelmethodsoperatorlearning} where evaluations of \(\Gamma\) are given on many input functions,
our surrogate \(\Gamma^\dagger\) is constrained to satisfy the FRG flow equations. As a result, the representation is informed by the underlying functional PDE rather than by paired input-output data. } 
Moreover, the Gaussian process viewpoint naturally provides uncertainty
quantification and a Bayesian interpretation, which could be exploited in
inverse problems associated with functional differential equations.

In terms of applications, the proposed framework is particularly promising in
situations where traditional FRG techniques are difficult to use. Potential
targets include large deviation problems for dynamical systems and phenomena
such as spontaneous stochasticity. Our experiments also show that the GP
method can handle inhomogeneous fields, suggesting extensions to more complex
field theories, for instance the study of kink solutions in the $\phi^4$ model.










\section{Acknowledgments.} 
We are grateful to Malo Tarpin for many insightful discussions and suggestions that have substantially improved this work.
MD, HO, and XY acknowledge support from the Air Force Office of Scientific Research under MURI awards number FA9550-20-1-0358 (Machine Learning and Physics-Based Modeling and Simulation), FOA-AFRL-AFOSR-2023-0004 (Mathematics of Digital Twins), the Department of Energy under award number DE-SC0023163 (SEA-CROGS: Scalable, Efficient, and Accelerated Causal Reasoning Operators, Graphs and Spikes for Earth and Embedded Systems), the National Science Foundation under award number 2425909 (Discovering the Law of Stress Transfer and Earthquake Dynamics in a Fault Network using a Computational Graph Discovery Approach). HO acknowledges support from the DoD Vannevar Bush Faculty Fellowship Program under ONR award number N00014-18-1-2363. \GE{GE and MH were partially funded for this work by the Simons Foundation, Collaboration Grant MPS-1151713.} FA was supported by the Applied Mathematics (M2DT MMICC center) activity within the U.S. Department of Energy, Office of Science, Advanced Scientific Computing Research, under Contract DE-AC02-06CH11357.

\appendix

\section{Optimal recovery}
\label{app: optimal recovery}
{
Let \(\mathcal{U}\) be a Banach space with norm
\(\|\cdot\|_{\mathcal{U}}\), and let \(w:\mathcal{U}\to\mathbb{R}^m\) be a
measurement map with components \(w_i\in\mathcal{U}^*\) (the dual space of $\mathcal{U}$), so that
\(w(\varphi)=(w_1(\varphi),\dots,w_m(\varphi))^T\). Given data \(z=w(\varphi)\), we reconstruct \(\varphi\) from \(z\) by optimal recovery.
Namely, we take \(\tilde w\) to be a data-consistent
reconstruction rule that minimizes the worst-case relative error
\[
  \tilde w \in {\rm argmin}_{v:\,\mathbb{R}^m\to\mathcal{U}\atop w(v(z))=z}
  \ \sup_{\varphi\in\mathcal{U}\setminus\{0\}}
  \frac{\|\varphi - v(w(\varphi))\|_{\mathcal{U}}}{\|\varphi\|_{\mathcal{U}}}.
\]
Equivalently, for each \(z\), \(\tilde w(z)\) is the minimax-optimal interpolant
among all \(\psi\in\mathcal{U}\) satisfying \(w(\psi)=z\). As shown in 
\cite{owhadi2019operator}, this coincides with the minimum-norm interpolant:
\[
  \tilde{w}(z)
  := {\rm argmin}_{\psi \in \mathcal{U}}
      \|\psi\|_{\mathcal{U}}^2
  \quad\text{subject to}\quad
  w(\psi) = z.
\]
In this paper, we do not evaluate \(\tilde w\) explicitly. Rather, it provides
a useful conceptual link to optimal recovery and may serve as a starting point
for a convergence analysis of the proposed algorithm, as discussed in
\Cref{sec:Conclusion}.
}

\section{Further Experiments}
\label{app:extra_experiments}
\XY{In this section, we present two additional numerical examples to illustrate our methods. The first concerns the exact Gaussian-model solution of the Wetterich--Morris equation (Subsection~\ref{subsec:wetterich:gaussian}). Here, we use an exponential regulator rather than the Litim regulator used in \Cref{subsec:num:wp_gaussian}.}

The second experiment investigates the Wetterich flow on the continuum interval \([0,1]\) for a \(\phi^4\) model, whose bare action includes a quartic self-interaction term.  
Because of this nonlinear self-interaction, the corresponding flow equation cannot be solved in closed form, and the effective average action \(\Gamma_\kappa[\phi]\) is no longer available as an explicit analytic expression, in contrast to the Gaussian benchmarks. This experiment therefore examines the performance of our GP-based solver on a genuinely interacting, non-Gaussian flow without an explicit solution. We  draw a randomized set of collocation functions and learn the flow on that set. We then evaluate the learned flow on spatially constant functions, which allows a direct comparison with the local-potential approximation (LPA) method (see Appendix~\ref{appendix:LPA_intro}). \XY{Hence, although this example yields results similar to the LPA method, its purpose is to demonstrate that our approach can directly solve the Wetterich equation on non-constant fields, which the LPA cannot.
}

\subsection{The Wetterich-Morris equation: Gaussian-model}
\label{subsec:wetterich:gaussian}
We now turn to the Wetterich--Morris equation  
\begin{equation}\label{wmeqrs}
\partial_\kappa \Gamma_\kappa[\phi]
=\frac{1}{2} \int_0^1\!\!\int_0^1 \partial_\kappa R_\kappa(x,y)\,
\Bigl(\Gamma_\kappa^{(2)}[\phi]+R_\kappa\Bigr)^{-1}(x,y)\,\mathrm{d}x\,\mathrm{d}y,
\end{equation}
where \(\Gamma_\kappa^{(2)}[\phi]\) denotes the linear operator defined by the second functional derivative with
respect to \(\phi\).
We consider \eqref{wmeqrs} with effective average action
\(\Gamma_\kappa\) given explicitly by  
\begin{align}
\label{eq:wrg_esol}
\Gamma_{\kappa}[\phi] 
\;=&\;
\frac{1}{2}\,\int_0^1 \Bigl[\gamma\,(\nabla \phi(x))^{2}
+ \gamma\, m^{2}\,\phi(x)^{2}\Bigr]\mathrm{d}x
\;+\;
\frac{1}{2}\,\ln\!\frac{\det\bigl[-\gamma\,\Delta + \gamma\,m^{2} + R_{\kappa}\bigr]}
{\det\bigl[-\gamma\,\Delta + \gamma\,m^{2}\bigr]},
\end{align}
where \(m>0\) is the mass and \(\gamma>0\) is a small prefactor in front of the
quadratic term that moderates the influence of the Laplacian in the numerics.
The bare action for \eqref{wmeqrs} at the UV scale \(\kappa_{\mathrm{UV}}\) is
chosen by evaluating \eqref{eq:wrg_esol} at \(\kappa=\kappa_{\mathrm{UV}}\).
We refer to \eqref{eq:wrg_esol} as the Gaussian model, since for every scale
\(\kappa\) the corresponding bare action is a quadratic functional of \(\phi\)
(up to an additive constant). 

As in the Wilson--Polchinski experiment, we work in
\(
H^{1}_{\mathrm{per}}([0,1];\R)
\)
and in the truncated Fourier subspace \(V_P\) defined in \Cref{subsec:num:wp_gaussian}, representing fields by their Fourier
coefficients.  
The exponential regulator
\[
R_{\kappa}(p)
= \frac{\lvert p\rvert^{2\alpha}}
       {e^{\,(\lvert p\rvert/\kappa)^{2\alpha}} - 1},
\qquad \alpha = 1,
\]
enters both the logarithmic determinant in $\Gamma_\kappa$ and the flow~\eqref{wmeqrs} through its
real-space kernel $R_\kappa(x,y)$ and its scale derivative $\partial_\kappa R_\kappa(x,y)$ obtained by inverse
Fourier transform.  

We again place a GP prior on the restriction of $\Gamma_\kappa$ to $V_P$,
modeling $\Gamma_\kappa$ 
as a zero-mean GP with the kernel \eqref{eq:wp_wr_kernel}.

\subsubsection{Numerical setup}
For this benchmark, we use the explicit solution in~\eqref{eq:wrg_esol} with \(\gamma = 10^{-3}\) and \(m^{2} = 1\). We reuse the same truncation \(P\), the same Fourier representation, and the same sampling
strategy as in Section~\ref{subsec:num:wp_gaussian}.  
In particular, we work with \(N=1000\) collocation fields
\(\Phi=\{\phi^{(i)}\}_{i=1}^{N}\subset V_P\) and an independent test ensemble
\(\widetilde\Phi=\{\tilde\phi^{(j)}\}_{j=1}^{N_{\mathrm{test}}}\subset V_P\) with \(N_{\mathrm{test}}=200\), generated by the
correlated-Gaussian and i.i.d.\ Gaussian procedures described in \Cref{subsec:num:wp_gaussian}, including the same mild \(1/|p|\) decay on
high modes.

We use the explicit solution in  \eqref{eq:wrg_esol} to evaluate the effective action at the ultraviolet scale \(\kappa_{\mathrm{UV}}=10\), and take this as the initial condition for the flow. The associated ODE system in \(\kappa\) is then integrated from \(\kappa_{\mathrm{UV}}\) down to \(\kappa_{\mathrm{IR}}=10^{-10}\), and the values of \(\Gamma_\kappa\) are recorded on the same grid \(\{\kappa_\ell\}_{\ell=0}^{N_T}\) as in the Wilson--Polchinski example. We denote the GP solution at scale $\kappa$ by $\Gamma_\kappa^\dagger$.
At each \(\kappa_\ell\) we evaluate both the exact $\Gamma_{\kappa_\ell}$ and the surrogate
$\Gamma_{\kappa_\ell}^\dagger$ on \(\Phi\) and \(\widetilde\Phi\).

The collocation and test errors are defined exactly as in the Wilson--Polchinski experiment, but with
\(W_\kappa(J)\) replaced by \(\Gamma_\kappa(\phi)\).  
For clarity we denote these by \(\varepsilon_\Gamma^{\mathrm{coll}}(\kappa_\ell)\),
\(\varepsilon_\Gamma^{\mathrm{test}}(\kappa_\ell)\),
and their averages \(\bar\varepsilon_\Gamma^{\mathrm{coll}}\),
\(\bar\varepsilon_\Gamma^{\mathrm{test}}\), which are reported in Table~\ref{tb:wg}.

\subsubsection{Numerical results and discussion}

Figure~\ref{fig:Wetterich:GM} summarizes the behavior of the GP-based solver for the Wetterich Gaussian model.  
Panel~(A) displays three representative fields from the collocation set.  
Panel~(B) compares, for the same fields, the trajectories
\(\kappa\mapsto\Gamma_\kappa[\phi^{(i)}]\) with the GP surrogate
\(\kappa\mapsto\Gamma_\kappa^\dagger[\phi^{(i)}]\). The curves demonstrate accuracy across the full RG
range.  
Panel~(C) shows the collocation errors \(\varepsilon_\Gamma^{\mathrm{coll}}(\kappa)\) as a function of \(\kappa\); they remain uniformly small along the flow.

Panels~(D)–(F) repeat these diagnostics for three test fields and for the entire test ensemble
\(\widetilde\Phi\).  
The out-of-sample trajectories \(\Gamma_\kappa[\tilde\phi^{(j)}]\) and
\(\Gamma_\kappa^\dagger[\tilde\phi^{(j)}]\) remain close in absolute size, even
though the test errors \(\varepsilon_\Gamma^{\mathrm{test}}(\kappa)\) are one to
three orders of magnitude larger than the collocation errors and are still small on
the scale of the function values. 
Together with the averages \(\bar\varepsilon_\Gamma^{\mathrm{coll}}\) and \(\bar\varepsilon_\Gamma^{\mathrm{test}}\) in
Table~\ref{tb:wg}, these results show that the GP surrogate provides an accurate and
stable approximation of the Wetterich flow for this Gaussian benchmark, both on and off the training set.

\begin{figure}[!hbtp]
	\centering 
    \begin{subfigure}[b]{0.32\textwidth}
		\includegraphics[width=\textwidth]{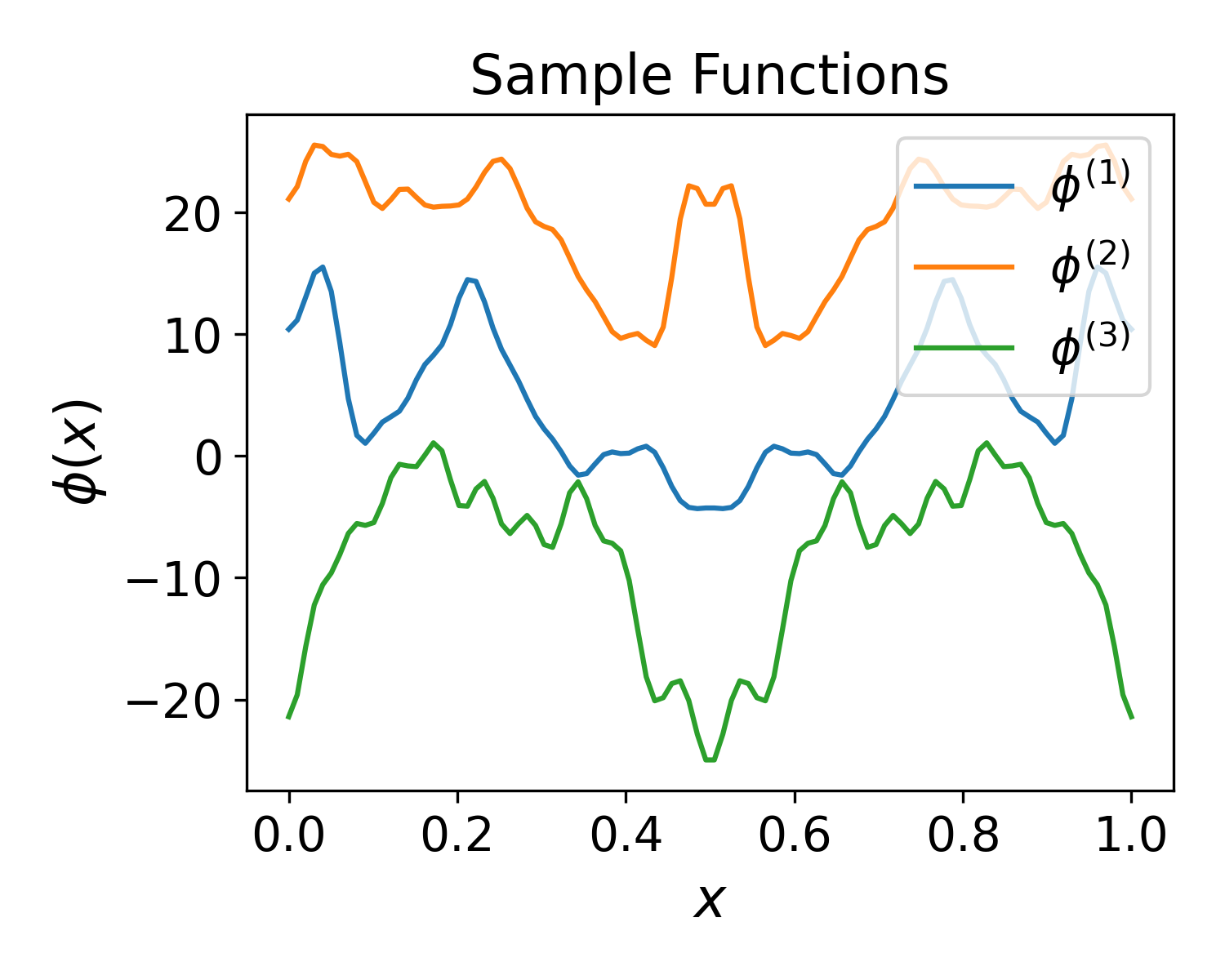}
		\caption{The first three of 1000 collocation points.}
		\label{fig:WetterichGaussian:collocation_fields}
	\end{subfigure}
	\begin{subfigure}[b]{0.32\textwidth}
		\includegraphics[width=\textwidth]{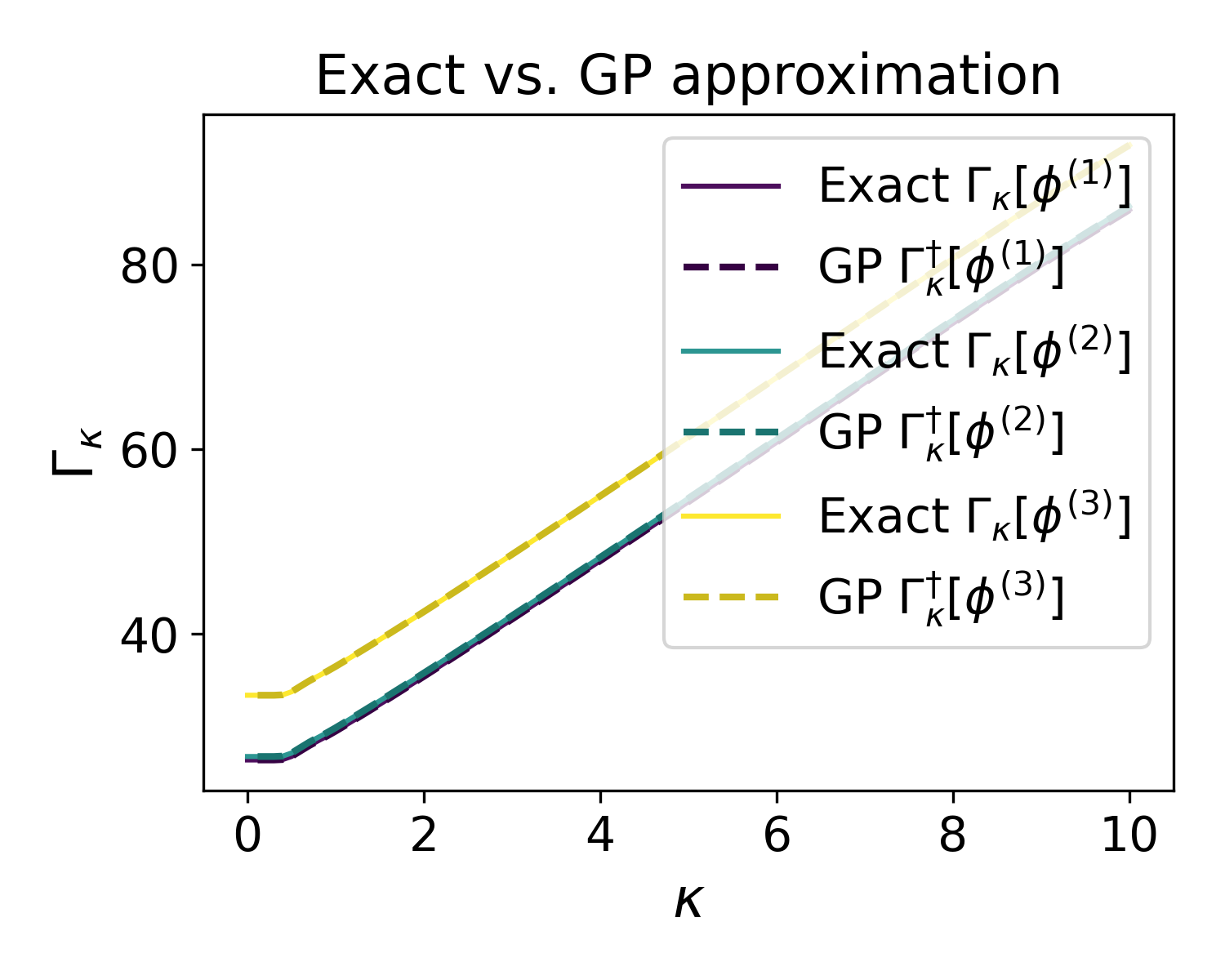}
		\caption{Exact solutions vs. GP approximations.}
		\label{fig:WetterichGaussian:collocation_fields}
	\end{subfigure} 
    \begin{subfigure}[b]{0.32\textwidth}
		\includegraphics[width=\textwidth]{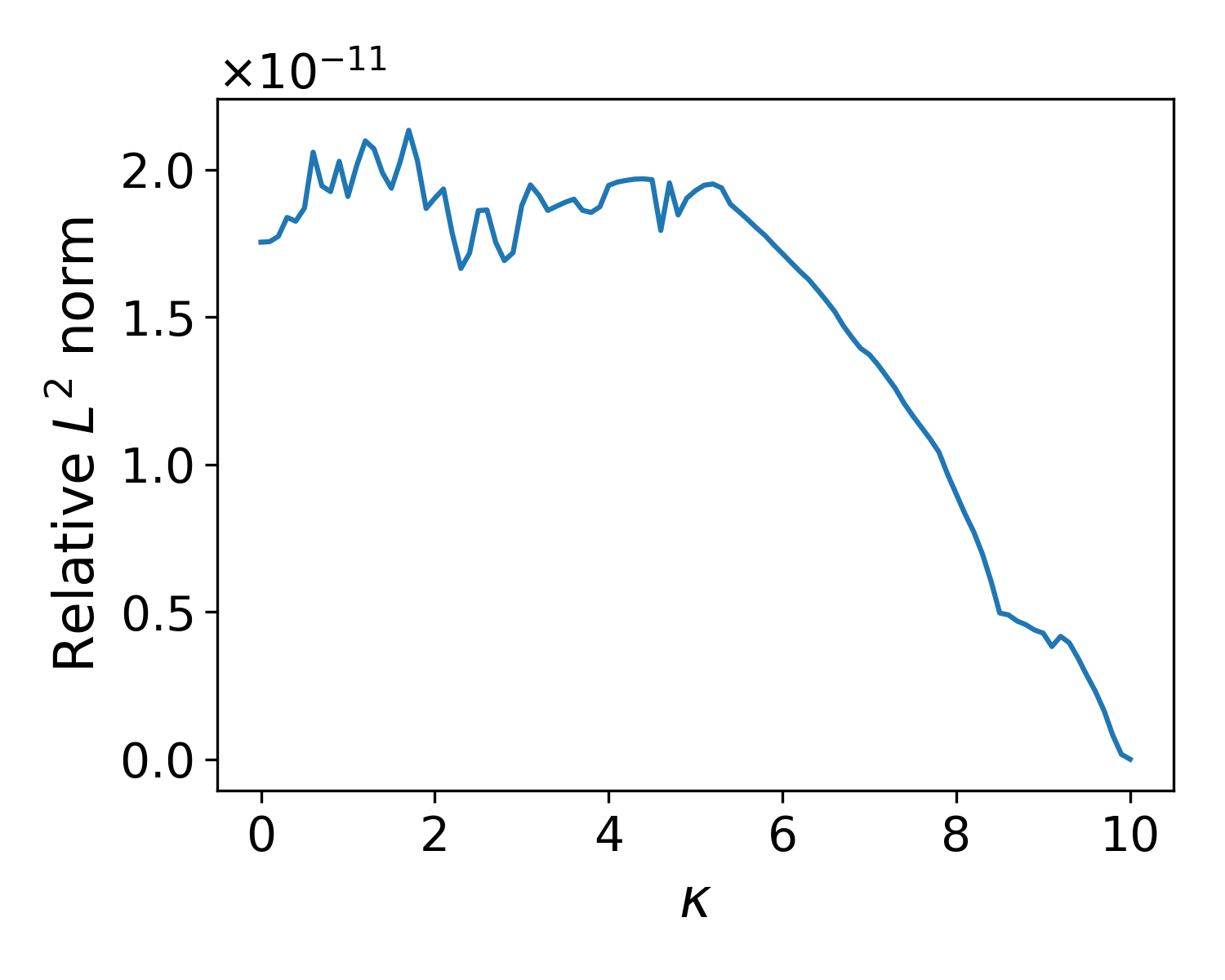}
		\caption{Relative $L^2$ error per scale variable $\kappa$ at collocation points.}
		\label{fig:WetterichGaussian:collocation_fields:RL2}
	\end{subfigure} \\[1ex] 
    \begin{subfigure}[b]{0.32\textwidth}
		\includegraphics[width=\textwidth]{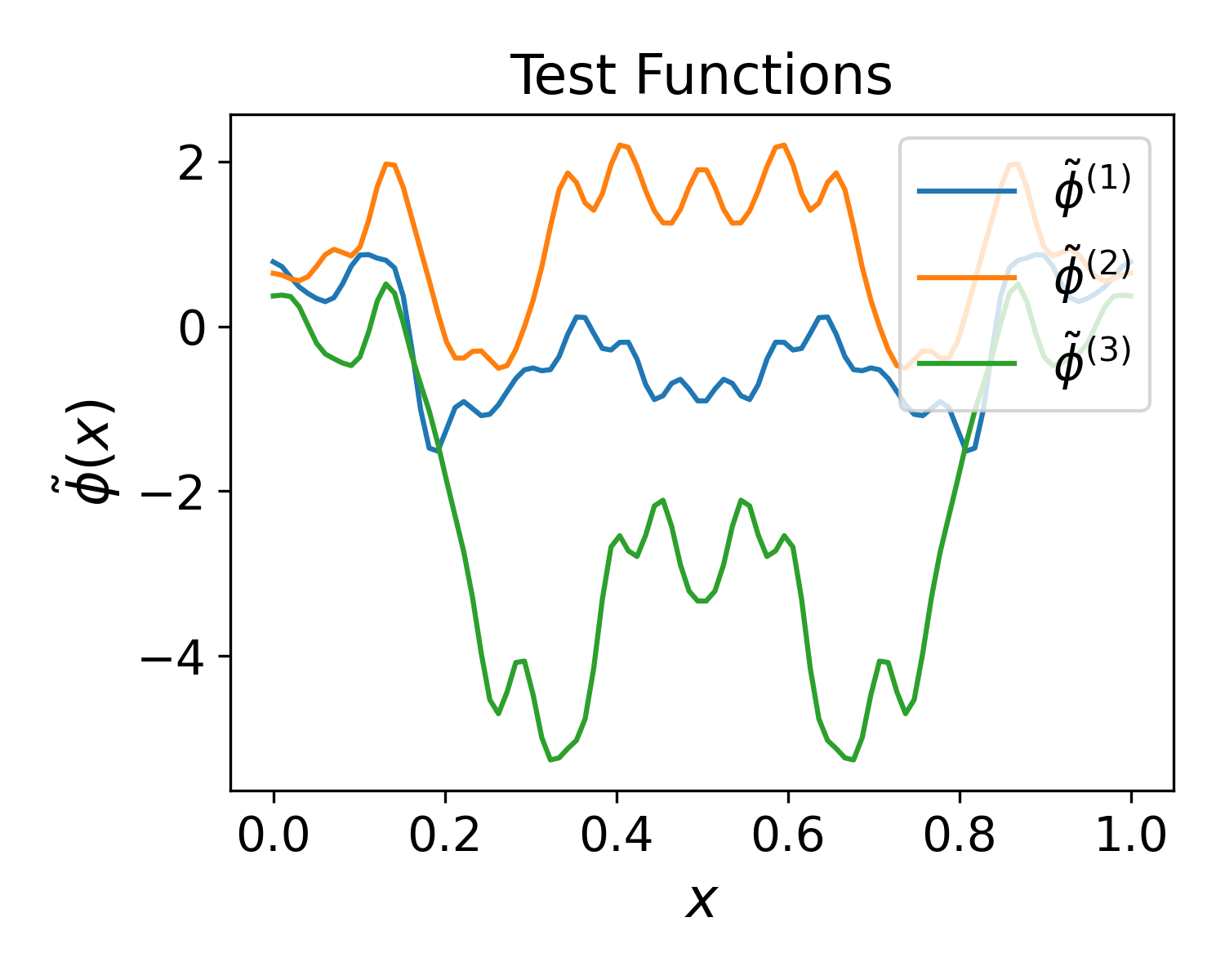}
		\caption{The first three of 200 test points.}
		\label{fig::collocation_fields}
	\end{subfigure}
	\begin{subfigure}[b]{0.32\textwidth}
		\includegraphics[width=\textwidth]{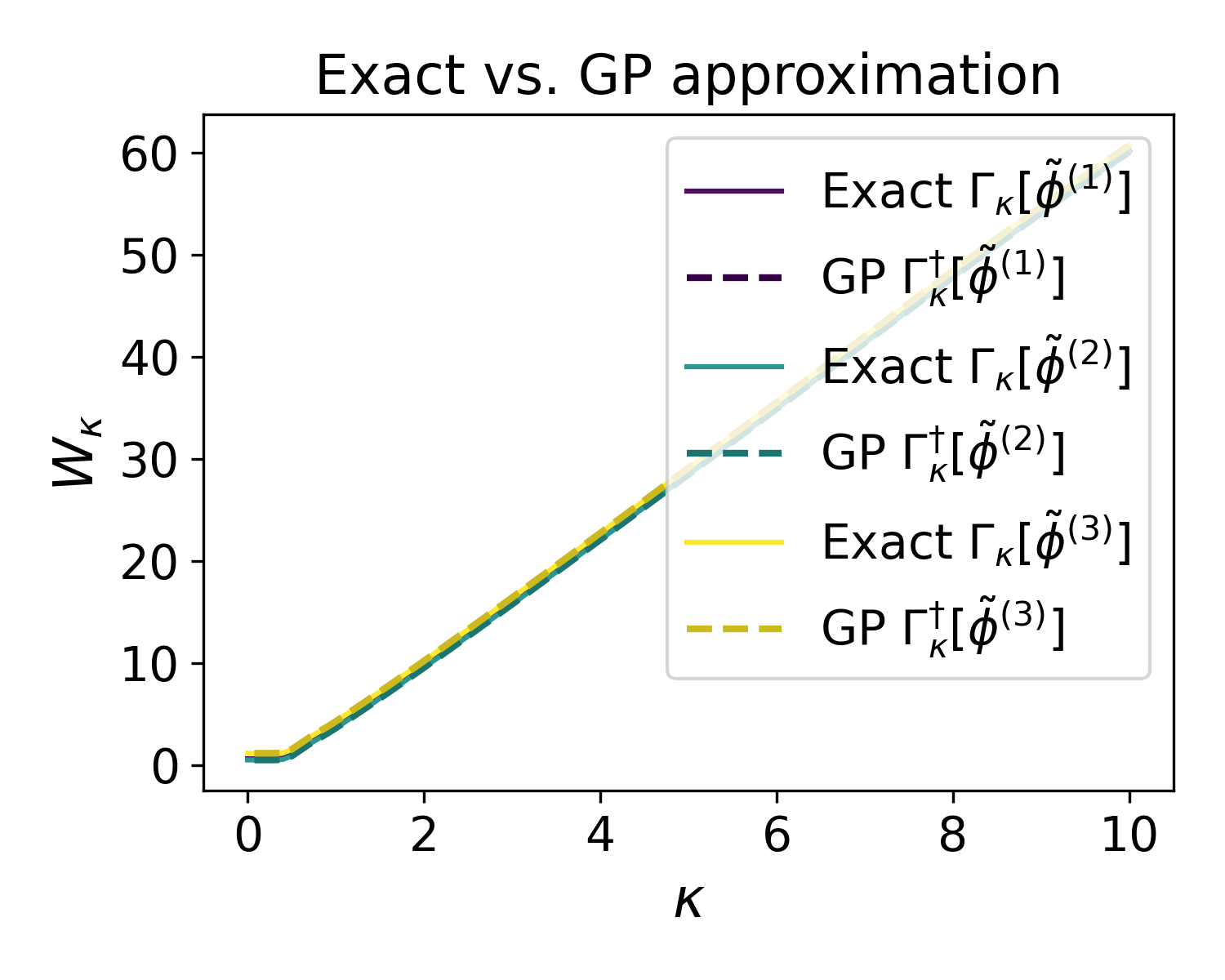}
		\caption{Exact solutions vs. GP approximations.}
		\label{fig:WetterichGaussian:test_fields}
	\end{subfigure}
	\begin{subfigure}[b]{0.32\textwidth}
		\includegraphics[width=\textwidth]{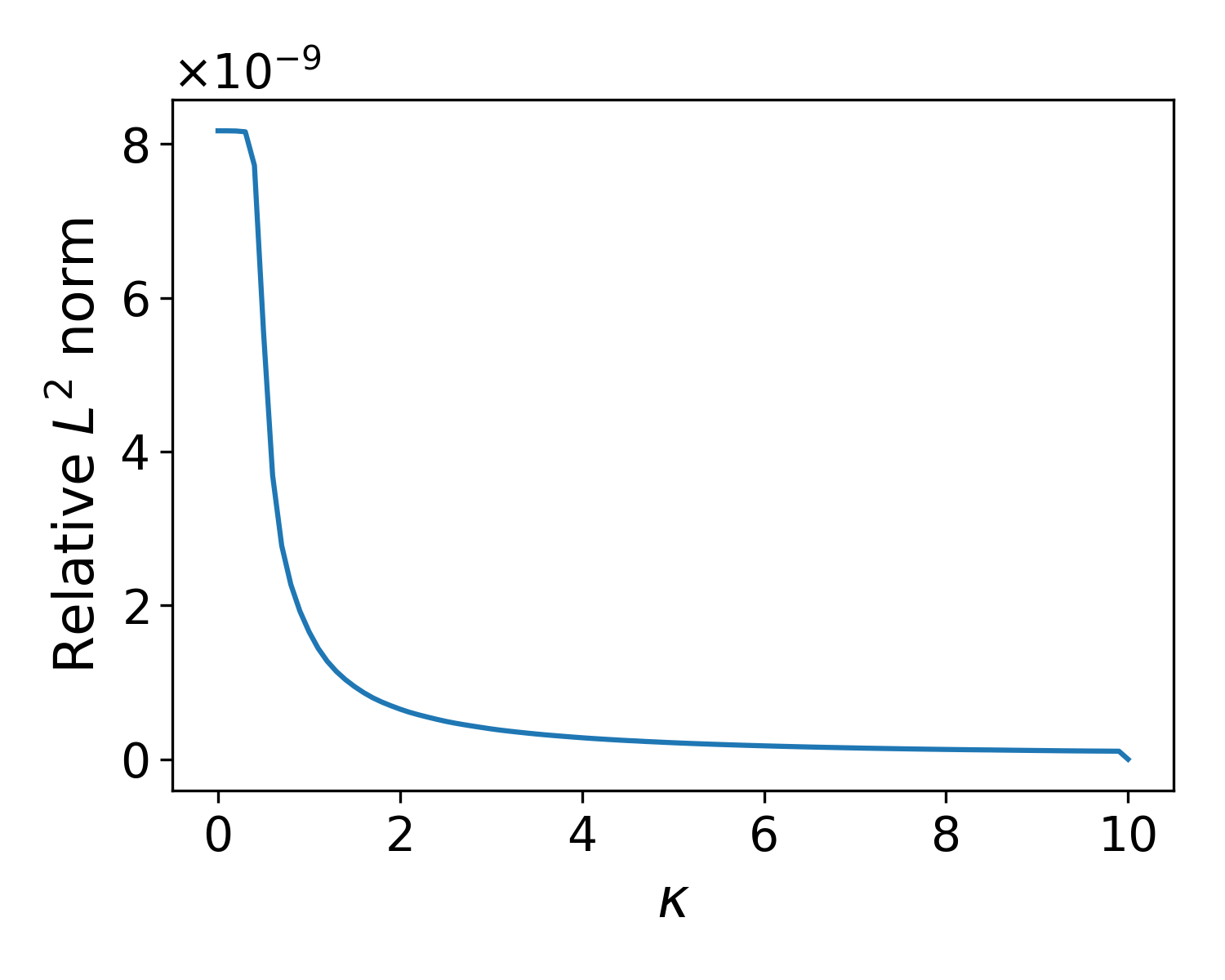}
		\caption{Relative $L^2$ error per scale variable $\kappa$ at test points.}
		\label{fig:WetterichGaussian:test_fields:RL2}
	\end{subfigure} \\[1ex]
	\caption{Numerical results of the Wetterich Gaussian model. Subfigures (A)--(C) show results at collocation points, while (D)--(F) present corresponding results at testing points. For visualization, only the first 3 of 1000 collocation points and the first 3 of 200 testing points are displayed. The relative $L^2$ error is evaluated for each scale variable $\kappa$.}
	\label{fig:Wetterich:GM}
\end{figure}

\begin{table}[ht]
  \centering
  \caption{The Wetterich Gaussian Model: Average relative $L^2$ errors on collocation and test sets for varying sample sizes $N$.}
  \label{tab:avg-rel-l2-errors-horizontal}
  \begin{tabular}{|c|c|c|c|c|}
    \hline
                  &  $N=100$ & $N=200$ & $N=500$ & $N=1000$ \\ 
    \hline
    Collocation set 
                  &   $1.31\times 10^{-9}$  &  
                   $1.54\times 10^{-10}$ & $5.40\times 10^{-11}$  &  $1.48\times 10^{-11}$  \\ 
    \hline
    Test set    
                  &    $1.27 \times 10^{-6}$  & $4.62\times 10^{-8}$ & $3.07\times 10^{-9}$  &  $8.62\times 10^{-10}$  \\ 
    \hline
  \end{tabular}
\label{tb:wg}
\end{table}

\subsection{The Wetterich equation: continuum $\phi^{4}$-model.} 
\label{subsec:wetterich:phi4}
In this subsection, we consider the Wetterich--Morris equation
\begin{equation}\label{eq:wetterich:phi4}
\partial_\kappa \Gamma_\kappa[\phi]=\frac{1}{2} \int \partial_\kappa R_\kappa(x,y)\bigg(\frac{\delta^2 \Gamma_\kappa[\phi]}{\delta \phi\,\delta \phi}+R_\kappa\bigg)^{-1}(x,y)\,dx\,dy, 
\end{equation}
with the \(\phi^4\) bare  action on the periodic unit interval $[0,1]$ at $\kappa=\kappa_{\mathrm{UV}}:$
\begin{align}
\label{eq:wetterich:bare_action}
\Gamma_{\kappa_{\mathrm{UV}}}[\phi]
    = \int_{0}^{1}\!\mathrm{d}x\;\left[\frac{1}{2}(\partial_x\phi)^2
      + \frac{m^2}{2}\,\phi^2 + \frac{\lambda}{4!}\,\phi^4\right],
    \quad \kappa_{\mathrm{UV}}=10,\;m^2=-1.5,\;\lambda=1, 
\end{align}
and the Litim regulator in Fourier representation 
\begin{align}
\label{eq:wetterich:phi4:litim}
R_k(q^2)=(k^2-q^2)\,\Theta(k^2-q^2). 
\end{align} 
We work in the periodic Sobolev space
\[
H^1_{\mathrm{per}}([0,1])
=\bigl\{\phi\in H^1([0,1]) : \phi(0)=\phi(1)\bigr\},
\]
and use the standard real Fourier basis \(\{e_\alpha\}_{\alpha=0}^{\infty}\subset H^1_{\mathrm{per}}([0,1])\) given by
\[
  e_0(x)=1,\qquad
  e_\alpha(x)=
  \begin{cases}
    \sqrt{2}\,\cos\bigl(2\pi\,p(\alpha)\,x\bigr), & \alpha \text{ odd},\\[4pt]
    \sqrt{2}\,\sin\bigl(2\pi\,p(\alpha)\,x\bigr), & \alpha \text{ even},\ \alpha>0,
  \end{cases}
\]
with \(p(\alpha)=\lfloor(\alpha+1)/2\rfloor\).  
Any \(\phi\in H^1_{\mathrm{per}}([0,1])\) then has the expansion
\[
  \phi(x)=\sum_{\alpha=0}^{\infty} c_\alpha\,e_\alpha(x),
  \qquad
  c_\alpha = \int_{0}^{1}\phi(x)\,e_\alpha(x)\,\mathrm{d}x.
\]
For the numerical experiments we restrict to the finite–dimensional subspace
\begin{equation}
\label{eq:wetterich:phi4:subspace}
V_P
=\Bigl\{
  \phi\in H^1_{\mathrm{per}}([0,1]) : 
  \phi(x)=\sum_{\alpha=0}^{2P} c_\alpha\,e_\alpha(x)
\Bigr\},
\end{equation}
of dimension \(2P+1\).

\subsubsection{Our Gaussian Process Method} Based on the physical prior, we build our GP prior \HO{for} $\Gamma$ for solving \eqref{eq:wetterich:phi4} based on the LPA ansatz (see Appendix \ref{appendix:LPA_intro} for an introduction of the LPA method)
\begin{align}
 \Gamma_k[\phi]
  = \int_{0}^{1}\!\mathrm{d}x\;\left[
    \frac{1}{2}\bigl(\partial_x\phi(x)\bigr)^{2}
    + U_k\!\bigl(\phi(x)^{2}\bigr)
  \right],
\end{align}
More precisely, we  place a fully nonparametric prior on the local potential by assuming
\[
U_k(u)\;\sim\;\mathcal{GP}\bigl(0,\;K_{U}\bigr),
\]
a zero‐mean GP in the scalar argument \(u=\phi^{2}\).  Because \(\Gamma_\kappa[\phi]\) depends linearly on \(U_\kappa\), this induces a GP prior on the entire functional \(\phi\mapsto\Gamma_\kappa[\phi]\).  Under these assumptions,
\[
\Gamma_\kappa[\cdot]\;\sim\;
\mathcal{GP}\Bigl(m_{\Gamma},C_{\Gamma}\Bigr), \quad 
m_{\Gamma}(\phi)=\frac{1}{2}\!\int_{0}^{1}(\partial_{x}\phi)^{2}dx,\;
C_{\Gamma}(\phi,\phi')=\!\int_{0}^{1}\!\!\int_{0}^{1}
K_{U}\!\bigl(\phi(x)^{2},\phi'(y)^{2}\bigr)\,dx\,dy.
\]
This construction thus carries the nonparametric GP prior on \(U_\kappa\) all the way through to the full functional \(\Gamma_\kappa[\phi]\).  Moreover, the choice of the scalar kernel \(K_U\) is inspired by the form of the bare action in \eqref{eq:wetterich:bare_action}.  In particular, for the \(\phi^4\) model we adopt the additive kernel
\begin{align}
\label{eq:wetterichphi4:additive_kernel}
K_U(u,u') \;=\; 1 \;+\; u\,u' \;+\; \bigl(u\,u'\bigr)^2 + \operatorname{exp}(-|u - u'|^2/(2\sigma^2)). 
\end{align}

\subsubsection{Numerical experiment setup}
\label{subsec:wetterich:phi4:setup}

We restrict the fields to the finite-dimensional subspace \(V_P\subset H^1_{\mathrm{per}}([0,1])\) defined in \eqref{eq:wetterich:phi4:subspace}, spanned by the first \(2P+1\) real Fourier modes.  To construct the collocation ensemble, we draw \(N_{\mathrm{col}}=1000\) random coefficient vectors \(c^{(i)}=(c^{(i)}_0,\dots,c^{(i)}_{2P})\in\R^{2P+1}\) from a correlated Gaussian prior with covariance \(\Sigma=A^\top A\) (for a random Gaussian matrix \(A\)), and then apply a mild high-frequency decay
\[
c^{(i)}_\alpha \;\leftarrow\;
\frac{c^{(i)}_\alpha}{\bigl(1+\lvert p(\alpha)\rvert\bigr)^{3/2}},
\]
\XY{thereby biasing the sampled fields toward regular configurations while still retaining nontrivial high-frequency modes and improving the conditioning of the computation of the functional Hessian,}
where \(p(\alpha)\) is the Fourier mode index.
 Each coefficient vector defines a real periodic field
\[
  \phi^{(i)}(x) = \sum_{\alpha=0}^{2P} c^{(i)}_\alpha\,e_\alpha(x),
  \qquad i=1,\dots,N_{\mathrm{col}}.
\]
The Gram matrix \(K(\Phi,\Phi)\) induced by the scalar kernel \(K_U\) in \eqref{eq:wetterichphi4:additive_kernel} is assembled once and stabilized with a small nugget.

For each RG scale \(\kappa\), the GP posterior defines a surrogate functional $\Gamma_\kappa^\dagger[\phi]$ 
which approximates \(\Gamma_\kappa[\phi]\) for all \(\phi\in V_P\).   Inserting \(\Gamma_\kappa^\dagger\) into the right-hand side of the Wetterich equation \eqref{eq:wetterich:phi4}, and projecting onto the collocation ensemble, yields a closed system of ODEs in \(\kappa\) for these coefficients. At the ultraviolet scale \(\kappa_{\mathrm{UV}}\) we initialize the flow with the bare action \eqref{eq:wetterich:bare_action}. Concretely, for each collocation field \(\phi^{(i)}\), we compute its bare functional value \(\Gamma_{\kappa_{\mathrm{UV}}}[\phi^{(i)}]\), using the real Fourier representation to compute the kinetic term exactly and the potential term by averaging the integrand over the grid.  Collecting these values into the vector
\[
  \Gamma_{\kappa_{\mathrm{UV}}}(\Phi)
  :=
  \bigl(\Gamma_{\kappa_{\mathrm{UV}}}[\phi^{(1)}],\dots,
        \Gamma_{\kappa_{\mathrm{UV}}}[\phi^{(N_{\mathrm{col}})}]\bigr)^{\!\top},
\]
we treat \(\Gamma_{\kappa_{\mathrm{UV}}}(\Phi)\) as the initial condition of the ODE system. We integrate the resulting ODE system backward in scale from \(\kappa_{\mathrm{UV}}=10\) down to \(\kappa_{\mathrm{IR}}=2\).  The solver outputs \((\Gamma^\dagger_\kappa[\phi^{(i)}])_{i=1}^{N_{\mathrm{col}}}\) along a uniform grid of RG scales \(\{\kappa_\ell\}_{\ell=0}^{N_T}\subset[2,10]\).

Since no explicit solution is available in this case, we compare the GP surrogate $\Gamma_\kappa^\dagger(\phi)$ with an independently computed LPA solution $U_\kappa(\phi^2)$. 
The latter is obtained by solving the flow for the local potential on a grid of $\phi^2$ values using finite differences and the same Litim regulator~\eqref{eq:wetterich:phi4:litim}. Concretely, we sample a uniform grid of values
\(\{\widetilde{\phi}^{(j)}\}_{j=1}^{N_{\mathrm{test}}}\subset[0,4]\)
and regard each \(\widetilde{\phi}^{(j)}\) as a constant test field.
In the real Fourier basis, these fields are represented by activating only the constant mode.  For the LPA reference, we independently solve the standard LPA flow equation for the local potential \(U_\kappa(u)\) (with \(u=\phi^2\)) using the same bare parameters as in \eqref{eq:wetterich:bare_action}. The variable \(u\) is discretized on a uniform grid, and the resulting ODE system for \(\{U_\kappa(u_j)\}\) is integrated in the RG scale from the ultraviolet cutoff \(\kappa_{\mathrm{UV}}=10\) down to an infrared scale \(\kappa_{\mathrm{IR}}=2\). We compare the LPA potential $U_\kappa(\phi^2)$ with the GP surrogate by evaluating both on the same $(\kappa,\phi)$ grid. \XY{The comparison between the GP method and LPA in this section is not intended to demonstrate that we can reproduce LPA’s results \GE{which, of course, are only approximate.} Rather, it illustrates the ability of our method to solve the functional on non-constant fields, which lie outside the scope of the LPA approximation, and to generalize the resulting functional approximation to previously unseen fields.}

\begin{figure}[!hbtp]
	\centering          
	\begin{subfigure}[b]{0.32\textwidth}
		\includegraphics[width=\textwidth]{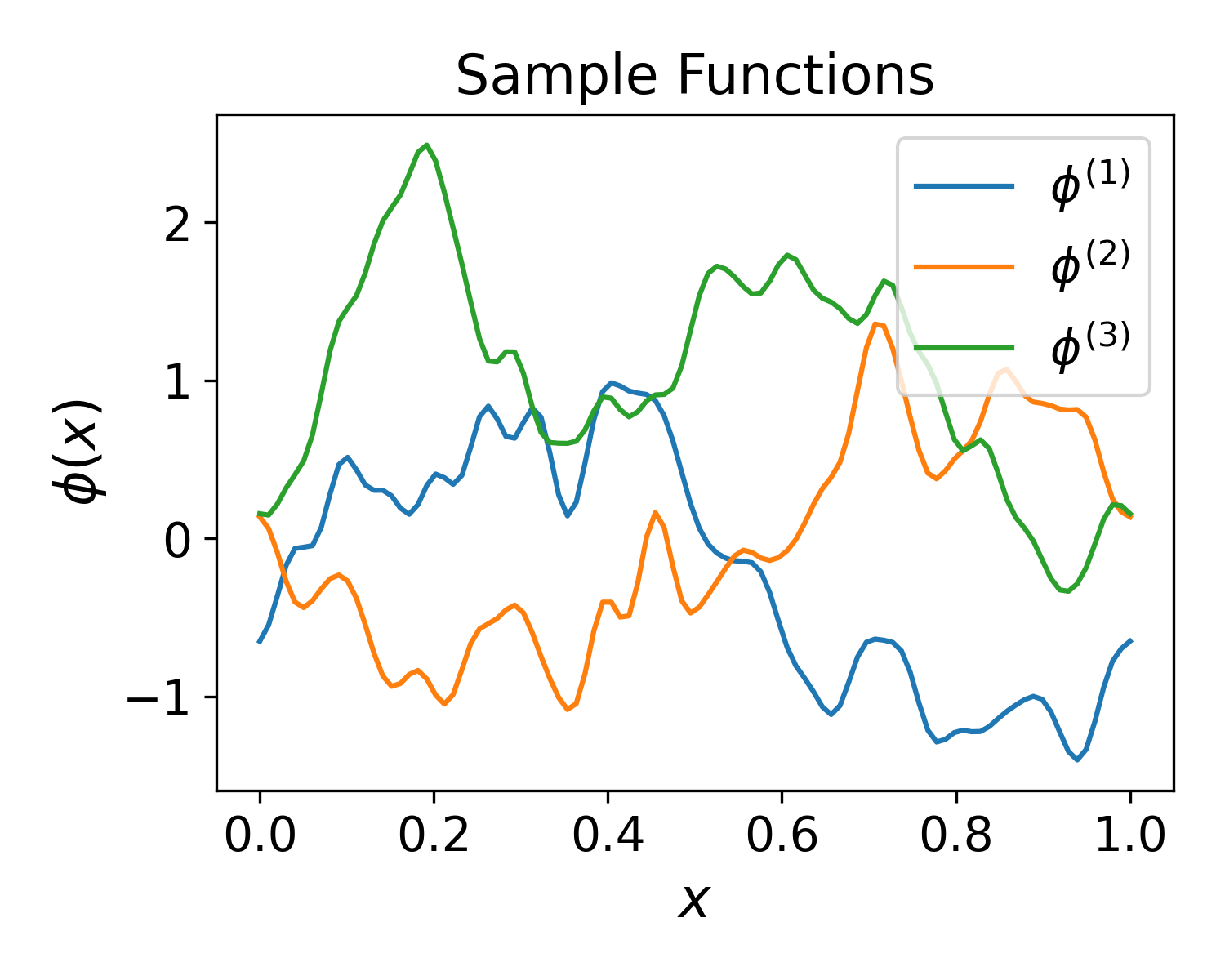}
		\caption{Three representative training fields $\{\phi_i\}_{i=1}^{3}$.}
		\label{fig:Wetterich_phi4:collocation_fields}
	\end{subfigure} 
	\begin{subfigure}[b]{0.32\textwidth}
		\includegraphics[width=\textwidth]{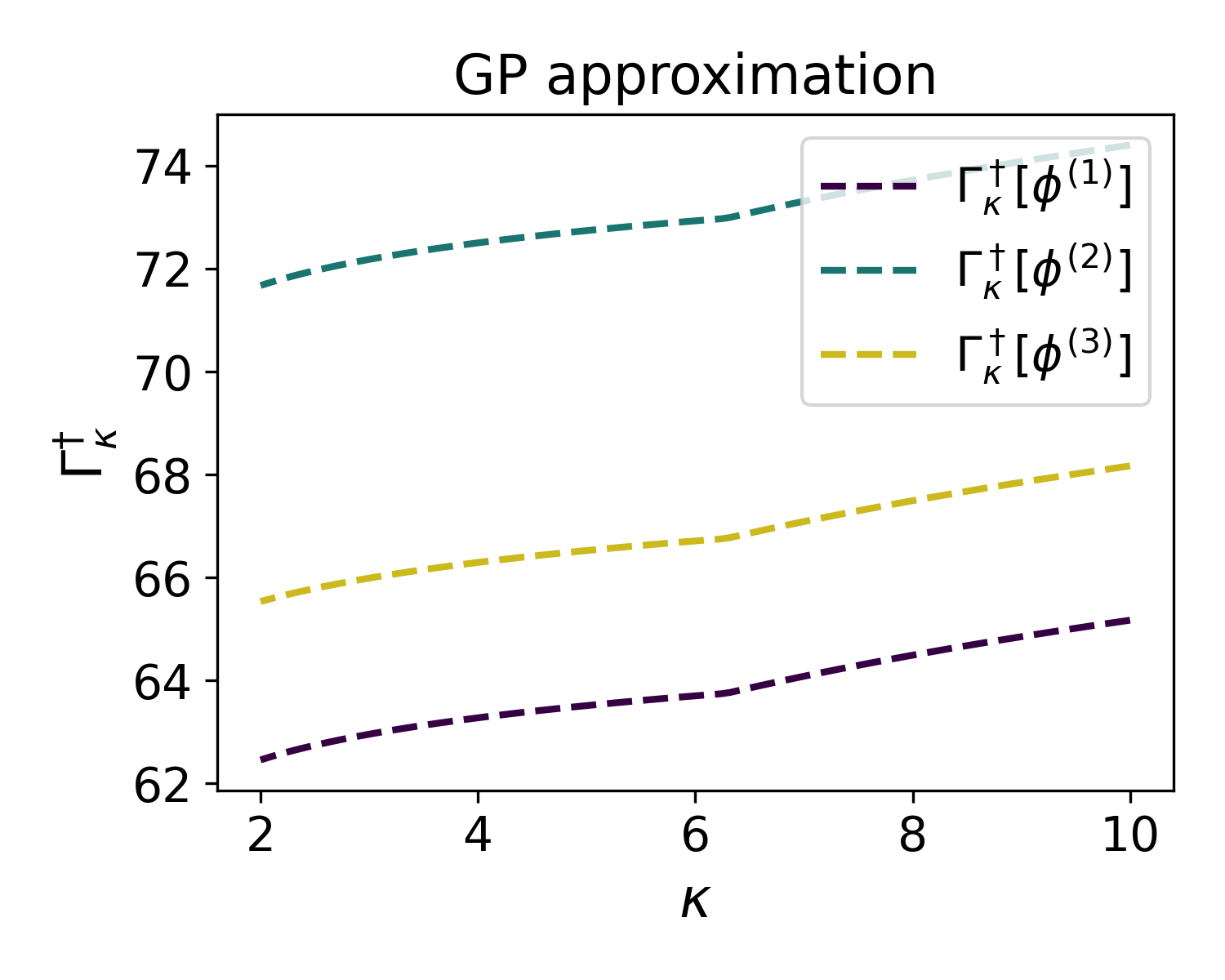}
		\caption{GP surrogate $\Gamma_{\kappa}^{\dagger}[\phi_i]$ along the same RG scales $k$.}
\label{fig:Wetterich_phi4:collocation_fields:Gamma}
	\end{subfigure} \\
	\begin{subfigure}[b]{0.32\textwidth}
		\includegraphics[width=\textwidth]{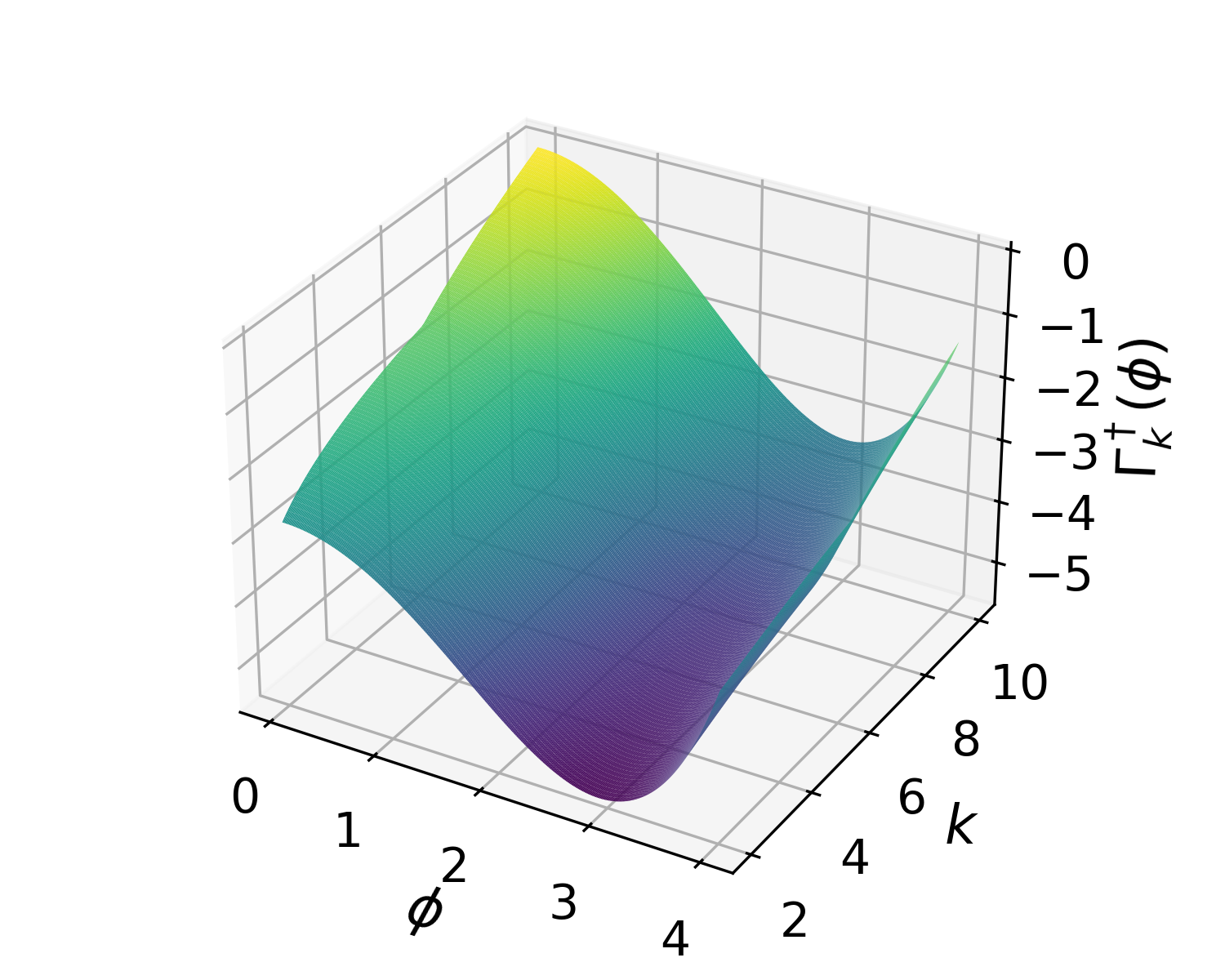}
		\caption{Surrogate $\Gamma_k^{\dagger}(\phi)$ on constant fields $\phi\in[0,4]$.}
	\label{fig:Wetterich_phi4:test_fields}
	\end{subfigure} 
	\begin{subfigure}[b]{0.32\textwidth}
		\includegraphics[width=\textwidth]{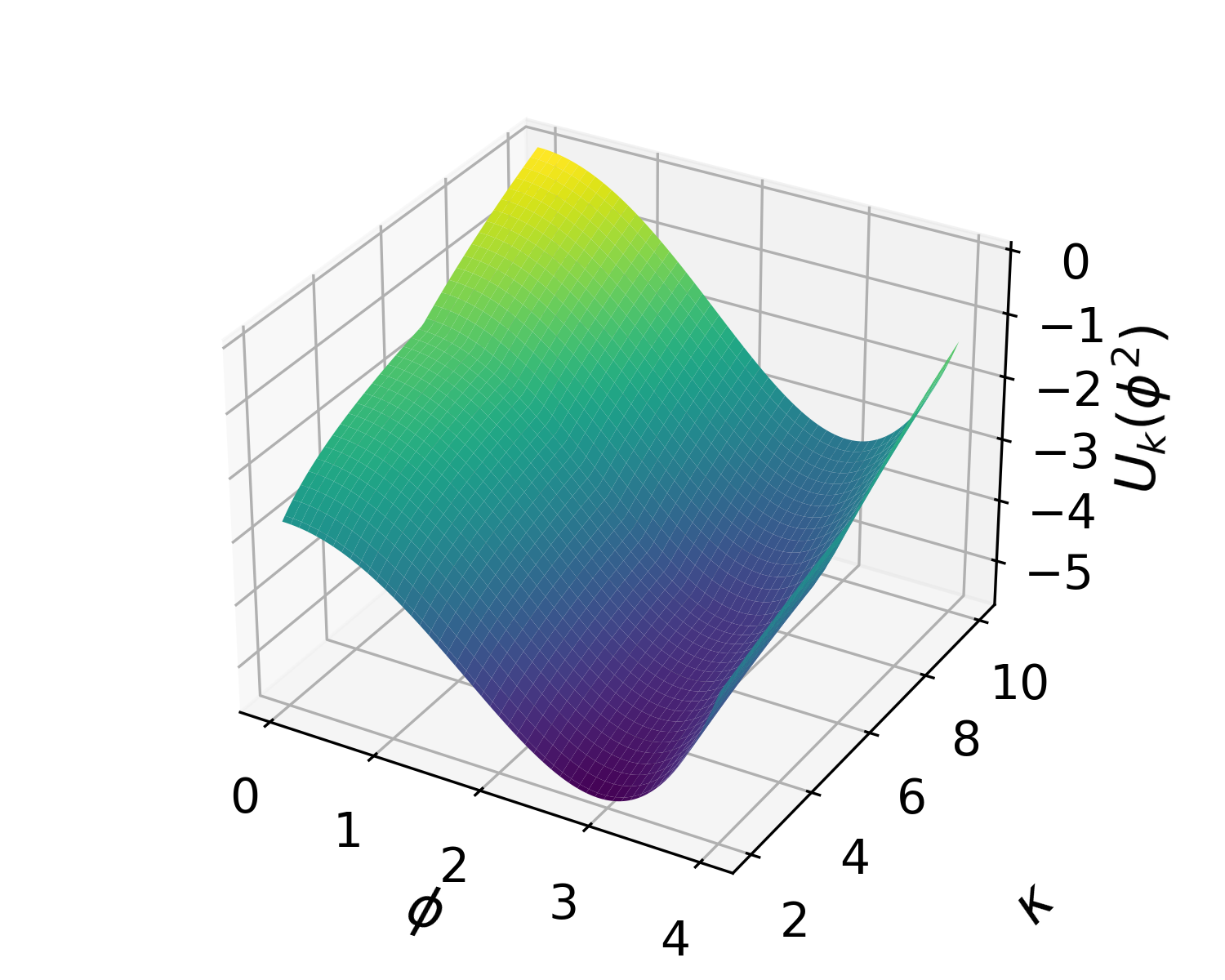}
		\caption{LPA reference $U_\kappa(\phi^{2})$ on the same constants.}
		\label{fig:Wetterich_phi4:test_fields:RL2}
	\end{subfigure} 	
    \begin{subfigure}[b]{0.32\textwidth}
		\includegraphics[width=\textwidth]{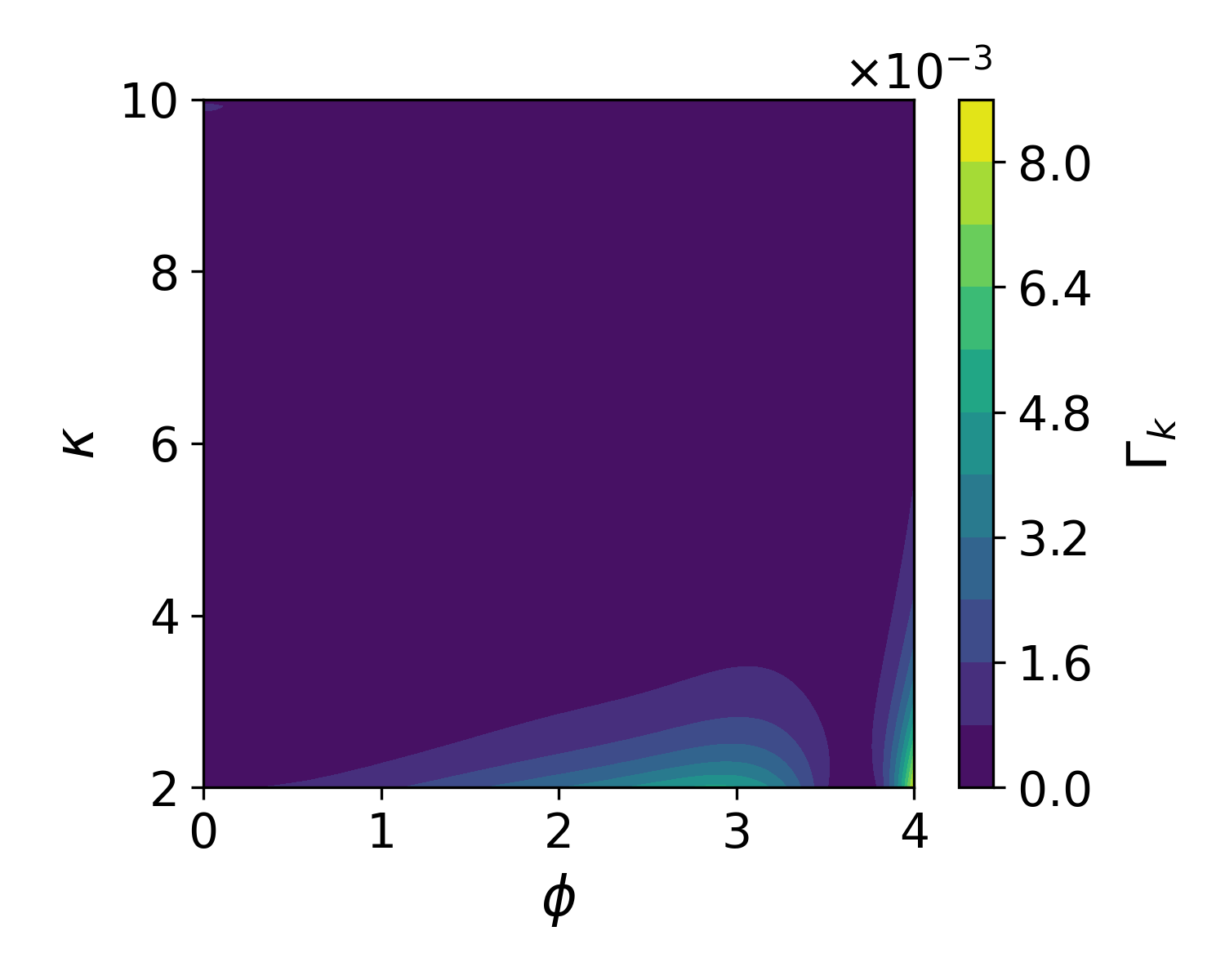}
		\caption{$\lvert\Gamma_{\kappa}^{\dagger}-U_\kappa\rvert/\lvert U_\kappa\rvert$ over $(\kappa,\phi)\in[2,10]\times[0,4]$.}
		\label{fig:Wetterich_phi4:test_fields:RL2}
	\end{subfigure}
	\caption{Numerical results for the Wetterich flow with the $\phi^4$ bare action. Panels (A)--(B) illustrate three of the $N_\text{col}=1000$ training fields and the corresponding GP predictions, while panels (C)--(D) plot the surrogate and the LPA potential on constant-field inputs in $[0, 4]$. Panel (E) maps the pointwise relative difference across RG scales $\kappa\in[2,10]$ and field amplitudes $\phi\in[0,4]$.}
	\label{fig:Wetterich:GM}
\end{figure}

\subsubsection{Numerical results and discussion}
\label{subsec:wetterich:phi4:results}

Figure~\ref{fig:Wetterich:GM} summarizes the performance of our GP-based solver for the interacting \(\phi^4\) Wetterich flow.  Panel~(A) displays three representative training fields drawn from the collocation ensemble \(\Phi\).  Panel~(B) plots the corresponding GP surrogate evaluations along the RG trajectory.  

Panel~(C) shows the surface \(\Gamma^\dagger_\kappa(\phi)\) for \(\kappa\in[2,10]\) and \(\phi\in[0,4]\), while panel~(D) presents the LPA reference potential \(U_\kappa(\phi^2)\) evaluated on the same \((\kappa,\phi)\) grid.  The two surfaces are visually indistinguishable at the scale of the plots, indicating that the functional GP, which is trained on general nonconstant fields, closely tracks the local potential extracted from a dedicated LPA computation when restricted to homogeneous configurations.

Panel~(E) quantifies this agreement by plotting the pointwise relative difference between the LPA and GP results on constant fields,
\[
  \varepsilon(\kappa,\phi)
  =
  \frac{\bigl\lvert\Gamma_\kappa^\dagger(\phi)-U_\kappa(\phi^2)\bigr\rvert}
       {\lvert U_\kappa(\phi^2)\rvert},
  \qquad
  (\kappa,\phi)\in[2,10]\times[0,4].
\]
Across this entire grid the errors remain in the range \(10^{-3}\)–\(10^{-2}\).

Taken together, these diagnostics show that the GP prior for \(\Gamma\), engineered from the LPA ansatz and the \(\phi^4\) structure of the bare action, yields a surrogate that accurately captures the RG evolution of the effective action while retaining excellent quantitative agreement with the LPA potential on constant fields.  The plots in panels~(A)--(B) demonstrate accurate interpolation on the training ensemble, while the constant-field comparison in panels~(C)–(E) plays the role of an out-of-sample test and confirms  generalization of the surrogate across the entire RG flow.

\section{Mathematical presentation of the FRG flow problem}
\label{secfre}

For a mathematical audience, it is convenient to recall a probabilistic
interpretation of the functional renormalization group (FRG), following
\cite{hudes2025}.  Let $\psi$ be an underlying ``true'' field and consider a
noisy observation
\[
  \phi = \psi + \epsilon,
\]
where $\epsilon$ is a Gaussian error with zero mean and covariance 
\GE{matrix $R_\kappa^{-1}$ or equivalently precision matrix $R_\kappa$}.  We assume that $\epsilon$ is independent of $\psi$.  By Bayes' rule,
the conditional law of $\psi$ given the observation $\phi$ is
\begin{equation}\label{eq:bayes_result}
  P[\psi \mid \phi]
  = \frac{P[\psi]\;P[\epsilon = \phi - \psi]}{P[\phi]}.
\end{equation}

To encode this conditional distribution, one introduces an external source (or
test function) $J$, which should be thought of as an element of the dual of the
field space, and defines the
conditional generating functional
\[
  Z_\kappa[J\mid \phi]
  := \mathbb{E}\!\left[\exp\!\Bigl(\int_x J(x)\,\psi(x)\,\mathrm{d}x\Bigr)
                     \,\Bigm|\, \phi\right].
\]
The associated conditional cumulant-generating functional is then
\[
  W_\kappa[J \mid \phi]
  := \ln Z_\kappa[J\mid \phi].
\]

From $W_\kappa[J \mid \phi]$ one defines a scale-dependent functional
$\Gamma_\kappa[\phi]$ via a generalized Legendre transform in the source
variable $J$:
\begin{equation*}
  \Gamma_\kappa[\phi]
  = \sup_{J}
    \left\{
      \int_x J(x)\,\phi(x)\,\mathrm{d}x
      \;-\;
      W_\kappa[J \mid \phi]
    \right\}.
\end{equation*}
In the FRG literature, $\Gamma_\kappa$ is referred to as the \emph{effective
average action}. As the scale parameter $\kappa$ varies, $\Gamma_\kappa$
interpolates between the bare (microscopic) action in the ultraviolet limit
$\kappa\to\infty,$ $R_\kappa\to \infty$ \GE{(perfect measurements)} and the full effective action in the infrared limit
$\kappa\to 0$, $R_\kappa \to O$ \GE{(no measurements)}, which governs the macroscopic behavior. Starting from the probabilistic setup above, one can derive, as shown in
\cite{hudes2025}, the Wetterich--Morris flow equation \eqref{eq: WM renorm} for
$\Gamma_\kappa$.

\section{Derivative and Hessian of kernel interpolants}
\label{app: derivative and hessian}

We now discuss the computation of functional derivatives and Hessians of the
kernel interpolant \eqref{eq:Gamma-surrogate-kappa}.  For a fixed scale
\(\kappa\) we suppress the \(\kappa\)–dependence and write the surrogate in the
form
\begin{equation}\label{eq:Gamma-kernel-appendix}
  \Gamma^\dagger[\phi]
  = \sum_{i=1}^N c_i\,K\bigl(w(\phi), w(\phi^{(i)})\bigr),
\end{equation}
where \(\phi\in\mathcal{U}\) is a generic field,
\(\phi^{(i)}\) are the collocation fields, and
\(w:\mathcal{U}\to\mathbb{R}^m\) is the feature map introduced in
\Cref{subsec:functional-evolution}.  The vector
\(c=(c_1,\dots,c_N)^\top\) collects the coefficients appearing in the
representer formula, for example
\(c = \bigl(K(Z,Z)+\varepsilon I\bigr)^{-1} g_\kappa\) in
\eqref{eq:Gamma-surrogate-kappa}.  In this appendix we treat \(c\) as fixed and
focus on the dependence on the field \(\phi\).

Throughout, we consider G\^ateaux derivatives of \(\Gamma^\dagger\) along
directions \(\eta,v\in\mathcal{U}\).
We assume that the feature map \(w\) is linear, so that
\(w(\phi+\varepsilon\eta)=w(\phi)+\varepsilon w(\eta)\) for all
\(\phi,\eta\in\mathcal{U}\) and \(\varepsilon\in\mathbb{R}\); this covers the
cases of pointwise sampling and Fourier coefficients considered below.

\subsection{First variation}

Fix a collocation field \(\phi^{(i)}\) and abbreviate
\(z = w(\phi)\in\mathbb{R}^m\), \(z^{(i)}=w(\phi^{(i)})\in\mathbb{R}^m\).
Consider the directional derivative of a single kernel term,
\[
  \frac{\mathrm{d}}{\mathrm{d}\varepsilon}
  K\bigl(w(\phi+\varepsilon\eta), w(\phi^{(i)})\bigr)\Big|_{\varepsilon=0}.
\]
Using the chain rule and linearity of \(w\) we obtain
\begin{equation}\label{eq: functional derivative kernel}
  \frac{\mathrm{d}}{\mathrm{d}\varepsilon}
  K\bigl(w(\phi+\varepsilon\eta), w(\phi^{(i)})\bigr)\Big|_{\varepsilon=0}
  = \nabla_1 K(z,z^{(i)}) \cdot w(\eta),
\end{equation}
where \(\nabla_1\) denotes the gradient of \(K(\cdot,z^{(i)})\) with respect to
its first argument, and the dot denotes the Euclidean inner product in
\(\mathbb{R}^m\).

By linearity of the sum in \eqref{eq:Gamma-kernel-appendix}, the first
variation of \(\Gamma^\dagger\) at \(\phi\) in the direction \(\eta\) is
\begin{equation*}
  \left\langle \frac{\delta\Gamma^\dagger}{\delta\phi},\eta\right\rangle =\frac{\mathrm{d}}{\mathrm{d}\varepsilon}
  \Gamma^\dagger[\phi+\varepsilon\eta]\Big|_{\varepsilon=0}
  =
  \sum_{i=1}^N c_i\,
  \nabla_1 K\bigl(w(\phi), w(\phi^{(i)})\bigr)\cdot w(\eta).
\end{equation*}
This expression for the functional derivative $\delta\Gamma^\dagger/\delta\phi$ will take a concrete form
once a specific form of the feature map
\(w\) is chosen.

\subsection{Second variation}

A similar calculation yields the second variation.  
Differentiating \eqref{eq: functional derivative kernel} once more in the
direction \(v\) gives
\begin{align*}
  \frac{\mathrm{d}}{\mathrm{d}\varepsilon}
  \Bigl[\nabla_1 K\bigl(w(\phi+\varepsilon v), w(\phi^{(i)})\bigr)\cdot w(\eta)\Bigr]
  \Big|_{\varepsilon=0}
  &= \bigl(w(\eta)\bigr)^\top\,
     \nabla_1^2 K\bigl(w(\phi), w(\phi^{(i)})\bigr)\,
     w(v),
\end{align*}
where \(\nabla_1^2 K\) denotes the Hessian of \(K(\cdot,z^{(i)})\) with respect
to its first argument.  
\XY{Summing over \(i\) with weights \(c_i\) then yields the bilinear form
\(\langle \eta,\,\delta^2\Gamma^\dagger/(\delta\phi\,\delta\phi)\, v\rangle\)
for the second variation.  
As in the first-variation case, this pairing provides an explicit expression
for the second-variation kernel $\delta^2\Gamma^\dagger/(\delta\phi\delta\phi)$ once the feature map \(w\) is specified.}  
We now record the resulting formulas for the two choices considered in this paper.


\subsection{Examples of differrent measurements} \XY{In this subsection, we detail the functional derivative calculations for kernel interpolants in two examples: pointwise measurements and measurements given by Fourier coefficients.}

Suppose the feature map is given by pointwise samples at physical collocation
points \(x_1,\dots,x_m\), so that
\[
  w(\phi) = \bigl(\phi(x_1),\dots,\phi(x_m)\bigr) \in\mathbb{R}^m.
\]
In this case
\[
  w(\eta) = \bigl(\eta(x_1),\dots,\eta(x_m)\bigr),
\]
and \eqref{eq: functional derivative kernel} becomes
\begin{align*}
  \nabla_1 K\bigl(w(\phi),w(\phi^{(i)})\bigr)\cdot w(\eta)
  &= \sum_{j=1}^m
       \partial_{z_j} K\bigl(w(\phi),w(\phi^{(i)})\bigr)\,\eta(x_j)\\
  &= \int
       \Biggl[
         \sum_{j=1}^m
           \partial_{z_j} K\bigl(w(\phi),w(\phi^{(i)})\bigr)\,
           \delta_{x_j}(x)
       \Biggr] \eta(x)\,\mathrm{d}x,
\end{align*}
where \(\partial_{z_j}\) denotes differentiation with respect to the \(j\)th
component of the first kernel argument and \(\delta_{x_j}\) is the Dirac
measure at \(x_j\).
We may therefore identify the functional derivative of a single kernel term as
\[
  \frac{\delta}{\delta \phi(x)}
  K\bigl(w(\phi),w(\phi^{(i)})\bigr)
  =
  \sum_{j=1}^m
    \partial_{z_j} K\bigl(w(\phi),w(\phi^{(i)})\bigr)\,\delta_{x_j}(x).
\]
Integrals involving \(\delta\Gamma^\dagger/\delta\phi\) are thus reduced to
evaluations of the kernel gradient at the finite set of spatial points
\(\{x_j\}_{j=1}^m\).

For the second variation we use the Hessian \(\nabla_1^2 K\) and the 
representation \(w(\eta)=(\eta(x_j))_{j=1}^m\).  A calculation analogous to
the one above yields
\begin{align*}
  &w(\eta)^\top \nabla_1^2 K\bigl(w(\phi),w(\phi^{(i)})\bigr) w(v)
  =
  \sum_{j,k=1}^m
    \eta(x_k)\,\partial_{z_k}\partial_{z_j}
    K\bigl(w(\phi),w(\phi^{(i)})\bigr)\,v(x_j)\\
  =& \int\!\!\int
       \eta(y)
       \Biggl[
         \sum_{j,k=1}^m
           \partial_{z_k}\partial_{z_j}
           K\bigl(w(\phi),w(\phi^{(i)})\bigr)\,
           \delta_{x_k}(y)\,\delta_{x_j}(x)
       \Biggr]
       v(x)\,\mathrm{d}x\,\mathrm{d}y.
\end{align*}
Hence the second functional derivative of a single kernel term is the
distribution
\[
  \frac{\delta^2}{\delta \phi(x)\,\delta \phi(y)}
  K\bigl(w(\phi),w(\phi^{(i)})\bigr)
  =
  \sum_{j,k=1}^m
    \partial_{z_k}\partial_{z_j}
    K\bigl(w(\phi),w(\phi^{(i)})\bigr)\,
    \delta_{x_k}(y)\,\delta_{x_j}(x).
\]
Again, any integral involving the Hessian of \(\Gamma^\dagger\) only depends on
the physical collocation points \(\{x_j\}\) and on the kernel Hessian
evaluated at the feature vector \(w(\phi)\).

Next, we consider the case where \(w\) records Fourier coefficients.  For
simplicity, let \(\{e_j\}_{j=1}^m\) be a finite family of (complex or real)
Fourier modes on the spatial domain and set
\[
  w(\phi)
  = \bigl(\hat \phi_j\bigr)_{j=1}^m,
  \qquad
  \hat \phi_j := \int \phi(x)\,e_j(x)\,\mathrm{d}x.
\]
Then
\[
  w(\eta) = \bigl(\hat \eta_j\bigr)_{j=1}^m,
  \qquad
  \hat \eta_j := \int \eta(x)\,e_j(x)\,\mathrm{d}x.
\]
Proceeding as in the pointwise case, we obtain from
\eqref{eq: functional derivative kernel}
\begin{align*}
  \nabla_1 K\bigl(w(\phi),w(\phi^{(i)})\bigr)\cdot w(\eta)
  &= \sum_{j=1}^m
       \partial_{z_j} K\bigl(w(\phi),w(\phi^{(i)})\bigr)\,\hat \eta_j\\
  &= \int
       \Biggl[
         \sum_{j=1}^m
           \partial_{z_j} K\bigl(w(\phi),w(\phi^{(i)})\bigr)\,e_j(x)
       \Biggr] \eta(x)\,\mathrm{d}x.
\end{align*}
Thus
\[
  \frac{\delta}{\delta \phi(x)}
  K\bigl(w(\phi),w(\phi^{(i)})\bigr)
  =
  \sum_{j=1}^m
    \partial_{z_j} K\bigl(w(\phi),w(\phi^{(i)})\bigr)\,e_j(x).
\]

For the second derivative we obtain, by the same argument,
\begin{align*}
  &w(\eta)^\top \nabla_1^2 K\bigl(w(\phi),w(\phi^{(i)})\bigr) w(v)
  =
  \sum_{j,k=1}^m
    \hat \eta_k\,
    \partial_{z_k}\partial_{z_j}
    K\bigl(w(\phi),w(\phi^{(i)})\bigr)\,
    \hat v_j\\
  =& \int\!\!\int
       \eta(y)
       \Biggl[
         \sum_{j,k=1}^m
           \partial_{z_k}\partial_{z_j}
           K\bigl(w(\phi),w(\phi^{(i)})\bigr)\,
           e_k(y)\,e_j(x)
       \Biggr]
       v(x)\,\mathrm{d}x\,\mathrm{d}y,
\end{align*}
so that
\[
  \frac{\delta^2}{\delta \phi(x)\,\delta \phi(y)}
  K\bigl(w(\phi),w(\phi^{(i)})\bigr)
  =
  \sum_{j,k=1}^m
    \partial_{z_k}\partial_{z_j}
    K\bigl(w(\phi),w(\phi^{(i)})\bigr)\,
    e_k(y)\,e_j(x).
\]

In both the pointwise and Fourier settings, the functional derivatives and
Hessians of \(\Gamma^\dagger\) are obtained by summing the corresponding
expressions over \(i=1,\dots,N\) with weights \(c_i\), as in
\eqref{eq:Gamma-kernel-appendix}.

\section{Local Potential Approximation (LPA)}
\label{appendix:LPA_intro}

We consider the effective average action \(\Gamma_\kappa[\phi]\) satisfying the Wetterich flow
\[
\partial_\kappa \Gamma_\kappa[\phi]
=\frac{1}{2}\,\mathrm{Tr}\!\Big[\big(\Gamma_\kappa^{(2)}[\phi]+R_\kappa\big)^{-1}\,\partial_\kappa R_\kappa\Big],
\qquad
\Gamma_\kappa^{(2)}[\phi](x,y)=\frac{\delta^2\Gamma_\kappa[\phi]}{\delta\phi(x)\,\delta\phi(y)},
\]
where the renormalization scale \(\kappa\) flows backward from a large terminal value \(\kappa_{\mathrm{UV}}\) to an infrared value \(\kappa_{\mathrm{IR}}\), and \(R_\kappa\) is a symmetric, translation-invariant regulator.

In the local potential approximation (LPA), \GE{one assumes} an ansatz with a standard quadratic gradient term encoding all scale dependence in a spatially local potential \cite{Delamotte_2012}. Concretely, on a periodic continuum domain \(\Omega\subset\mathbb{R}\) (for example \([0,1]\) with periodic boundary conditions)
\[
\Gamma_\kappa[\phi]
=\int_{\Omega}\!\Bigl(\tfrac{1}{2}\,|\partial_x\phi(x)|^2 + U_\kappa(\phi(x))\Bigr)\,\mathrm{d}x,
\]
and on a periodic lattice with \(N\) sites, spacing \(h\), and index set \(\mathbb{T}_N\), one sets
\[
\Gamma_\kappa[\phi]
=\sum_{i\in\mathbb{T}_N}\Bigl(\tfrac{1}{2}\,\bigl|(\nabla_h \phi)_i\bigr|^2 + U_\kappa(\phi_i)\Bigr)\,h,
\qquad
(\nabla_h \phi)_i:=\frac{\phi_{i+1}-\phi_i}{h}.
\]
The scalar function \(U_\kappa\)  is the scale-dependent effective potential to be determined. 
At a constant configuration \(\phi(x)\equiv\varphi\in\mathbb{R}\), the second variational derivative takes the form
\[
\Gamma_\kappa^{(2)}[\varphi]=-\Delta+U_\kappa''(\varphi)\,I \quad \text{(continuum)}, 
\qquad
\Gamma_\kappa^{(2)}[\varphi]=-\Delta_h+U_\kappa''(\varphi)\,I \quad \text{(lattice)},
\]
where \(-\Delta\) is the Laplacian on \(\Omega\) with periodic boundary conditions and \(-\Delta_h\) is the standard second–order discrete Laplacian.

Because the regulator is translation-invariant, both \(\Gamma_\kappa^{(2)}[\varphi]\) and \(R_\kappa\) are diagonal in a Fourier basis.  
We therefore choose a real orthonormal eigenbasis of the (continuous or discrete) Laplacian,
\[
  -\Delta\,e_{\alpha,j} = \lambda_\alpha\,e_{\alpha,j}
  \quad\text{or}\quad
  -\Delta_h\,e_{\alpha,j} = \lambda_\alpha\,e_{\alpha,j},
\]
where \(\alpha\in\mathcal{I}\) labels the distinct eigenvalues \(\lambda_\alpha\ge 0\), and
\(j=1,\dots,g_\alpha\) runs over an orthonormal basis of the corresponding eigenspace of multiplicity \(g_\alpha\in\mathbb{N}\).

By translation invariance, the regulator acts in mode:
\[
R_\kappa e_{\alpha,j} = r_\kappa(\lambda_\alpha)\,e_{\alpha,j},
\qquad
(\partial_\kappa R_\kappa)\,e_{\alpha,j} = \partial_\kappa r_\kappa(\lambda_\alpha)\,e_{\alpha,j},
\]
for some scalar profile \(\lambda\mapsto r_\kappa(\lambda)\).
At a constant field \(\varphi\), the operator inside the trace in the Wetterich equation is diagonal in the same basis:
\[
\big(\Gamma_\kappa^{(2)}[\varphi]+R_\kappa\big)e_{\alpha,j}
=\big(\lambda_\alpha + U_\kappa''(\varphi) + r_\kappa(\lambda_\alpha)\big)\,e_{\alpha,j},
\]
and hence
\[
\big(\Gamma_\kappa^{(2)}[\varphi]+R_\kappa\big)^{-1}e_{\alpha,j}
=\frac{1}{\lambda_\alpha + r_\kappa(\lambda_\alpha)+U_\kappa''(\varphi)}\,e_{\alpha,j}.
\]
Therefore, the trace reduces to a sum over distinct eigenvalues,
\[
\mathrm{Tr}\!\Big[\big(\Gamma_\kappa^{(2)}[\varphi]+R_\kappa\big)^{-1}\,\partial_\kappa R_\kappa\Big]
=\sum_{\alpha\in\mathcal{I}} g_\alpha\,
\frac{\partial_\kappa r_\kappa(\lambda_\alpha)}
     {\lambda_\alpha + r_\kappa(\lambda_\alpha) + U_\kappa''(\varphi)}.
\]

To treat the continuum and lattice cases uniformly, we work with the potential per unit volume (continuum) or per site (lattice). For a constant field \(\varphi\), define
\[
U_\kappa(\varphi)
:= \frac{1}{\mathrm{vol}(\Omega)}\,\Gamma_\kappa[\phi\equiv\varphi],
\qquad
\mathrm{vol}(\Omega):=
\begin{cases}
|\Omega|, & \text{continuum},\\[4pt]
N, & \text{lattice},
\end{cases}
\]
so that \(U_\kappa\) always denotes the effective potential normalized by the size of the domain.
With this convention,
\[
\partial_\kappa U_\kappa(\varphi)
=\frac{c_\Omega}{2}\sum_{\alpha\in\mathcal{I}} g_\alpha\,
\frac{\partial_\kappa r_\kappa(\lambda_\alpha)}
     {\lambda_\alpha + r_\kappa(\lambda_\alpha) + U_\kappa''(\varphi)},
\qquad
c_\Omega:=\frac{1}{\mathrm{vol}(\Omega)}.
\]
This is the LPA equation. It applies to both settings; in particular \(c_\Omega=1\) on \([0,1]\) and \(c_\Omega=1/N\) on a lattice when \(U_\kappa\) is defined per site.

A commonly used choice is the Litim profile, defined modewise by
\[
r_\kappa(\lambda)=(\kappa^2-\lambda)_+,
\qquad
\partial_\kappa r_\kappa(\lambda)=2\kappa\,\mathbf{1}_{\{\lambda<\kappa^2\}}.
\]
On the \emph{active} set \(\{\lambda_\alpha<\kappa^2\}\) the denominator simplifies to \(\kappa^2+U_\kappa''(\varphi)\).  
Introducing the active-mode count
\[
S(\kappa):=\sum_{\lambda_\alpha<\kappa^2} g_\alpha,
\]
the equation collapses to
\[
\partial_\kappa U_\kappa(\varphi)
=c_\Omega\,\kappa\,\frac{S(\kappa)}{\kappa^2+U_\kappa''(\varphi)}.
\]
Integrating this equation backward from \(\kappa_{\mathrm{UV}}\) to \(\kappa_{\mathrm{IR}}\) yields the infrared potential \(U_{\kappa_{\mathrm{IR}}}\), which can then be used to compute physical observables.
\section{Transfer Matrix Approach}
\label{app_sec:TM}

We outline the transfer-matrix computation used to obtain reference solutions
for the one-dimensional $\phi^4$ model studied in
\Cref{subsec:num:wetterich_lattice}.  We consider a one-dimensional periodic lattice with $N_x$ sites and lattice
spacing $a>0$.  A field configuration is
$\phi = (\phi_1,\dots,\phi_{N_x}) \in \mathbb{R}^{N_x}$, and the bare action in
the presence of an external source $c = (c_x)_{x=1}^{N_x}$ is
\[
  S[\phi;c]
  =
  a \sum_{x=1}^{N_x}
  \left[
    \frac{1}{2}
      \left(\frac{\phi_{x+1}-\phi_x}{a}\right)^2
    + U(\phi_x;c_x)
  \right],
  \qquad
  \phi_{N_x+1} = \phi_1,
\]
with local potential
\[
  U(\phi;c_x)
  :=
  \frac{1}{2} m^2 \phi^2
  + \frac{\lambda}{24}\,\phi^4
  - c_x \phi.
\]
The partition function and cumulant-generating functional with $\mathcal{D}\phi:=\prod_{x=1}^{N-x} d\phi_x$
are
\[
  Z[c]
  :=
  \int \mathrm{e}^{-S[\phi;c]}\,\mathcal{D}\phi,
  \qquad
  W[c] := \ln Z[c].
\]

To obtain a finite-dimensional representation, we discretise the range of field
values.  We choose an interval $[-\phi_{\mathrm{max}},\phi_{\mathrm{max}}]$ and a grid
\[
  \{\phi^{(1)},\dots,\phi^{(N_\phi)}\} \subset [-\phi_{\mathrm{max}},\phi_{\mathrm{max}}],
\]
and restrict the field at each site to this grid,
\[
  \phi_x \in \{\phi^{(1)},\dots,\phi^{(N_\phi)}\},
  \qquad x=1,\dots,N_x.
\]
For this discretized model the path integral becomes a finite sum and can be
written exactly in terms of a transfer matrix. \XY{We comment that this discretization may introduce inaccuracies when \(|\phi|>\phi_{\mathrm{max}}\)} \GE{but is chosen here for consistency with the kernel approximation.}

The transfer matrix $T_x$ encodes the Boltzmann weight associated with the 
nearest-neighbour pair
between sites $x$ and $x+1$.  It is an $N_\phi\times N_\phi$ matrix with
entries
\begin{equation}\label{eq:naive_transfer_matrix}
  (T_x)_{ij}
  :=
  \exp\!\left[
    -\frac{a}{2}
    \left(
      \left(\frac{\phi^{(j)}-\phi^{(i)}}{a}\right)^2
      + U(\phi^{(i)};c_x) + U(\phi^{(j)};c_{x+1})
    \right)
  \right],
  \qquad i,j = 1,\dots,N_\phi.
\end{equation}
The kinetic term $(\phi_{x+1}-\phi_x)^2$ couples neighbouring field values
$\phi^{(i)}$ and $\phi^{(j)}$, while the local potentials at sites $x$ and
$x+1$ are split symmetrically between the two ends of the bond.  The partition
function of the discretised model is
\[
  Z[c]
  = \operatorname{Tr}\bigl(T_1 T_2 \cdots T_{N_x}\bigr),
  \qquad
  W[c] = \ln \operatorname{Tr}\bigl(T_1 T_2 \cdots T_{N_x}\bigr).
\]

Let
\[
  \phi_x[c] := \langle \phi_x \rangle_c
\]
denote the expectation value of the field at site $x$ in the Gibbs measure with
weight $\mathrm{e}^{-S[\phi;c]}$.  Since the source couples as $-c_x\phi_x$, we
have
\[
  \frac{\delta W[c]}{\delta c_x}
  = \bigl\langle \phi_x \bigr\rangle_c
  = \phi_x[c].
\]

In the transfer-matrix representation, introduce the diagonal matrix
\[
  \Phi_{ij} := \phi^{(i)} \delta_{ij},
  \qquad i,j = 1,\dots,N_\phi,
\]
which represents multiplication by the local field value on field-space
indices.  Differentiating \eqref{eq:naive_transfer_matrix} gives
\[
  \frac{\partial T_x}{\partial c_x}
  = a\,\Phi\,T_x.
\]
Using $Z[c] = \operatorname{Tr}(T_1\cdots T_{N_x})$ and cyclicity of the trace,
we obtain
\begin{equation}
  \phi_x[c]
  = \frac{\delta W[c]}{\delta c_x}
  =
  \frac{\operatorname{Tr}\bigl(\Phi\,T_1\cdots T_{N_x}\bigr)}
       {\operatorname{Tr}\bigl(T_1\cdots T_{N_x}\bigr)}.
  \label{eq:TM-phi-expectation}
\end{equation}
Here $\Phi$ is an operator (a diagonal matrix); the scalar $\phi_x[c]$ is
obtained after taking the trace.

For the observables used in this work (magnetization and susceptibility), it
suffices to consider homogeneous sources $c_x \equiv c$.  Then all transfer
matrices coincide, $T_x = T(c)$, and translation invariance implies that
$\phi_x[c]$ is independent of $x$.  We denote this common value by $\bar\phi(c)$
and identify it with the magnetisation per site,
\[
  m(c)
  :=
  \frac{1}{N_x}\sum_{x=1}^{N_x} \langle \phi_x \rangle_c
  = \bar\phi(c)
  =
  \frac{\operatorname{Tr}\bigl(\Phi\,T(c)^{N_x}\bigr)}
       {\operatorname{Tr}\bigl(T(c)^{N_x}\bigr)}.
\]
Thus, in our notation the magnetisation per site is precisely the expectation
value of the local field.

The effective action $\Gamma[\phi]$ is the Legendre transform of $W[c]$,
\[
  \Gamma[\phi]
  = \sum_{x=1}^{N_x} c_x\,\phi_x[c] - W[c],
  \qquad
  \phi_x[c] = \frac{\delta W[c]}{\delta c_x}.
\]
For a homogeneous configuration $\phi_x \equiv \phi$ and $c_x \equiv c$,  
the effective potential
\[
  U(\phi)
  := \frac{\Gamma[\phi]}{N_x}
\]
is the central quantity used to compute the susceptibility.
In the double limit $N_\phi\to\infty$ and $\phi_{\mathrm{max}}\to\infty$, with a correspondingly
refined field-space grid, $U(\phi)$ converges to the exact effective potential
of the continuum model.

If \eqref{eq:naive_transfer_matrix} is used directly in numerical computations,
overflow can occur for large homogeneous sources $c$.  To avoid this, we use an
equivalent transfer matrix based on a shifted local potential.  Fix a reference
field value $\phi_0$ and introduce
\begin{equation}
  U_2(\phi)
  := \frac{1}{2} m^2 \phi^2
     + \frac{\lambda}{24}\,\phi^4
     - c\,(\phi - \phi_0).
\end{equation}
The corresponding transfer matrix is
\begin{align}
    (T_2)_{ij}
    := \exp\!\left[
      -\frac{a}{2}\left(
        \left(\frac{\phi^{(j)}-\phi^{(i)}}{a}\right)^2
        + U_2(\phi^{(i)}) + U_2(\phi^{(j)})
      \right)
    \right].
\end{align}
This shifted formulation is algebraically equivalent to the original one but is
numerically more stable for large $|c|$.

\section{Details of the GP representation for the Wetterich flow}
\label{app:wetterich_gp_details}
In this section, we present the implementation details for solving the Wetterich equation on the lattice  in \Cref{subsec:num:wetterich_lattice}.
\subsection{GP parametrization of the local potential}

In our numerical implementation, the scale-dependent local potential \(U_\kappa\) on a field window \([-\phi_{\mathrm{max}},\phi_{\mathrm{max}}]\) 
 is modeled on the continuum as
\begin{equation}
\label{eq:parametrized_Uk}
U_\kappa(\phi) = m_0(\phi) + v_\kappa(\phi) + g_\kappa(\phi),\qquad \phi\in[-\phi_{\mathrm{max}},\phi_{\mathrm{max}}],
\end{equation}
with three distinct contributions. The baseline
\[
m_0(\phi):=\tfrac12 m^2\phi^2+\tfrac{\lambda}{24}\phi^4
\]
is the bare lattice \(\phi^4\) potential, fixed for all \(\kappa\). The parametric correction
\[
v_\kappa(\phi)
=
\operatorname{softplus}(\theta_{2,\kappa})\,\phi^2/2
+
\operatorname{softplus}(\theta_{4,\kappa})\,\phi^4/12
\]
tracks the running of the quadratic and quartic couplings through a latent parameter vector \(\theta_\kappa=(\theta_{2,\kappa},\theta_{4,\kappa})^\top\in\mathbb{R}^2\). The map \(\operatorname{softplus}(x)=\log(1+e^x)\) is smooth and strictly positive, so that the effective couplings in \(v_\kappa\) are positive by construction. The remainder
\[
g_\kappa \sim \mathcal{GP}(0,K_U)
\]
is a zero-mean GP with a fixed covariance kernel \(K_U\) on \([-\phi_{\mathrm{max}},\phi_{\mathrm{max}}]\), given concretely below.
The latter term captures non-polynomial and higher-order structure that is not represented by the quadratic-quartic component.

A direct use of \eqref{eq:parametrized_Uk} in the Wetterich equation \eqref{eq:wett:lattice} would proceed as follows. At each RG scale \(\kappa\), the flow provides the values of \(U_\kappa\) at a set of field points, and from these data one would infer the latent parameters \(\theta_\kappa\) and the GP remainder \(g_\kappa\) that best explain the current profile. The curvature \(U_\kappa''\) entering \eqref{eq:wett:lattice} is then evaluated from the reconstructed \(U_\kappa\). In practice, this procedure requires solving, at each scale, a nonlinear constrained estimation problem for \(\theta_\kappa\) coupled to GP interpolation conditions for \(g_\kappa\). This is too expensive for long RG trajectories.

{
To obtain a stable and efficient solver, we use an interleaved predictor-projector scheme: on each recorded scale interval we (i) advance the flow using a linear-Gaussian surrogate that yields an explicit curvature closure, and then (ii) project the predicted profile onto the nonlinear constrained GP model. The projection replaces the predicted state and is used as the initial condition for the next predictor step. 

We discretize the field window $[-\phi_{\max},\phi_{\max}]$ at $N_\phi$ points $\{\phi_j\}_{j=1}^{N_\phi}$ and collect the sampled potentials in the vector
\[
Y_\kappa := \big(U_\kappa(\phi_j)\big)_{j=1}^{N_\phi} \in \mathbb{R}^{N_\phi}.
\]
We solve the Wetterich equation \eqref{eq:wett:lattice} using the local ansatz \eqref{eq:local} for $\Gamma_\kappa$. The resulting ODE is integrated from $\kappa_{\mathrm{UV}}$ down to $\kappa_{\mathrm{IR}}$ and recorded on a discrete scale grid,
\[
\{(\kappa_s,Y_s)\}_{s=0}^S,\qquad
\kappa_0=\kappa_{\mathrm{UV}}>\dots>\kappa_S=\kappa_{\mathrm{IR}},
\]
with $Y_s=\big(U_{\kappa_s}(\phi_j)\big)_{j=1}^{N_\phi}$.

\subsection*{Prediction: linear-Gaussian surrogate}
On each interval $[\kappa_s,\kappa_{s+1}]$ we first advance the flow with a linear-Gaussian surrogate. Concretely, we relax the softplus parametrization and write
\[
v_\kappa(\phi)=a_{2,\kappa}\,\phi^2/2+a_{4,\kappa}\,\phi^4/12,
\qquad
a_\kappa :=
\begin{bmatrix} a_{2,\kappa}\\[2pt] a_{4,\kappa}\end{bmatrix}\in\mathbb{R}^2,
\]
and place a Gaussian prior on the coefficients,
$a_\kappa\sim\mathcal{N}(0,\Sigma_a)$, where $\Sigma_a$ is a fixed $2\times2$ covariance matrix (diagonal and small in the experiments). The remainder is modeled as
\[
g_\kappa \sim \mathcal{GP}(0,K_U),
\]
independent of $a_\kappa$, where $K_U$ denotes the kernel used in the surrogate. Introducing the feature map
\[
f(\phi):=(\phi^2/2,\phi^4/12)^\top,
\]
the surrogate GP model reads
\begin{equation}
\label{eq:linearized_model}
U_\kappa(\phi)=m_0(\phi)+f(\phi)^\top a_\kappa+g_\kappa(\phi).
\end{equation}
Under these assumptions, the induced prior on $U_\kappa$ in the prediction step is
\begin{equation}
\label{eq:relaxedUk}
U_\kappa \sim \mathcal{GP}(m_0,K^{\mathrm{I}}),\qquad
K^{\mathrm{I}}(\phi,\phi') := f(\phi)^\top \Sigma_a f(\phi') + K_U(\phi,\phi').
\end{equation}
This model is Gaussian in the latent variables $(a_\kappa,g_\kappa)$ and leads to a linear dependence of the curvature $U_\kappa''$ on the potential samples $Y_\kappa$. Integrating the resulting closed ODE from $\kappa_s$ to $\kappa_{s+1}$ produces a predicted profile, which we denote by $\tilde Y_{s+1}$.

\subsection*{Correction: constrained softplus GP reconstruction}
At the correction step, we return to the nonlinear GP model \eqref{eq:parametrized_Uk} and use the predictor output $\tilde Y_{s+1}$ as data to reconstruct a constrained representation of $U_{\kappa_{s+1}}$. Recall that the discrete profile collects the sampled values
\[
(Y_{s+1})_j \;=\; U_{\kappa_{s+1}}(\phi_j),\qquad j=1,\dots,N_\phi,
\]
on the fixed field grid $X:=\{\phi_j\}_{j=1}^{N_\phi}\subset[-\phi_{\max},\phi_{\max}]$. At each scale $\kappa_{s+1}$ we adopt the nonlinear GP model
\[
U_{\kappa_{s+1}}(\phi)
= m_0(\phi)
+ \operatorname{softplus}\big(\theta_{2,\kappa_{s+1}}\big)\,\phi^2/2
+ \operatorname{softplus}\big(\theta_{4,\kappa_{s+1}}\big)\,\phi^4/12
+ g_{\kappa_{s+1}}(\phi),
\]
where $(\theta_{2,\kappa_{s+1}},\theta_{4,\kappa_{s+1}})^\top\in\mathbb{R}^2$ are latent parameters and
\[
g_{\kappa_{s+1}} \sim \mathcal{GP}(0,K_U).
\]
The GP remainder $g_{\kappa_{s+1}}$ captures the nonpolynomial structure that cannot be represented by a quartic polynomial. For a fixed $\theta_{\kappa_{s+1}}$, the polynomial part is deterministic, and we interpret the residual at the grid points,
\[
r_{\theta_{\kappa_{s+1}}}(\phi_j)
:= (\tilde Y_{s+1})_j
- m_0(\phi_j)
- \operatorname{softplus}\big(\theta_{2,\kappa_{s+1}}\big)\,\phi_j^2/2
- \operatorname{softplus}\big(\theta_{4,\kappa_{s+1}}\big)\,\phi_j^4/12,
\]
as observations of $g_{\kappa_{s+1}}$ at $\phi_j$. Standard GP regression with kernel $K_U$ then yields a posterior mean $g_{\kappa_{s+1},\mathrm{post}}(\,\cdot\,;\theta_{\kappa_{s+1}})$, and the size of the implied correction is quantified by the squared RKHS norm
$\big\|g_{\kappa_{s+1},\mathrm{post}}(\,\cdot\,;\theta_{\kappa_{s+1}})\big\|_{K_U}^2$.

We determine $\theta_{\kappa_{s+1}}$ by solving the two--dimensional optimization problem
\begin{equation}
\label{eq:theta_objective}
\min_{\theta_{\kappa_{s+1}}\in\mathbb{R}^2}\;
\frac{1}{2}\,\big\|g_{\kappa_{s+1},\mathrm{post}}(\,\cdot\,;\theta_{\kappa_{s+1}})\big\|_{K_U}^2
\;+\;\frac{\gamma}{2}\,\big\|\theta_{\kappa_{s+1}}-\theta_{\kappa_s}\big\|_2^2,
\end{equation}
where $\gamma>0$ is a small regularization parameter and $\theta_{\kappa_s}$ denotes the parameter obtained at the previous scale. In practice, \eqref{eq:theta_objective} is minimized at each scale by a damped Gauss--Newton iteration. The resulting minimizer $\theta_{\kappa_{s+1}}$ and posterior mean $g_{\kappa_{s+1},\mathrm{post}}$ define the corrected profile $Y_{s+1}$, which is then used as the initial condition for the next prediction step.

In particular, the infrared parameters $\theta_{\mathrm{IR}}:=\theta_{\kappa_S}$ and remainder $g_{\mathrm{IR}}:=g_{\kappa_S,\mathrm{post}}$ define the reconstructed infrared potential
\[
U_{\mathrm{IR}}(\phi)
= m_0(\phi)
+ \operatorname{softplus}\big(\theta_{\mathrm{IR},2}\big)\,\phi^2/2
+ \operatorname{softplus}\big(\theta_{\mathrm{IR},4}\big)\,\phi^4/12
+ g_{\mathrm{IR}}(\phi),
\]
from which all physical observables, including the magnetization and susceptibility used in Section~\ref{subsec:num:wetterich_lattice}, are subsequently computed.
}

\section{Magnetization and Susceptibility}
\label{app_sec:mag_sus}
In the infrared, the scalar effective potential \(U_{\kappa_{\mathrm{IR}}}\) encodes equilibrium thermodynamics for spatially homogeneous configurations. Introducing a homogeneous external source \(c\) that couples linearly to the field selects a preferred orientation and probes the system’s linear response. The \emph{magnetization} \(m(c)\) is the thermodynamic order parameter conjugate to \(c\); it captures symmetry breaking and phase behavior and is defined as the equilibrium field that minimizes the tilted potential
\[
\mathcal{F}_c(\phi):=U_{\kappa_{\mathrm{IR}}}(\phi)-c\,\phi,
\qquad
m(c):=\operatorname*{arg\,min}_{\phi\in\mathbb{R}}\ \mathcal{F}_c(\phi).
\]
Whenever \(U_{\kappa_{\mathrm{IR}}}\) is twice differentiable and strictly convex at \(m(c)\), first-order optimality yields the equivalent scalar equation
\[
U_{\kappa_{\mathrm{IR}}}'\big(m(c)\big)=c,\qquad U_{\kappa_{\mathrm{IR}}}''\big(m(c)\big)>0.
\]
The \emph{susceptibility} \(\chi(c)\) quantifies the linear response of the order parameter to the source and is given by
\[
\chi(c):=\frac{d\,m(c)}{dc}=\frac{1}{U_{\kappa_{\mathrm{IR}}}''\big(m(c)\big)}.
\]
These two quantities are standard diagnostics in statistical physics: \(m(c)\) tracks the emergence of spontaneous order, while \(\chi(c)\) measures fluctuations and exhibits enhancement near criticality (through small curvature \(U_{\kappa_{\mathrm{IR}}}''\)). They are also natural validation targets for our reconstruction of \(U_{\kappa_{\mathrm{IR}}}\), because they depend on both its shape and its curvature and admit high-quality lattice Monte Carlo and transfer-matrix benchmarks for direct comparison.

\bibliographystyle{plain}
\bibliography{bibliography}

\end{document}